\documentclass[runningheads]{llncs}
\PassOptionsToPackage{nocompress}{cite}
\PassOptionsToPackage{pagebackref,colorlinks=true,citecolor=blue,linkcolor=blue,urlcolor=blue}{hyperref}
\usepackage[final,year=2026,ID=8707]{eccv}
\usepackage{eccvabbrv}
\usepackage{graphicx}
\usepackage{booktabs}
\usepackage[accsupp]{axessibility}  
\usepackage{float}
\usepackage{hyperref}
\usepackage{orcidlink}
\usepackage{comment}
\usepackage{kotex}
\usepackage[table]{xcolor}
\usepackage{colortbl}
\usepackage{multirow}
\usepackage{adjustbox}

\definecolor{best}{RGB}{244,204,204}
\definecolor{second}{RGB}{252,229,205}
\definecolor{third}{RGB}{255,242,204}
\begin{document}
\AtBeginDocument{\hypersetup{citecolor=blue,linkcolor=blue,urlcolor=blue}}

\title{TRiGS: Temporal Rigid-Body Motion for Scalable 4D Gaussian Splatting} 

\titlerunning{Abbreviated paper title}

\author{%
Suwoong Yeom\inst{1}$^{*}$\orcidlink{0009-0007-4684-4785} \and
Joonsik Nam\inst{1}$^{*}$\orcidlink{0009-0001-7035-5603} \and
Seunggyu Choi\inst{1}$^{*}$\orcidlink{0009-0004-1901-9868} \and\\
Lucas~Yunkyu~Lee\inst{3}\orcidlink{0009-0003-3632-1105} \and
Sangmin Kim\inst{4}\orcidlink{0009-0003-7018-1724} \and
Jaesik Park\inst{4}\orcidlink{0000-0001-5541-409X} \and
Joonsoo Kim\inst{5}\orcidlink{0000-0002-6470-0773} \and\\
Kugjin~Yun\inst{5}\orcidlink{0009-0002-7574-2853} \and
Kyeongbo~Kong\inst{2}$^{\dagger}$\orcidlink{0000-0002-1135-7502} \and
Sukju Kang\inst{1}$^{\dagger}$\orcidlink{0000-0002-4809-956X}}

\authorrunning{F.~Author et al.}

\institute{%
Sogang University \and
Pusan National University \and
POSTECH \and
Seoul National University \and
Electronics and Telecommunications Research Institute\\
\small $^{*}$ Indicates Equal Contribution \quad $^{\dagger}$ Corresponding Authors\\
\small Project Page: \url{https://wwwjjn.github.io/TRiGS-project_page/}}

\maketitle

\vspace{-10pt}
\begin{figure}[H]
    \centering
    \includegraphics[width=0.95\textwidth]{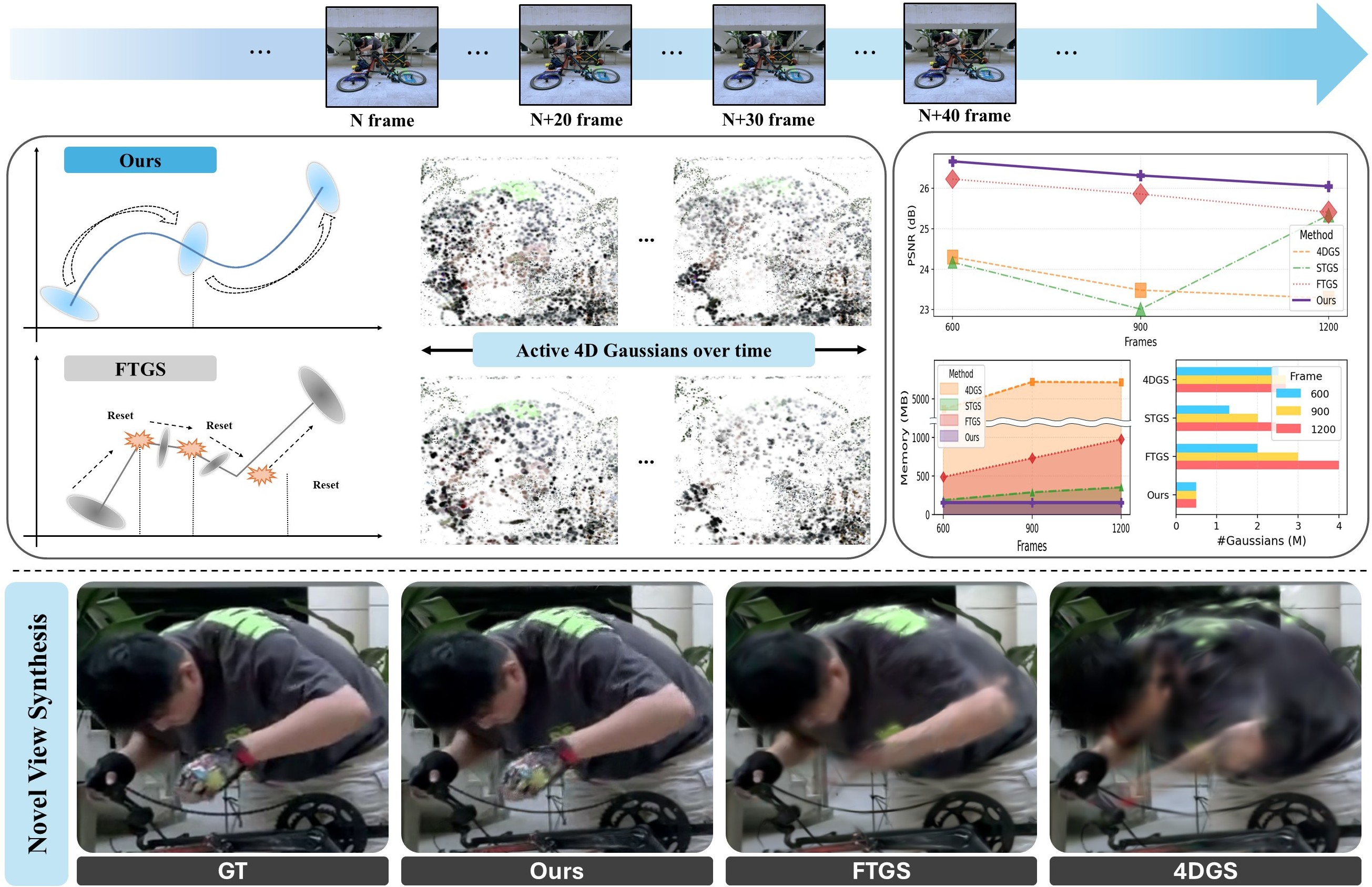}
    \caption{\textbf{Comparison of temporal modeling for extended dynamic scenes.}
    \textbf{(Left)} While previous methods like FTGS rely on fragmented piecewise linear approximations that force primitives to repeatedly reset, our continuous rigid-body transformations preserve the long-term temporal identity of active Gaussians. 
    \textbf{(Right)} By eliminating the need for unnecessary proliferation, TRiGS maintains a strictly constant, compact memory footprint while sustaining high rendering quality across extended sequences (up to 1200 frames), in stark contrast to baselines that suffer from severe memory bloat and performance drops. 
    \textbf{(Bottom)} Consequently, our approach delivers temporally stable, high-fidelity novel view synthesis without visual degradation or motion artifacts.}
    \label{fig:teaser}
\end{figure}
\vspace{-10pt}
\begin{abstract}
Recent 4D Gaussian Splatting (4DGS) methods achieve impressive dynamic scene reconstruction but often rely on piecewise linear velocity approximations and short temporal windows. This disjointed modeling leads to severe temporal fragmentation, forcing primitives to be repeatedly eliminated and regenerated to track complex nonlinear dynamics. This makeshift approximation eliminates the long-term temporal identity of objects and causes an inevitable proliferation of Gaussians, hindering scalability to extended video sequences. To address this, we propose TRiGS, a novel 4D representation that utilizes unified, continuous geometric transformations. By integrating $SE(3)$ transformations, hierarchical Bézier residuals, and learnable local anchors, TRiGS models geometrically consistent rigid motions for individual primitives. This continuous formulation preserves temporal identity and effectively mitigates unbounded memory growth. Extensive experiments demonstrate that TRiGS achieves high fidelity rendering on standard benchmarks while uniquely scaling to extended video sequences (e.g., 600 to 1200 frames) without severe memory bottlenecks, significantly outperforming prior works in temporal stability.
\end{abstract}

\section{Introduction}
Reconstructing dynamic scenes from video is a fundamental problem in computer vision, enabling immersive applications such as virtual reality, augmented reality, and free-viewpoint rendering. Recently, 3D Gaussian Splatting (3DGS) \cite{kerbl20233d, yu2024mip, guedon2024sugar, huang20242d} has advanced static scene rendering by achieving real-time performance while maintaining high visual fidelity. Building upon this success, recent works \cite{yang2024deformable, wu20244d, bae2024per, wu2025localdygs, huang2024sc, yang2023real, li2024spacetime} have extended 3DGS into the temporal domain to model complex 4D dynamic environments, leading to rapid progress in dynamic scene representation and rendering.

One representative approach for 4D dynamic scene reconstruction learns implicit deformation fields \cite{yang2024deformable,wu20244d,bae2024per, shaw2024swings, katsumata2024compact, yan2024street, wu2025localdygs, huang2024sc, chen2025dash} that describe each frame with respect to a canonical space. However, these methods struggle to maintain stable deformations as the temporal distance increases and displacements become large, a common issue in scenes with complex motions. To handle extended sequences, some approaches\cite{wu2025localdygs,li2025gifstream} divide the video into sequential streaming chunks; yet, this strategy incurs severe memory overhead and often leads to visual quality degradation. To overcome the limitations in modeling complex dynamics, recent works\cite{yang2023real,li2024spacetime,lee2024fully, wang2025freetimegs} have shifted toward directly optimizing 4D primitives or parameterizing motion with explicit functions. These explicit approaches currently achieve state-of-the-art performance by granting Gaussians extreme temporal flexibility. For example, they allow primitives to freely appear and disappear over time, or they approximate local motion using piecewise linear velocity and short temporal opacity windows.

Despite achieving high rendering quality on short video clips, these flexible formulations suffer from a fundamental limitation known as temporal fragmentation, as illustrated in Fig.~\ref{fig:teaser}. Since complex non-linear motion is constrained to linear approximations or short temporal windows, Gaussian primitives cannot accurately track continuous dynamic changes. Because a linear assumption inherently diverges from a true curved trajectory over time, the primitives gradually drift away from the actual geometry. To compensate for this growing spatial mismatch, the model is driven to repeatedly eliminate, split, and regenerate primitives. As visually evident in the left panel of Fig.~\ref{fig:teaser}, this disjointed modeling leads to an inevitable proliferation of Gaussians. This makeshift not only degrades the object’s long-term temporal identity but also incurs substantial memory overhead. Consequently, such fragmented representations are difficult to scale to extended video sequences beyond hundreds of frames under practical memory constraints.

To address these limitations, we propose TRiGS, a new dynamic rendering framework that integrates continuous geometric transformations into 4D Gaussian Splatting to model temporally robust rigid motions. TRiGS moves beyond fragmented formulations that approximate continuous nonlinear motion using discontinuous linear segments or impose short lifetimes on primitives. Instead, it directly optimizes the motion so that the Gaussians composing an object undergo a geometrically consistent, continuous rigid transformation over time.

Concretely, we achieve this continuous representation through three key components. First, we employ the exponential map of $SE(3)$ to unify complex rotation and translation into a single continuous rigid transformation. Second, to accurately capture nonlinear dynamics over time, we formulate hierarchical motion decomposition where continuous time varying offsets, parameterized by a quadratic Bézier curve, smoothly refine the base transformation parameters. Finally, to faithfully capture independent local motions rather than relying on a shared global center of rotation, we introduce learnable local anchors. Each Gaussian is associated with its own anchor that acts as an independent center of rotation, enabling the stable optimization of geometrically consistent and fine grained local rigid motions.

As a result, unlike piecewise linear modeling which easily loses track of complex dynamics, our continuous motion and anchor modeling consistently preserves the long-term temporal identity of Gaussians. It also reduces the need to compensate for motion approximation errors by repeatedly splitting and rearranging primitives, thereby suppressing unnecessary Gaussian proliferation and excessive re-initialization. Overall, TRiGS prevents temporal fragmentation while lowering memory usage, making dynamic scene reconstruction feasible for long sequences beyond hundreds of frames without severe memory bottlenecks.

 In summary, our main contributions are:
\begin{itemize}
\item We propose TRiGS, a novel 4D representation that utilizes unified, continuous geometric transformations rather than fragmented, piecewise linear velocity approximations. By integrating $SE(3)$ representations with learnable local anchors, it models geometrically consistent rigid motions, effectively overcoming the severe temporal fragmentation seen in existing methods.
\item We introduce a robust dynamic rendering pipeline designed to preserve the long-term temporal identity of primitives. Rather than repeatedly eliminating and regenerating Gaussians to handle complex nonlinear dynamics, we capture these continuous changes using time varying offsets. Coupled with a motion guided relocation strategy, it actively recycles redundant Gaussians, effectively mitigating unbounded memory growth.
\item We demonstrate through extensive experiments that TRiGS achieves high fidelity rendering on standard dynamic benchmarks while uniquely scaling to extended video sequences (e.g., 600 to 1200 frames). By maintaining a compact representation without severe memory bottlenecks, it significantly outperforms prior works in temporal stability and scalability.
\end{itemize}

\label{sec:intro}

\vspace{-5pt}
\section{Related Work}
\label{sec:related}

\subsection{Implicit Deformation for Dynamic 3D Gaussians}
Following 3D Gaussian Splatting\cite{kerbl20233d}, recent efforts have extended it to dynamic scenes. Representative approaches\cite{yang2024deformable, wu20244d, zhu2024motiongs,lu20243d, chen2025dash, yoon2025splinegs, kratimenos2024dynmf} involve defining a canonical space and implicitly predicting the deformations of Gaussians over time using Neural Fields. Subsequent studies have improved performance by dividing the scene into multiple local spaces\cite{wu2025localdygs, li2025gifstream, fu2025recon} or guiding the deformation based on sparse control points\cite{huang2024sc, lu2024scaffold, li2024dreammesh4d}. However, these implicit deformation field based strategies often struggle to maintain structural rigidity over time. Consequently, they demand careful tuning of initialization and regularizers to prevent drift. In long and complex scenes, errors accumulate, causing the model to lose geometric consistency and redundantly proliferate Gaussians to mask tearing and distortion artifacts.

\subsection{4D Attributes and Explicit Motion Modeling}
Parallel to these implicit approaches, 4D attributes\cite{yang2023real, li2024spacetime, gao20246dgs, gao20257dgs} were introduced to ensure spatiotemporal consistency by adding a temporal dimension to Gaussians. Subsequent studies utilized explicit functions\cite{lee2024fully, luiten2024dynamic, park2025splinegs, jin2025moe, wang2025shape} to optimize spatiotemporal displacements. Recently, methods\cite{katsumata2024compact, wang2025freetimegs} with compact velocity equations have been proposed, achieving high performance. Despite these advancements, existing explicit modeling approaches face a shared challenge: they optimize transformation parameters independently and struggle to correctly track the natural, curved motions of actual rigid bodies. Ultimately, these models patch the gaps by excessively duplicating Gaussians to cover rendering errors.

In this work, we argue that structurally coupled motion modeling is a principled approach for the 4DGS framework. Thus, departing from independent parameter optimization, we introduce the geometric coupling of parameters and a representation based on the Gaussian's inherent local coordinate system. By fundamentally preventing unnatural shape distortions and performing precise motion estimation, the proposed model effectively represents long dynamic scenes with only a small number of Gaussians.

\begin{figure*}[t]
\centerline{\includegraphics[width=\textwidth]{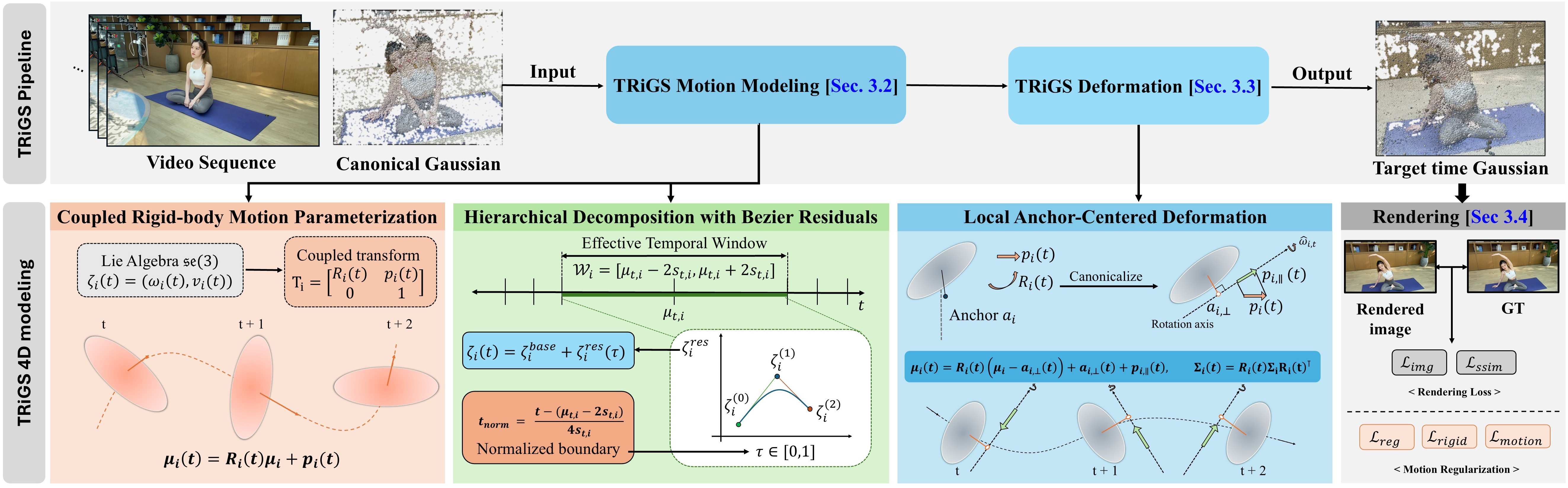}}
    \caption{\textbf{Overview of our framework.} 
    We comprehensively model the scene dynamics through three key components. First, the motion is parameterized via a hierarchical decomposition using Bézier residuals within an effective temporal window. Second, this Lie algebra representation is mapped to a coupled $SE(3)$ transformation. Third, the transformation is applied relative to a gauge-fixed local anchor ($a_{i,\perp}$) to ensure stable, articulate deformation. Finally, the deformed primitives are rendered to optimize both photometric and motion regularization objectives.
    }
    \label{figure_intro_static_dynamic}
\end{figure*}

\section{Method}
\label{sec:method}
We consider a calibrated video capturing a dynamic 3D scene. Our goal is to reconstruct a time-varying scene representation that supports novel-view rendering. To this end, we propose TRiGS, which comprehensively models complex scene dynamics through three key components: Coupled $SE(3)$ transformation for robust rigid-body motion modeling, Hierarchical Decomposition with B\'ezier Residuals, and a Local Anchor-Centered Deformation to capture local motion.
\subsection{Preliminaries}
We represent the scene as a set of 3D Gaussian primitives $\{g_i\}$.
Each primitive $g_i$ has a canonical mean $\mu_i\in\mathbb{R}^3$, covariance $\Sigma_i\in\mathbb{R}^{3\times 3}$, and canonical opacity $\alpha_i$.
Time $t$ is treated as a continuous scalar, and each primitive is associated with a central time $\mu_{t,i}$ .
\subsubsection{Temporal Opacity}
Temporal opacity is used to restrict the visibility of each primitive to a local neighborhood in time.
We introduce a temporal visibility function $\gamma_i:\mathbb{R}\to(0,1]$, parameterized by a temporal scale $s_{t,i}>0$, to define the time-dependent opacity as:
\begin{equation} 
\gamma_i(t) \;=\; \exp\!\Big(-\frac{(t-\mu_{t,i})^2}{2s_{t,i}^2}\Big), 
\qquad 
\alpha_i(t) \;=\; \alpha_i \cdot \gamma_i(t).
\label{eq:temporal_opacity}
\end{equation}
At query time $t$, the rendering process utilizes the time-dependent parameters $\mu_i(t), \Sigma_i(t), \alpha_i(t)$, where $\mu_i(t)$ and $\Sigma_i(t)$ are governed by the motion model.
\subsubsection{Linear Motion Modeling}
Like FreeTimeGS (FTGS) \cite{wang2025freetimegs}, a simple baseline models the primitive center using translation-only linear motion:
\begin{equation}
\mu_i(t) \;=\; \mu_i + v_i\,(t-\mu_{t,i}),
\label{eq:linear_motion}
\end{equation}
where $v_i\in\mathbb{R}^3$ is a per-primitive velocity.
To represent complex motions, a translation-only model often needs to decompose the trajectory with piecewise-linear representation, which typically requires many primitives as the motion becomes more complex. Under this formulation, the more complex the motion becomes, the more primitives are needed to maintain fidelity, inevitably causing severe temporal fragmentation and a significant surge in memory usage.
\subsection{Motion Modeling}
\subsubsection{Coupled Rigid-body Motion Parameterization}
To avoid allocating many time-localized primitives for complex motion, we model the motion of each Gaussian primitive with an $SE(3)$ formulation:
\begin{equation}
\mu_i(t) \;=\; R_i(t)\,\mu_i \;+\; p_i(t),
\label{eq:se3_motion_simple}
\end{equation}
where $R_i(t)\in SO(3)$ and $p_i(t)\in\mathbb{R}^3$.
However, optimizing $R_i(t)$ and $p_i(t)$ as independent parameters is often ill-conditioned, because rotation and translation can partially compensate each other under the rendering objective. This creates a continuum of near-equivalent solutions, yielding an entangled motion decomposition and accumulated drift over time. We therefore parameterize motion directly in the Lie algebra $\mathfrak{se}(3)$ and map it to $SE(3)$ via the exponential map, which enforces a coupled rigid transform and helps stabilize optimization without introducing additional disentanglement-specific losses. Concretely, we optimize a time-conditioned coefficient $\zeta_i(t)=(\omega_i(t), \nu_i(t))\in\mathfrak{se}(3)$ and define the relative log-parameter $\mathbf{u}_i(t):=\zeta_i(t)\Delta t$ with $\Delta t:=t-\mu_{t,i}$. The relative transform from the central time $\mu_{t,i}$ to $t$ is then
\begin{equation}
\mathbf{T}_i(\mu_{t,i}\!\to\! t)=\exp\!\big(\hat{\mathbf{u}}_i(t)\big)=
\begin{bmatrix}
R_i(t) & p_i(t)\\
0 & 1
\end{bmatrix},
\qquad \mathbf{u}_i(\mu_{t,i})=\mathbf{0}.
\label{eq:expmap_main}
\end{equation}
where $\hat{\mathbf{u}}_i(t)$ denotes the standard wedge operator in $\mathfrak{se}(3)$. This construction satisfies $\mathbf{T}_i(\mu_{t,i})=\mathbf{I}$ by design. Moreover, $R_i(t)$ and $p_i(t)$ are jointly determined by the same closed-form exponential map.
Let $\mathbf{u}_i(t)=(\phi_i(t),\upsilon_i(t))$ with $\phi_i(t)=\omega_i(t)\Delta t$ and $\upsilon_i(t)=\nu_i(t)\Delta t$.
We compute $R_i(t)$ via Rodrigues' formula and obtain $p_i(t)=J(\phi_i(t))\,\upsilon_i(t)$ using the $SO(3)$ left Jacobian $J(\cdot)$, rather than optimizing translation independently.
\subsubsection{Hierarchical Decomposition with B\'ezier Residuals}
To capture complex, non-linear motion over time, we model the time dependence of the Lie algebra coefficient $\zeta_i(t)$ using a hierarchical decomposition into a base term and a Bézier residual. Specifically, we decompose it into learnable base terms $\zeta_i^{\mathrm{base}} = (\omega_i^{\mathrm{base}},\nu_i^{\mathrm{base}})$ and time-varying residuals $\zeta_i^{\mathrm{res}} = (\omega_i^{\mathrm{res}}(\tau),\nu_i^{\mathrm{res}}(\tau))$ represented by a quadratic B\'ezier curve. To concentrate modeling capacity where primitive $i$ is effectively supervised, we localize the parameterization to an effective temporal window implied by temporal visibility $\gamma_i(t)$. Since $\gamma_i(t)$ rapidly decays away from $\mu_{t,i}$, primitive $i$ receives almost no gradients at distant times. Restricting the domain stabilizes optimization and avoids drift in weakly supervised regions. We define the effective window as
\begin{equation}
\mathcal{W}_i \;=\; [\,\mu_{t,i}-2s_{t,i},\ \mu_{t,i}+2s_{t,i}\,].
\label{eq:temporal_window}
\end{equation}
To evaluate the B\'ezier residual within $ \mathcal{W}_i $, we map the continuous global time $ t $ to a local normalized coordinate $ \tau \in [0,1] $. We compute the relative time progress $ t_{\mathrm{norm}} $ and clamp it to ensure the temporal offset remains bounded outside this active temporal extent:
\begin{equation}
t_{\mathrm{norm}} \;=\; \frac{t-(\mu_{t,i}-2s_{t,i})}{4s_{t,i}},\qquad \tau \;=\; 
\begin{cases} 
0, & \text{if } t_{\mathrm{norm}} < 0 \\[1ex]
t_{\mathrm{norm}}, & \text{if } 0 \le t_{\mathrm{norm}} \le 1 \\[1ex]
1, & \text{if } t_{\mathrm{norm}} > 1
\end{cases}
\label{eq:tau}
\end{equation}
We model the residual $\zeta_i^{\mathrm{res}}(\tau)$ as a quadratic B\'ezier curve with learnable coefficients, providing a smooth fine-grained refinement on top of the base motion. The final motion coefficient is obtained by adding the residual to the base:
\begin{gather}
\zeta_i^{\mathrm{res}}(\tau) = (1-\tau)^2 \zeta_i^{(0)} + 2(1-\tau)\tau \zeta_i^{(1)} + \tau^2 \zeta_i^{(2)}, \label{eq:bezier_v} \\
\zeta_i(t) = \zeta_i^{\mathrm{base}} + \zeta_i^{\mathrm{res}}(\tau).
\label{eq:twist_final}
\end{gather}
where $\zeta_i^{\mathrm{base}}, \zeta_i^{(k)}\in\mathfrak{se}(3)$ for $k\in\{0,1,2\}$ are learnable parameters.
\subsection{Local Anchor-Centered Deformation}
Applying an $SE(3)$ transform in Eq.~\ref{eq:se3_motion_simple} with a single shared anchor implicitly assumes that all primitives rotate around the same center. Dynamic scenes contain multiple parts undergoing independent local motions with distinct centers of rotation. To faithfully capture these independent motions, we introduce a learnable per-primitive local anchor $a_i\in\mathbb{R}^3$ and parameterize the $SE(3)$ transform around it. 

However, $a_i$ is not uniquely identifiable from translation. The same deformation can be explained by multiple $(a_i, p_i(t))$ pairs leading to optimization ambiguity. To alleviate this ambiguity, we canonicalize the $a_i$ as $a_{i,\perp}(t)$ by projecting $a_i$ onto the plane orthogonal to the rotation axis, and use only the axis-aligned translational component $\upsilon_{i,\parallel}(t)$:
\begin{equation}
a_{i,\perp}(t)=a_i-\langle a_i,\bar{\omega}_i(t)\rangle \bar{\omega}_i(t),
\qquad 
\upsilon_{i,\parallel}(t)=\langle \upsilon_i(t),\bar{\omega}_i(t)\rangle\,\bar{\omega}_i(t),
\end{equation}
where $\bar{\omega}_i(t)$ denotes the unit rotation axis direction. The anchor projection $a_{i,\perp}(t)$ acts as a gauge fixing, since $(I-R_i(t))\bar{\omega}_i(t)=0$ implies that the axis-parallel component of $a_i$ lies in the null space of $(I-R_i(t))$ and is thus unidentifiable. Consequently, we keep only the axis-aligned translation induced by the exponential map and denote it as $p_{i,\parallel}(t)=J(\phi_i(t))\,\upsilon_{i,\parallel}(t)$. With this local anchor parameterization, the deformed Gaussian primitives at time $t$ are given by
\begin{equation}
\mu_i(t) = R_i(t)\,(\mu_i - a_{i,\perp}(t)) + a_{i,\perp}(t) + p_{i,\parallel}(t), \qquad
\Sigma_i(t) = R_i(t)\,\Sigma_i\,R_i(t)^\top.
\label{eq:gaussian_deform_params}
\end{equation}
Please see the supplementary material for the closed-form $SE(3)$ exponential map, the gauge-fixing derivation, and small-angle numerical stabilization.
\subsection{Training}
Similar to 3DGS, we optimize Gaussian primitive parameters by minimizing a rendering objective between the rendered images and the ground-truth images:
\begin{equation}
\mathcal{L}
=
\lambda_{img}\,\mathcal{L}_{img}
+
\lambda_{ssim}\,\mathcal{L}_{ssim}
+
\lambda_{reg}\, \mathcal{L}_{reg}
+
\lambda_{motion}\,\mathcal{L}_{motion}
+
\lambda_{rigid}\,\mathcal{L}_{rigid}.
\label{eq:total_loss}
\end{equation}
\subsubsection{Motion regularization}
Although our primary supervision is the photometric rendering loss, the optimization remains ill-posed due to the large number of primitive parameters and the ambiguity of explaining image residuals, especially in dynamic scenes. Therefore, we regularize primitive parameters with $\mathcal{L}_{reg}$ following FTGS, enforce temporal smoothness with $\mathcal{L}_{motion}$, and encourage local motion coherence with $\mathcal{L}_{rigid}$. We denote the stop-gradient operator by $sg[\cdot]$.
\begin{gather}
b_i=\zeta_i^{(0)}-2\zeta_i^{(1)}+\zeta_i^{(2)},\qquad
\mathcal{L}_{motion}=\|b_i\|_2^2 .
\end{gather}
$\mathcal{L}_{motion}$ regularizes the quadratic B\'ezier residual by penalizing its acceleration term in both translation and rotation components. This discourages abrupt curvature in $\zeta_i(t)$ and promotes temporally smooth trajectories within the effective window.
\begin{gather}
K_{ij}=\exp\!\left(-\lambda_c\,\left\|\mathbf{c}_i-\mathbf{c}_j\right\|_2^2\right),\\
\mathcal{L}_{rigid}=\sum_i \sum_{j\in \mathcal{N}(i)}
K_{ij}\left(\left\|\nu_i^{\mathrm{base}}-\nu_j^{\mathrm{base}}\right\|_2^2 + \left\|\omega_i^{\mathrm{base}}-\omega_j^{\mathrm{base}}\right\|_2^2 \right),
\end{gather}
where $\mathbf{c}_i$ denotes the DC spherical-harmonics color of primitive $i$ in canonical space.
$\mathcal{L}_{rigid}$ promotes locally coherent motion by aligning the base motions of spatial neighbors. The appearance-based weight $K_{ij}$ serves as a soft affinity that suppresses incorrect coupling across object boundaries. Concretely, we compute the penalty over $k$-nearest-neighbor Gaussian primitives $j\in\mathcal{N}(i)$ based on their canonical centers $\mu_i$ in 3D. This encourages smooth motion within each region while preserving motion discontinuities at boundaries, reducing local drift and motion fragmentation. 
\subsubsection{Motion-guided Relocation}
To reuse capacity under a fixed primitive budget, we periodically recycle low-opacity primitives by cloning informative active ones. 
We split primitives using an opacity threshold $\tau_\alpha$:
\begin{equation}
\mathcal{I}=\{i~|~\sigma(\alpha_i)<\tau_\alpha\},\qquad
\mathcal{A}=\{i~|~\sigma(\alpha_i)\ge\tau_\alpha\}.
\end{equation}
We then sample a source primitive $i\in\mathcal{A}$ with probability $q_i \propto s_i$, where
\begin{equation}
s_i=\sum_{u\in\{\alpha,2D,t,d\nu\}} p_i^{(u)},\qquad
q_i=\frac{s_i}{\sum_{k\in\mathcal{A}} s_k}.
\end{equation}
Here $\{p_i^{(u)}\}$ are normalized difficulty cues defined in the supplement. The sampled primitive is cloned to replace an element in $\mathcal{I}$, reallocating primitives to hard regions while keeping the total count unchanged.

\section{Experiments}
\begin{table*}[t]
  \centering
  \caption{
  Quantitative comparison on the SelfCap dataset across 600, 900, and 1200 frame budgets.
  \colorbox{best}{best}, \colorbox{second}{second-best}, and \colorbox{third}{third-best} results are highlighted in the table.
   ($^{\ast}$ Results from our reproduction, as the official code is not publicly available.)
  }
  \label{tab:selfcap_quality}
  \tiny
  \setlength{\tabcolsep}{4pt}
  \renewcommand{\arraystretch}{1.0}
  \resizebox{\textwidth}{!}{%
  \begin{tabular}{l|ccc|ccc|ccc}
    \toprule
    \multirow{2}{*}[-0.4em]{Method} & \multicolumn{3}{c|}{600 frames} & \multicolumn{3}{c|}{900 frames} & \multicolumn{3}{c}{1200 frames} \\
    \cmidrule(lr){2-4} \cmidrule(lr){5-7} \cmidrule(lr){8-10}
    & PSNR$\uparrow$ & SSIM$\uparrow$ & LPIPS$\downarrow$
    & PSNR$\uparrow$ & SSIM$\uparrow$ & LPIPS$\downarrow$
    & PSNR$\uparrow$ & SSIM$\uparrow$ & LPIPS$\downarrow$ \\
    \midrule

    GIFStream\cite{li2025gifstream}
      & 23.76
      & \cellcolor{third}0.877
      & \cellcolor{third}0.134
      & \cellcolor{third}23.80
      & \cellcolor{third}0.871
      & \cellcolor{second}0.124
      & 24.27
      & \cellcolor{third}0.874
      & 0.117 \\

    4DGS\cite{yang2023real}
      & 24.30
      & 0.860
      & 0.227
      & 23.48
      & 0.851
      & 0.240
      & 23.28
      & 0.851
      & 0.233 \\

    Ex-4DGS\cite{lee2024fully}
      & 21.61
      & 0.801
      & 0.233
      & 20.93
      & 0.793
      & 0.248
      & 20.45
      & 0.787
      & 0.248 \\

    STGS\cite{li2024spacetime}
      & 24.17
      & 0.844
      & 0.161
      & 23.01
      & 0.864
      & 0.136
      & \cellcolor{third}25.34
      & 0.734
      & \cellcolor{second}0.100 \\

    LocalDyGS\cite{wu2025localdygs}
      & \cellcolor{third}25.49
      & 0.858
      & 0.240
      & 23.51
      & 0.838
      & 0.280
      & 24.24
      & \cellcolor{second}0.884
      & \cellcolor{third}0.116 \\

    FTGS$^{\ast}$\cite{wang2025freetimegs}
      & \cellcolor{second}26.23
      & \cellcolor{second}0.913
      & \cellcolor{second}0.121
      & \cellcolor{second}25.86
      & \cellcolor{second}0.898
      & \cellcolor{third}0.132
      & \cellcolor{second}25.41
      & 0.871
      & 0.146 \\

    \textbf{Ours}
      & \cellcolor{best}\textbf{26.67}
      & \cellcolor{best}\textbf{0.937}
      & \cellcolor{best}\textbf{0.113}
      & \cellcolor{best}\textbf{26.32}
      & \cellcolor{best}\textbf{0.916}
      & \cellcolor{best}\textbf{0.120}
      & \cellcolor{best}\textbf{26.05}
      & \cellcolor{best}\textbf{0.904}
      & \cellcolor{best}\textbf{0.099} \\

    \bottomrule
  \end{tabular}%
  }
\end{table*}

\begin{figure*}[!t]
\centering
\small

\begin{adjustbox}{valign=c} 
    \includegraphics[width=0.24\textwidth]{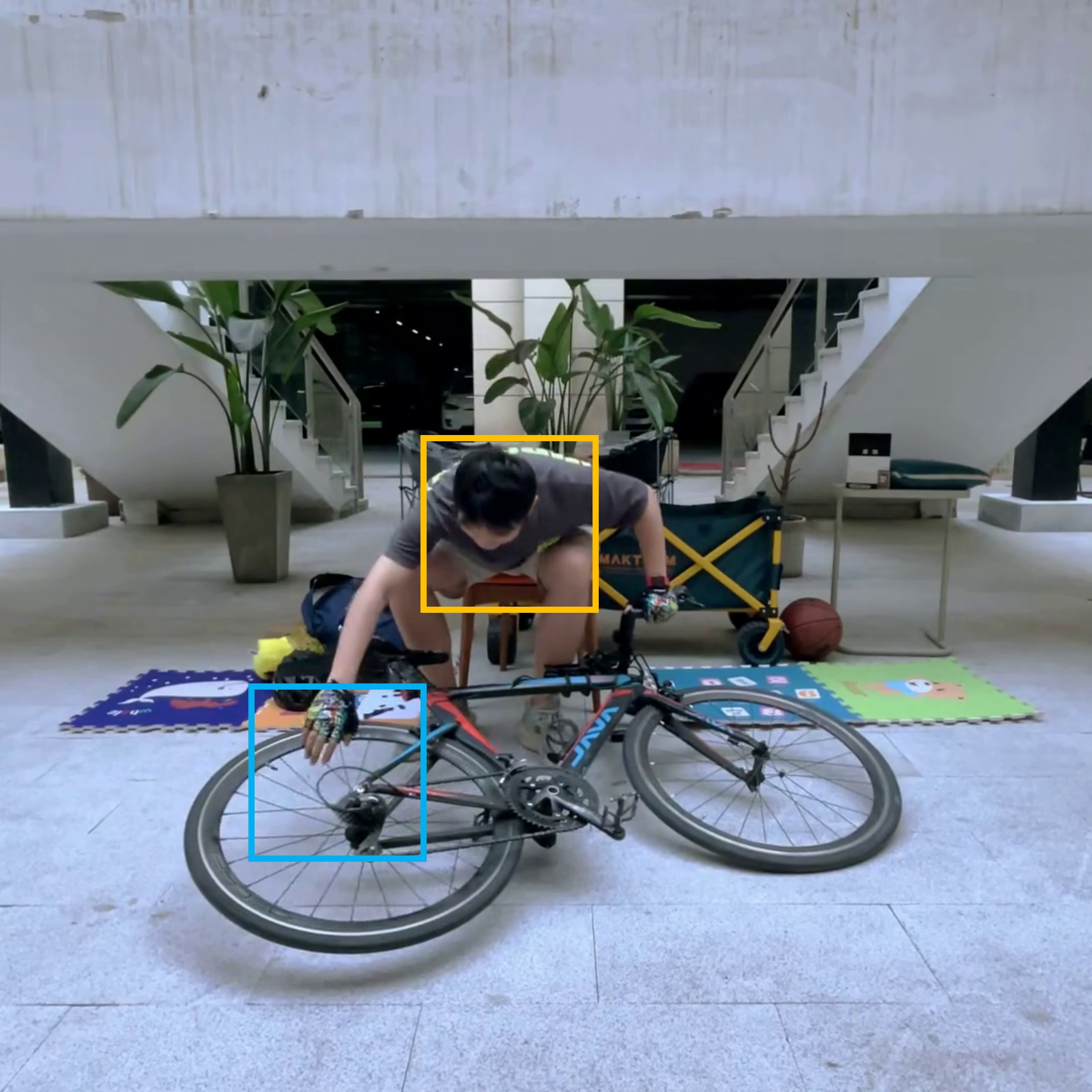}
\end{adjustbox}
\hfill
\begin{adjustbox}{valign=c} 
\begin{minipage}[c]{0.74\textwidth} 
    \centering
    \includegraphics[width=0.19\textwidth]{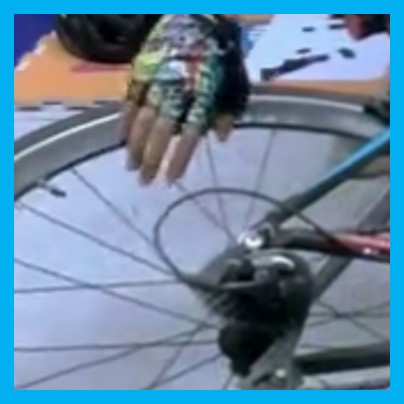}\hfill
    \includegraphics[width=0.19\textwidth]{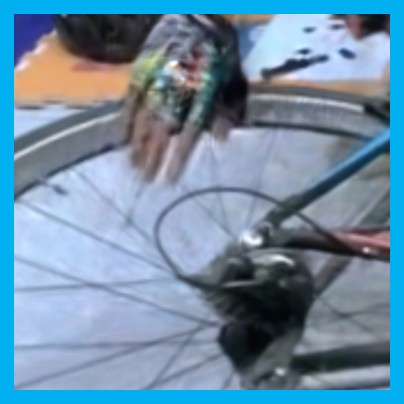}\hfill
    \includegraphics[width=0.19\textwidth]{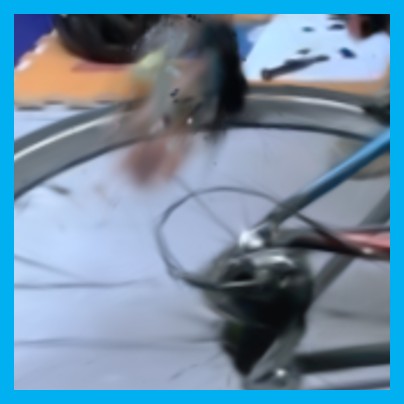}\hfill
    \includegraphics[width=0.19\textwidth]{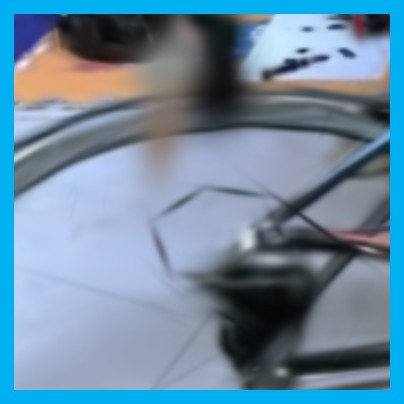}\hfill
    \includegraphics[width=0.19\textwidth]{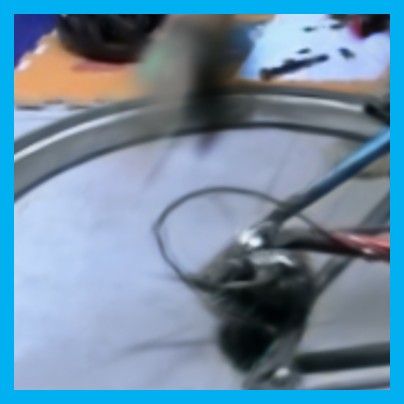}\\[0.6mm]
    \includegraphics[width=0.19\textwidth]{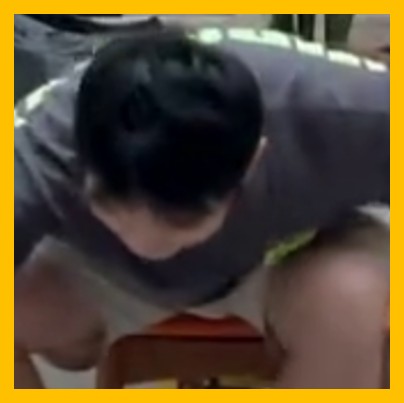}\hfill
    \includegraphics[width=0.19\textwidth]{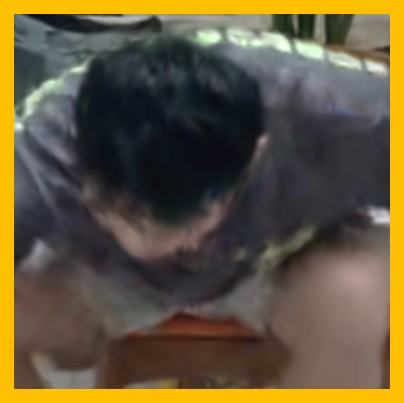}\hfill
    \includegraphics[width=0.19\textwidth]{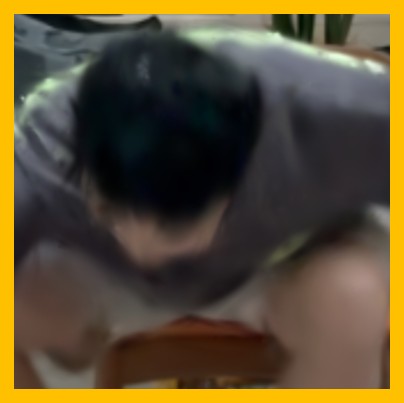}\hfill
    \includegraphics[width=0.19\textwidth]{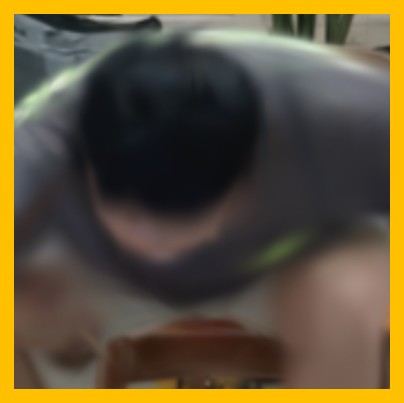}\hfill
    \includegraphics[width=0.19\textwidth]{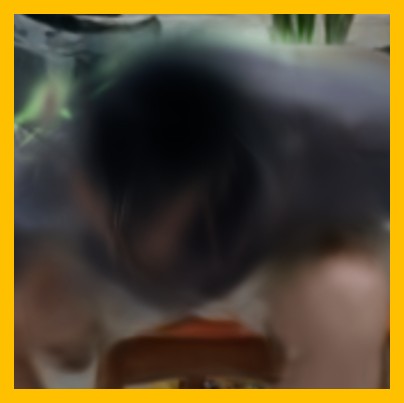}
\end{minipage}
\end{adjustbox}

\vspace{3mm} 

\begin{adjustbox}{valign=c}
    \includegraphics[width=0.24\textwidth]{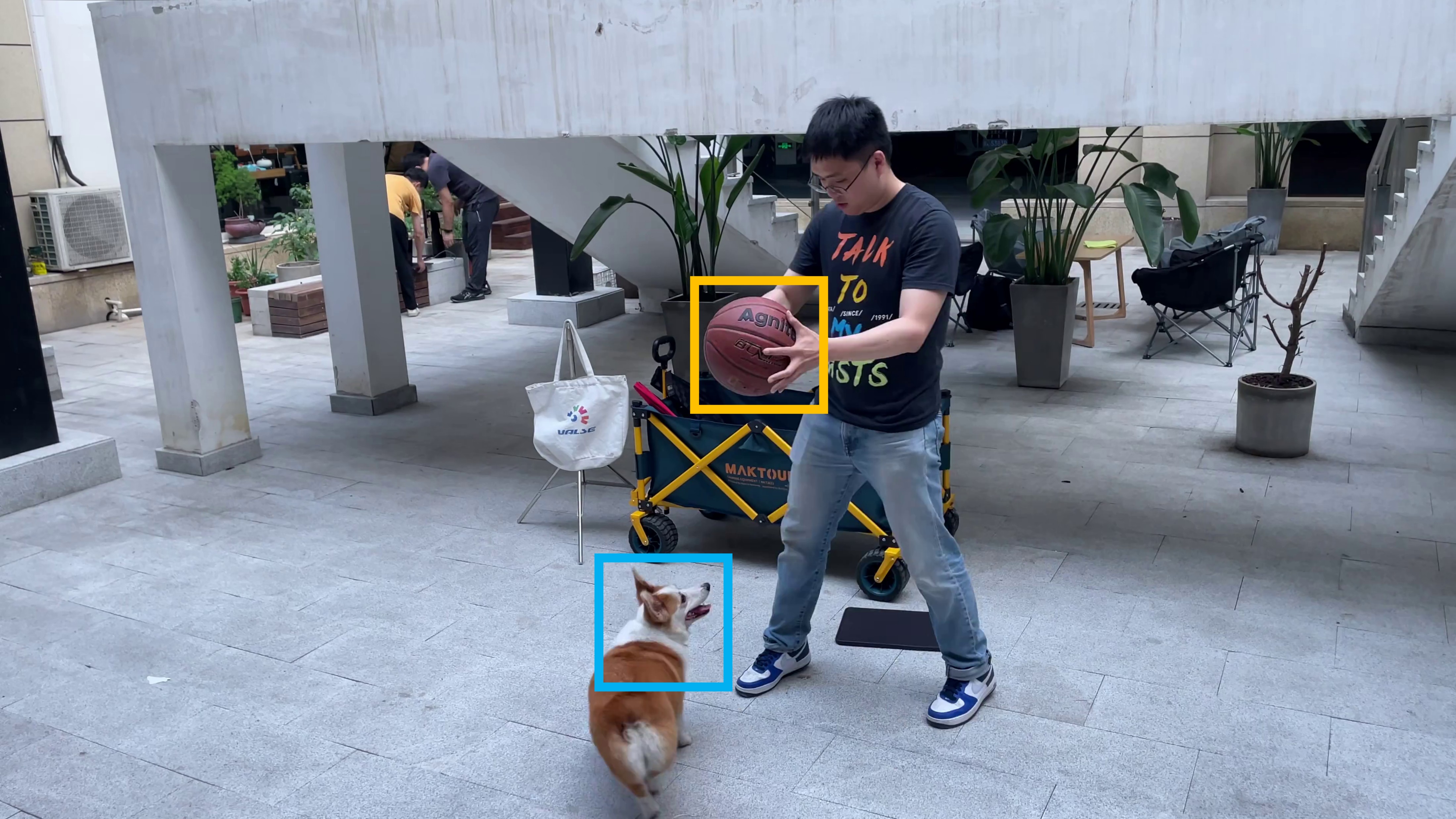}
\end{adjustbox}
\hfill
\begin{adjustbox}{valign=c}
\begin{minipage}[c]{0.74\textwidth}
    \centering
    \includegraphics[width=0.19\textwidth]{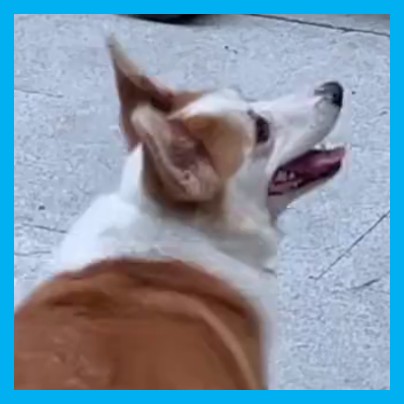}\hfill
    \includegraphics[width=0.19\textwidth]{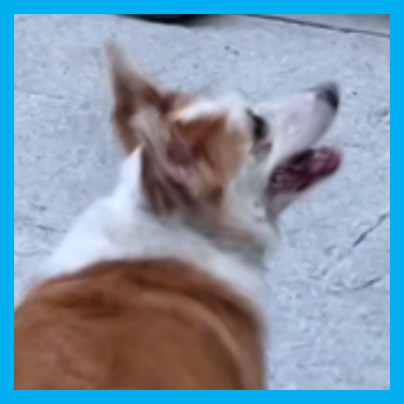}\hfill
    \includegraphics[width=0.19\textwidth]{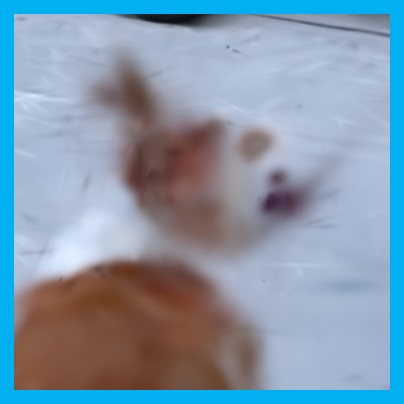}\hfill
    \includegraphics[width=0.19\textwidth]{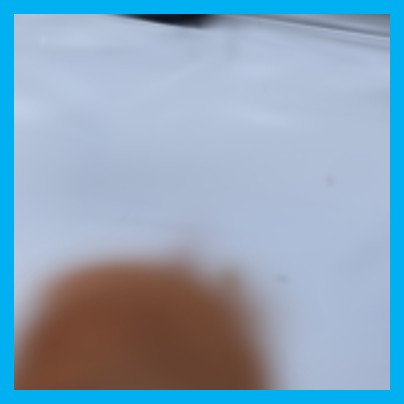}\hfill
    \includegraphics[width=0.19\textwidth]{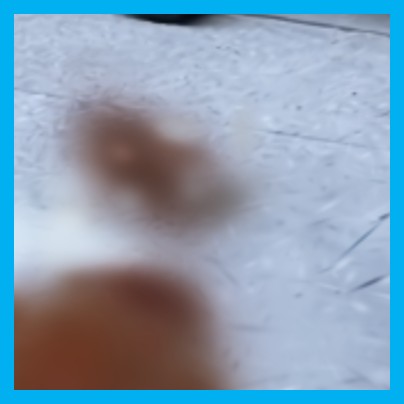}\\[0.6mm]
    \includegraphics[width=0.19\textwidth]{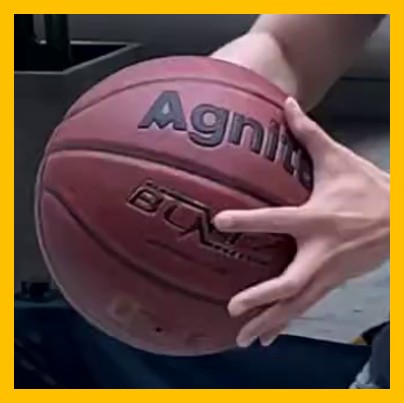}\hfill
    \includegraphics[width=0.19\textwidth]{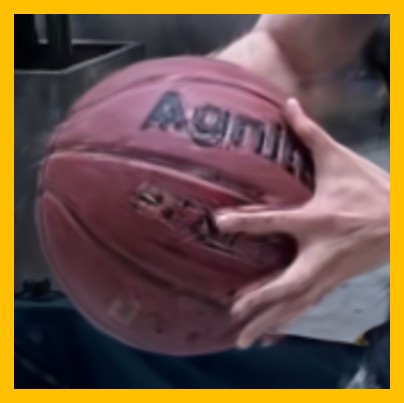}\hfill
    \includegraphics[width=0.19\textwidth]{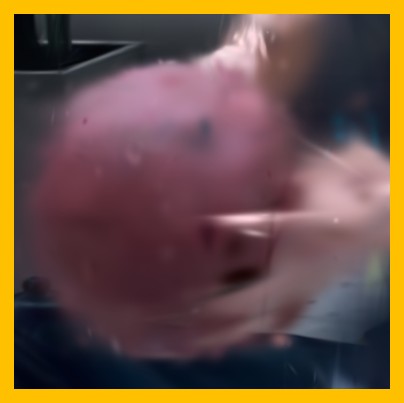}\hfill
    \includegraphics[width=0.19\textwidth]{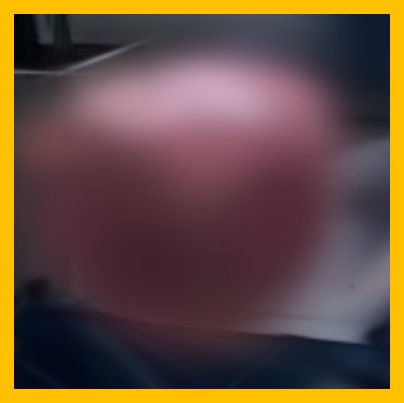}\hfill
    \includegraphics[width=0.19\textwidth]{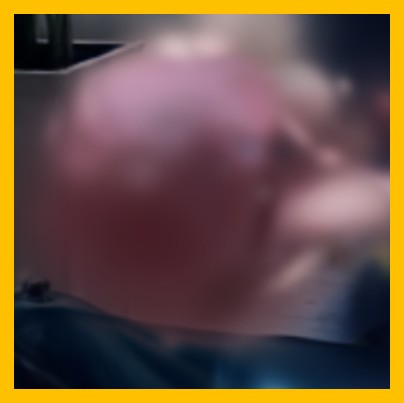}
\end{minipage}
\end{adjustbox}

\vspace{3mm}

\begin{adjustbox}{valign=c}
    \includegraphics[width=0.24\textwidth]{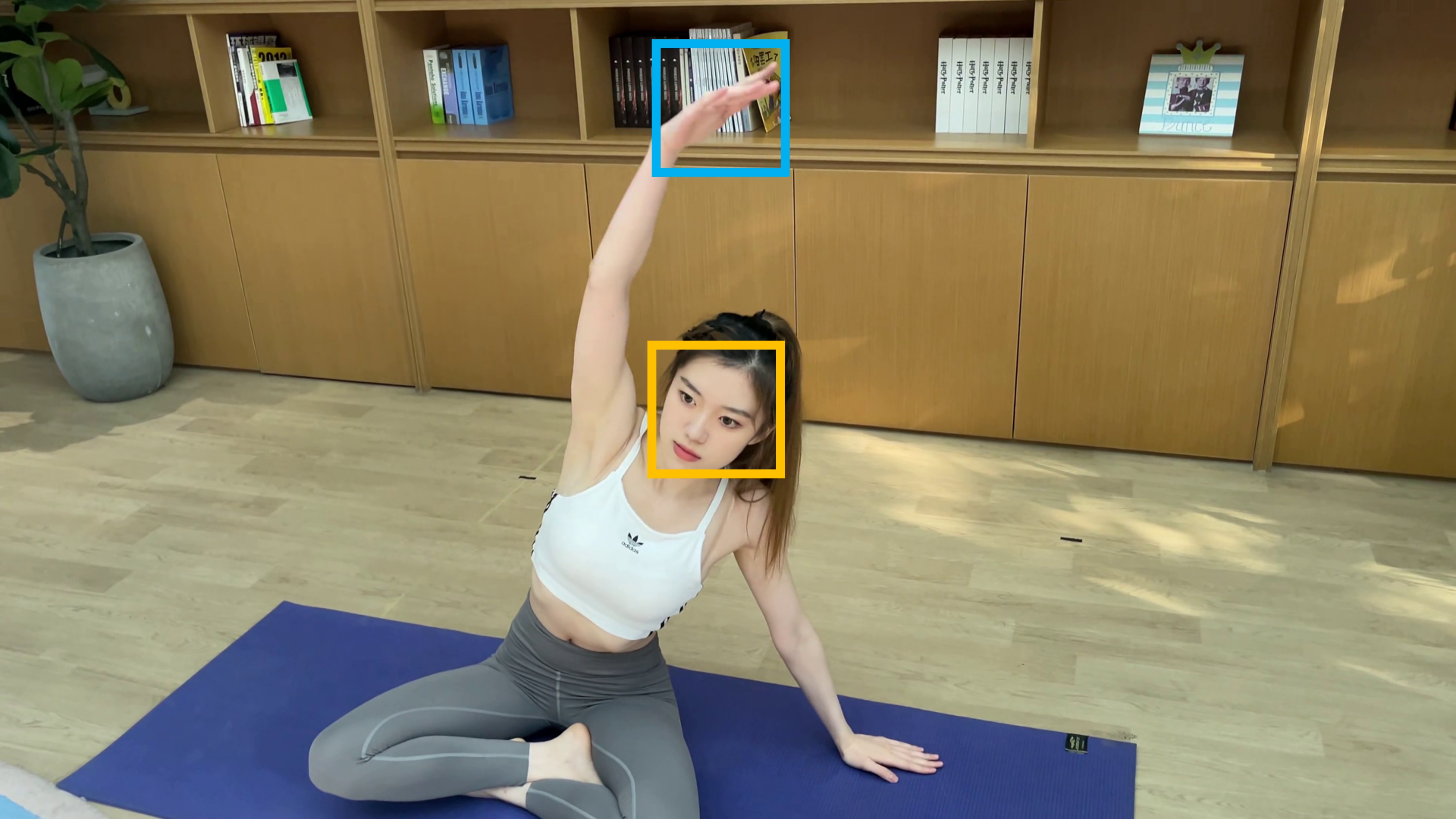}
\end{adjustbox}
\hfill
\begin{adjustbox}{valign=c}
\begin{minipage}[c]{0.74\textwidth}
    \centering
    \includegraphics[width=0.19\textwidth]{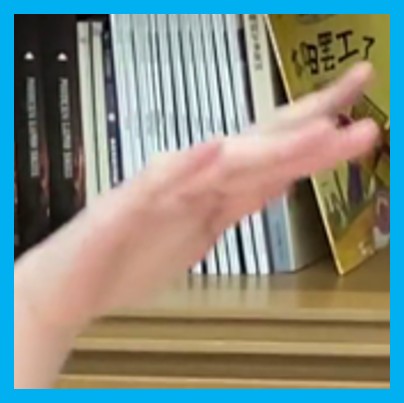}\hfill
    \includegraphics[width=0.19\textwidth]{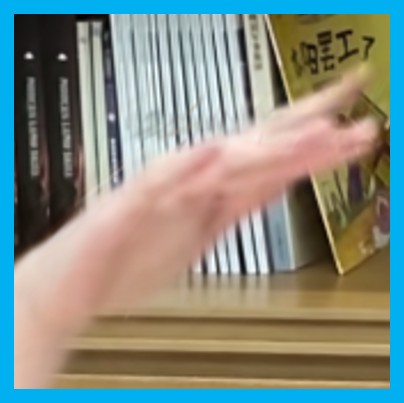}\hfill
    \includegraphics[width=0.19\textwidth]{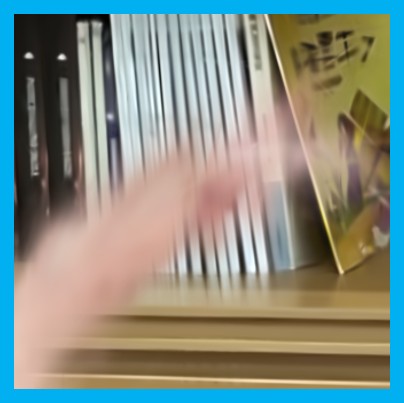}\hfill
    \includegraphics[width=0.19\textwidth]{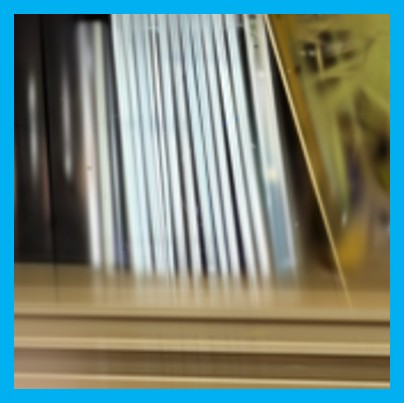}\hfill
    \includegraphics[width=0.19\textwidth]{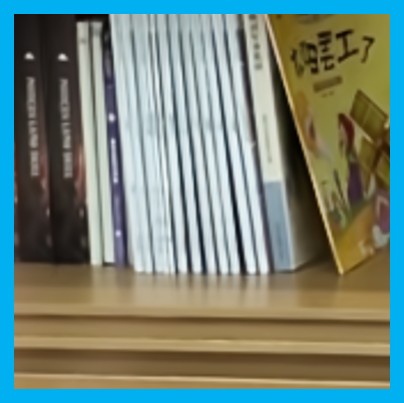}\\[0.6mm]
    \includegraphics[width=0.19\textwidth]{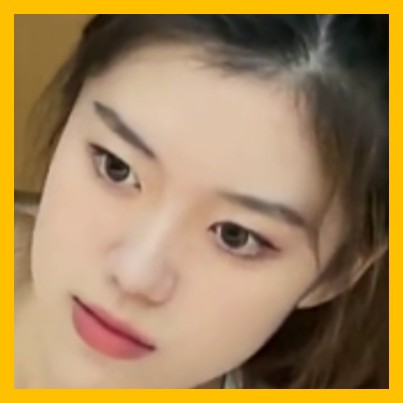}\hfill
    \includegraphics[width=0.19\textwidth]{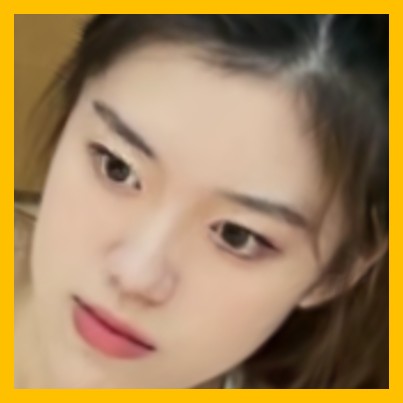}\hfill
    \includegraphics[width=0.19\textwidth]{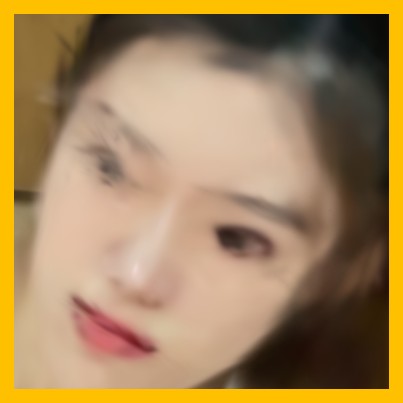}\hfill
    \includegraphics[width=0.19\textwidth]{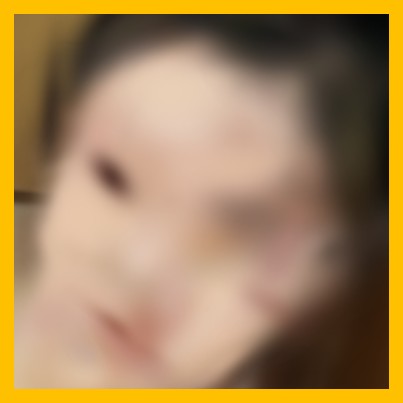}\hfill
    \includegraphics[width=0.19\textwidth]{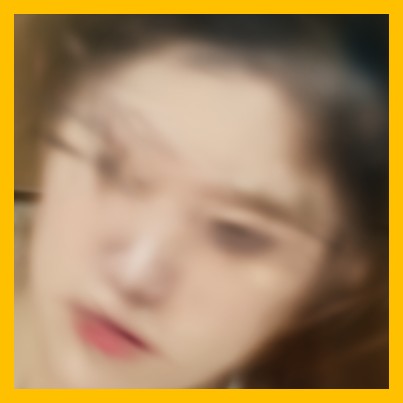}
\end{minipage}
\end{adjustbox}

\vspace{1mm}

\makebox[0.26\textwidth]{\scriptsize Full Image} \hfill 
\makebox[0.138\textwidth]{\scriptsize GT} \hfill
\makebox[0.138\textwidth]{\scriptsize \textbf{Ours}} \hfill
\makebox[0.138\textwidth]{\scriptsize FTGS \cite{wang2025freetimegs}} \hfill
\makebox[0.138\textwidth]{\scriptsize LocalDyGS \cite{wu2025localdygs}} \hfill
\makebox[0.138\textwidth]{\scriptsize STGS \cite{li2024spacetime}}

\caption{\textbf{Qualitative comparison on the SelfCap 1200-frame scenes.}}
\label{fig:visual_SelfCap1200_final}
\end{figure*}
\label{sec:experiments}
\vspace{-5pt}
\subsection{Implementation details}
\vspace{-5pt}
We implement our approach in PyTorch and optimize all parameters using Adam with the same settings as 3DGS\cite{kerbl20233d}. The model is trained for 30k iterations. The weights for motion regularization are set to $\lambda_{reg}=0.01$, $\lambda_{motion}=0.0001$, $\lambda_{rigid}=1.0$, $\lambda_c = 50$, and $K = 3$. We perform motion-guided relocation every $N = 100$ iterations using an opacity threshold $\tau_{\alpha} = 0.005$. To improve rendering quality, we use RoMa\cite{edstedt2024roma} to obtain a point cloud initialization. All experiments are conducted on a single RTX 4090 GPU.
\begin{figure*}[!t]
\centering
\small

\begin{minipage}[c]{0.26\textwidth}
    \centering
    \includegraphics[width=\textwidth, height=3.5cm, keepaspectratio]{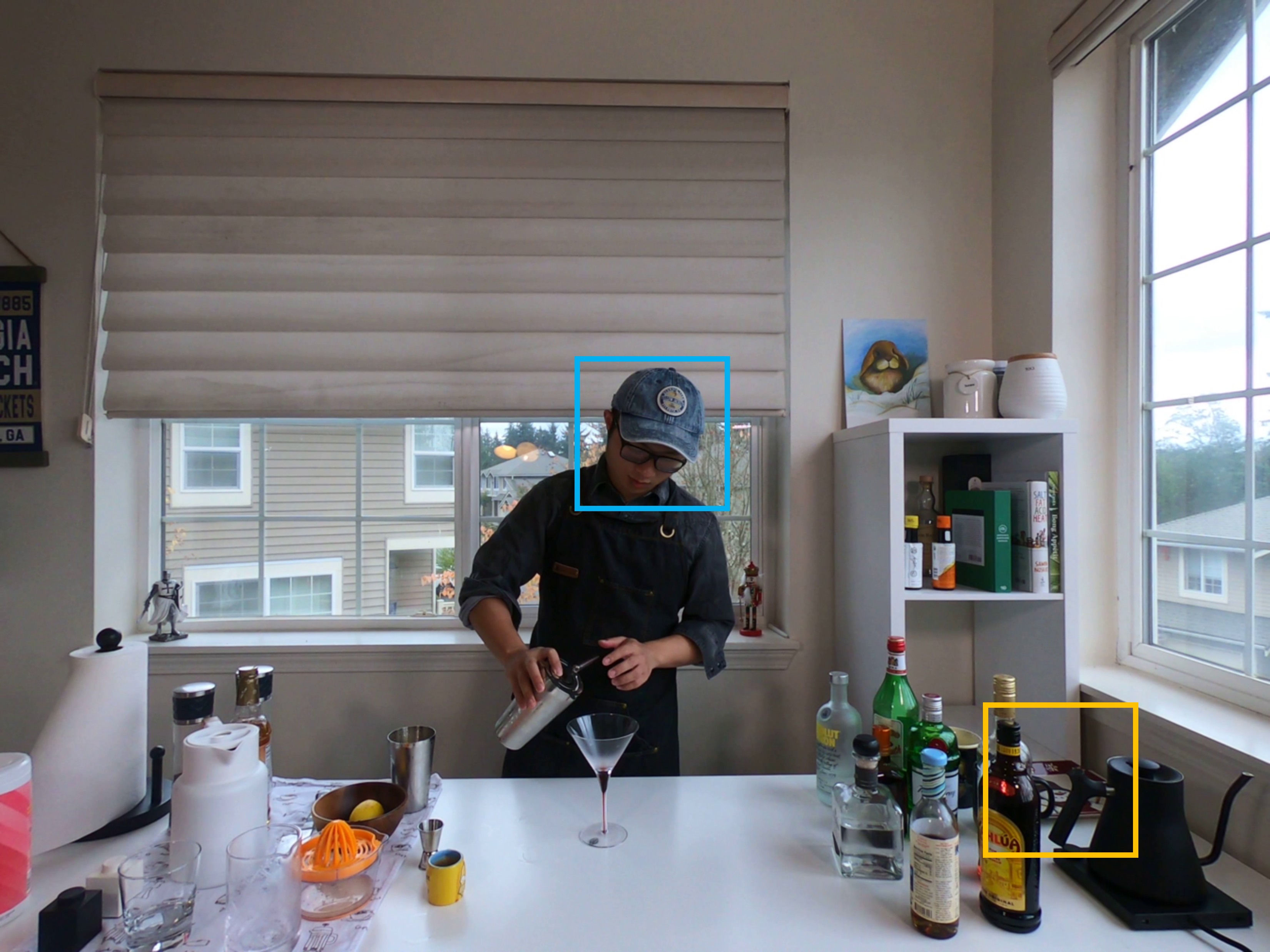}
\end{minipage}
\hfill
\begin{minipage}[c]{0.72\textwidth}
    \centering
    \includegraphics[width=0.19\textwidth]{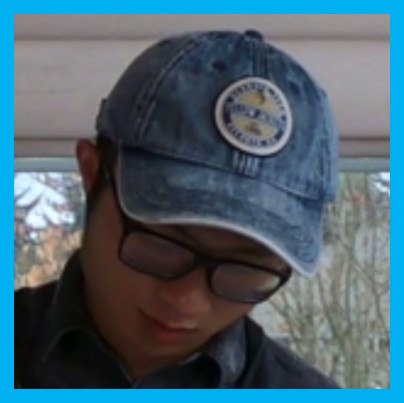}\hfill
    \includegraphics[width=0.19\textwidth]{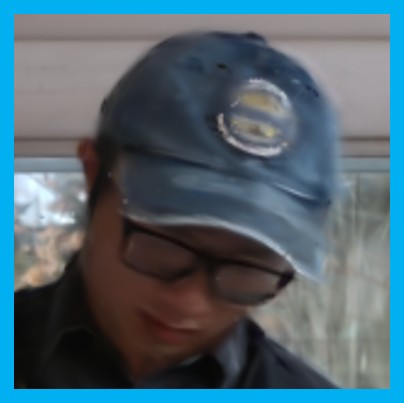}\hfill
    \includegraphics[width=0.19\textwidth]{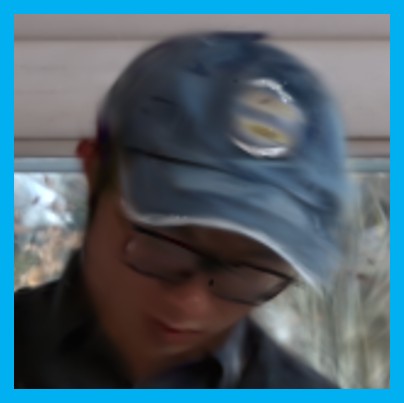}\hfill
    \includegraphics[width=0.19\textwidth]{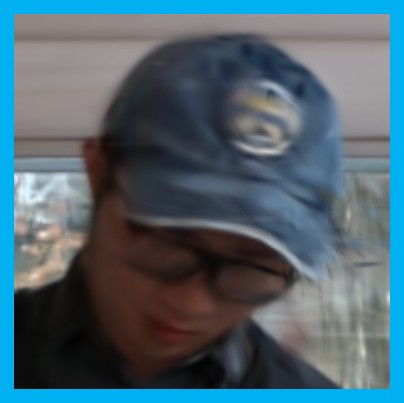}\hfill
    \includegraphics[width=0.19\textwidth]{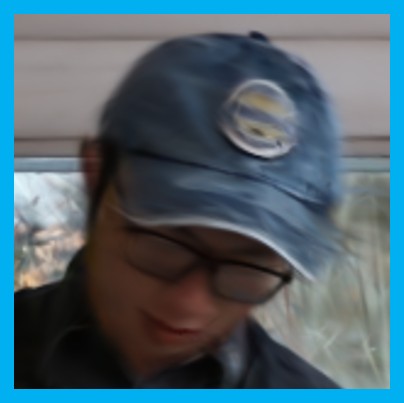}\\[0.5mm]
    \includegraphics[width=0.19\textwidth]{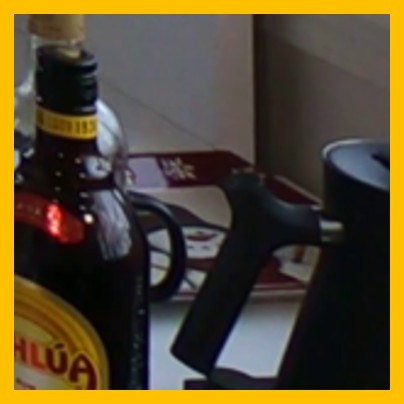}\hfill
    \includegraphics[width=0.19\textwidth]{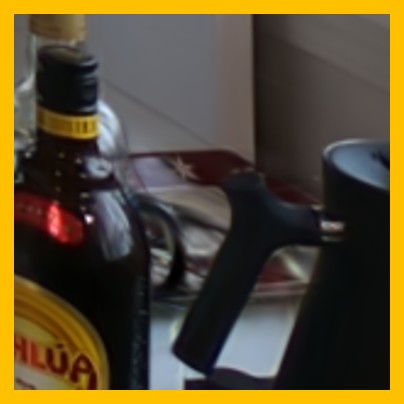}\hfill
    \includegraphics[width=0.19\textwidth]{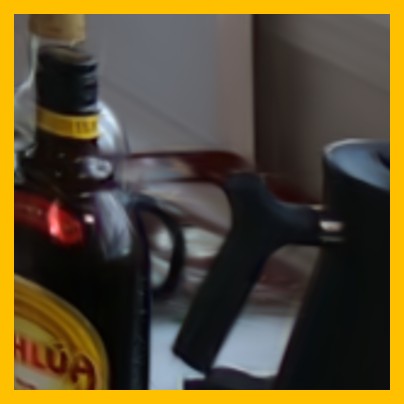}\hfill
    \includegraphics[width=0.19\textwidth]{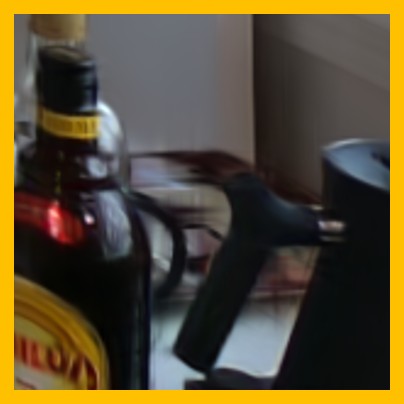}\hfill
    \includegraphics[width=0.19\textwidth]{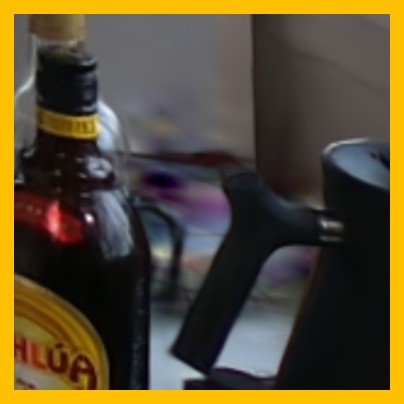}
\end{minipage}

\vspace{1mm} 

\begin{minipage}[c]{0.26\textwidth}
    \centering
    \includegraphics[width=\textwidth, height=3.5cm, keepaspectratio]{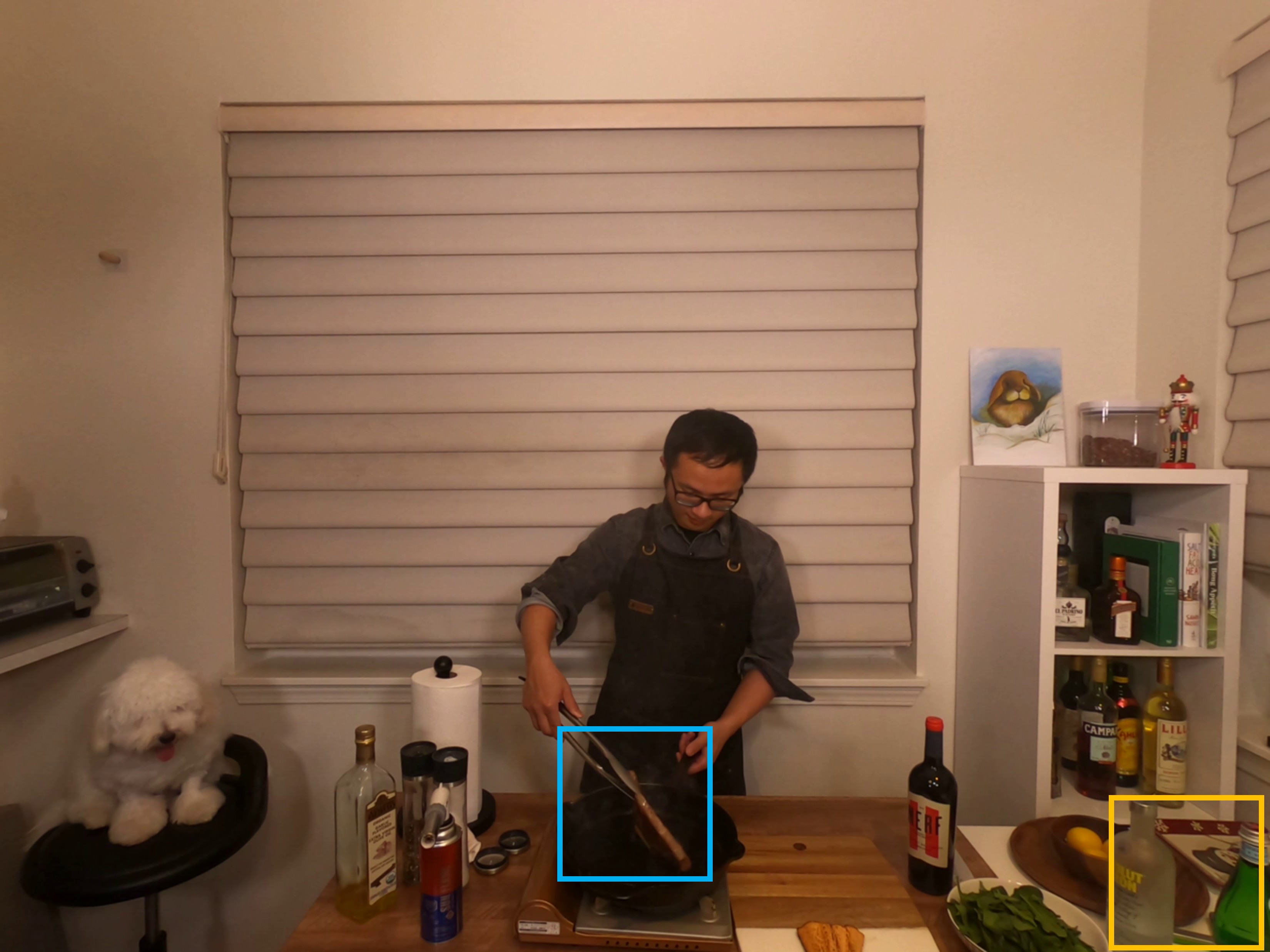}
\end{minipage}
\hfill
\begin{minipage}[c]{0.72\textwidth}
    \centering
    \includegraphics[width=0.19\textwidth]{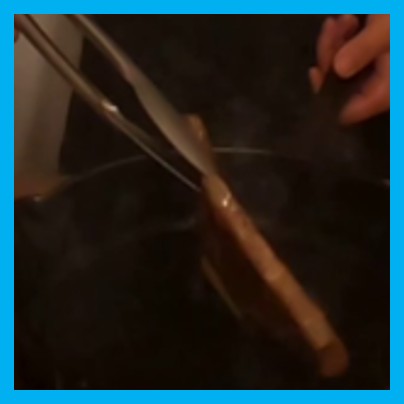}\hfill
    \includegraphics[width=0.19\textwidth]{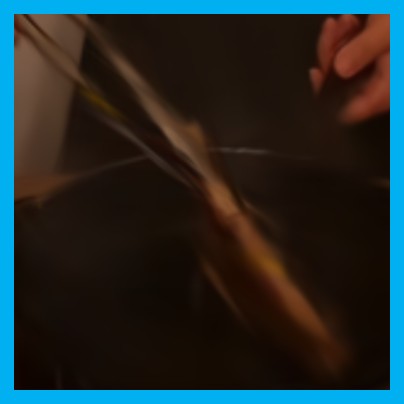}\hfill
    \includegraphics[width=0.19\textwidth]{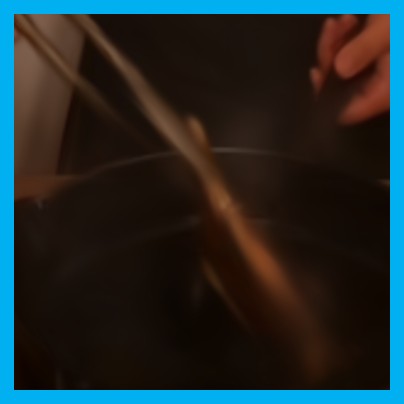}\hfill
    \includegraphics[width=0.19\textwidth]{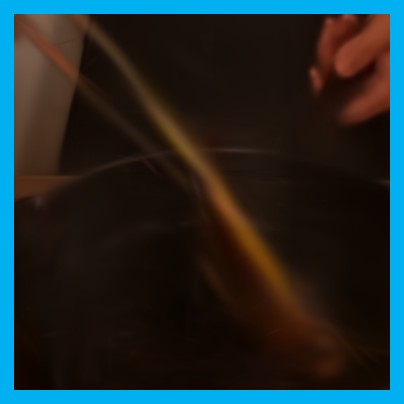}\hfill
    \includegraphics[width=0.19\textwidth]{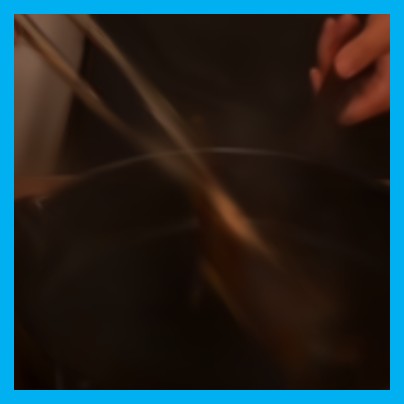}\\[0.5mm]
    \includegraphics[width=0.19\textwidth]{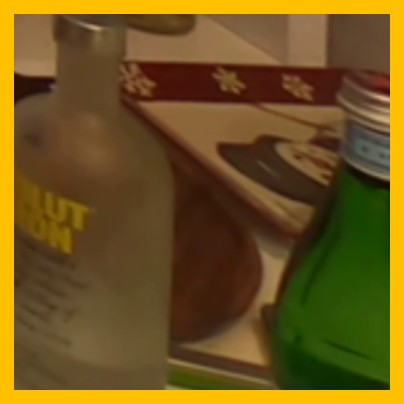}\hfill
    \includegraphics[width=0.19\textwidth]{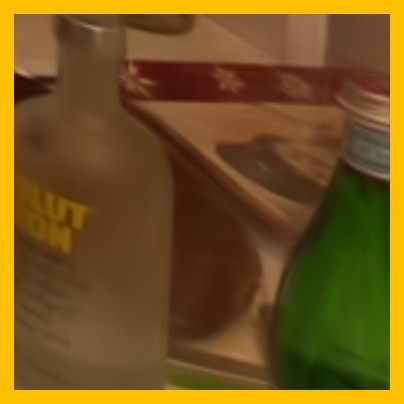}\hfill
    \includegraphics[width=0.19\textwidth]{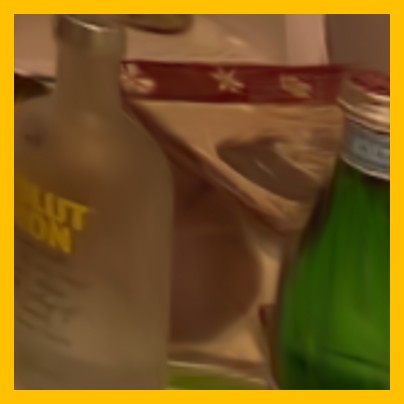}\hfill
    \includegraphics[width=0.19\textwidth]{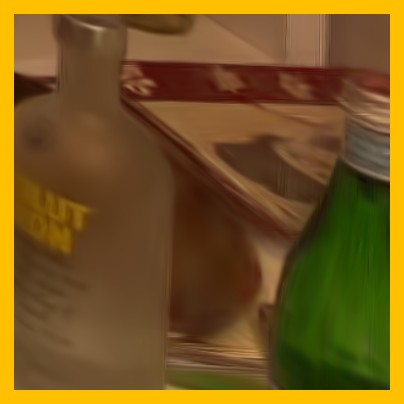}\hfill
    \includegraphics[width=0.19\textwidth]{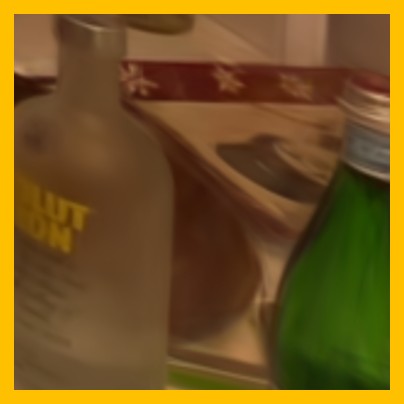}
\end{minipage}

\vspace{1mm}

\makebox[0.26\textwidth]{\scriptsize Full Image} \hfill
\makebox[0.13\textwidth]{\scriptsize GT} \hfill
\makebox[0.13\textwidth]{\scriptsize Ours} \hfill
\makebox[0.13\textwidth]{\scriptsize FTGS \cite{wang2025freetimegs}} \hfill
\makebox[0.13\textwidth]{\scriptsize Ex4DGS \cite{lee2024fully}} \hfill
\makebox[0.13\textwidth]{\scriptsize 4DGS \cite{yang2023real}}

\caption{\textbf{Qualitative results on the N3V dataset.}}
\label{fig:visual_comparison}
\end{figure*}
\vspace{-7pt}
\subsection{Experimental settings}
We evaluate TRiGS on the The Neural 3D Video(N3V)\cite{li2022neural} and SelfCap\cite{wang2025freetimegs} datasets against recent state-of-the-art dynamic synthesis methods \cite{li2025gifstream, wu2025localdygs, yang2023real, lee2024fully, li2024spacetime, wang2025freetimegs}. We assess rendering quality (PSNR, SSIM, LPIPS) and efficiency (FPS, Gaussian count, memory usage). For a fair comparison, all methods are initialized with the same RoMa-derived point cloud. Since the official code for FTGS is unavailable, we report results from our reproduction ($\ast$), which we validated by achieving comparable or superior performance to the original paper on the N3V dataset (Table~\ref{tab:comparison_results}).
\vspace{-5pt}
\subsubsection{SelfCap}
The SelfCap dataset provides challenging dynamic sequences characterized by fast and complex motions. It is captured using a multi-view system comprising 22 to 24 cameras at 60 FPS. For our evaluation, we utilize six distinct scenes: \textit{bike1}, \textit{bike2}, \textit{corgi1}, \textit{corgi2}, \textit{dance}, and \textit{yoga}. To rigorously assess long-term temporal stability and scalability, we edit the original videos to construct three variations of extended datasets consisting of 600, 900, and 1200 frames. The standard resolution is 3840$\times$2160, except for the bike scenes, which are provided at 1080$\times$1080.
\vspace{-15pt}
\subsubsection{Neural3DV}
The Neural 3D Video (N3V) dataset is a widely adopted standard benchmark for dynamic scene reconstruction. Following established protocols, we downsample the original 2704$\times$2028 videos (recorded at 30 FPS) and apply the official camera split for training and testing.
\vspace{-10pt}
\subsection{Comparisons}
\subsubsection{Quantitative comparisons}
\begin{table*}[t]
  \centering
  \caption{
  Quantitative comparison on N3V datasets.
  \colorbox{best}{best}, \colorbox{second}{second-best}, and \colorbox{third}{third-best} results are highlighted in the table.
  ($^{\dagger}$ Results reported in the original paper, using 0.5M initial points.)
  }
  \label{tab:comparison_results}
  \tiny
  \setlength{\tabcolsep}{2pt}
  \renewcommand{\arraystretch}{0.9}
  \resizebox{0.70\textwidth}{!}{%
  \begin{tabular}{lcccc}
    \toprule
    Method & PSNR$\uparrow$ & DSSIM$_1\downarrow$ & DSSIM$_2\downarrow$ & LPIPS$\downarrow$ \\
    \midrule
    HexPlane$^2$\cite{cao2023hexplane}             & 31.71 & - & \cellcolor{second}0.014 & 0.075 \\
    K-Planes\cite{fridovich2023k}                  & 31.63 & - & 0.018 & - \\
    MixVoxels\cite{wang2023mixed}                  & 31.73 & - & \cellcolor{third}0.015 & 0.064 \\
    HyperReel\cite{attal2023hyperreel}             & 31.10 & 0.036 & - & 0.096 \\
    NeRFPlayer\cite{song2023nerfplayer}            & 30.96 & 0.034 & - & 0.111 \\
    \midrule
    Deformable-3DGS\cite{yang2024deformable}       & 31.15 & 0.030 & - & 0.049 \\
    C-D3DGS\cite{katsumata2024compact}             & 30.46 & - & 0.022 & 0.150 \\
    SWinGS\cite{liu2024swings}                     & 31.10 & 0.030 & - & 0.096 \\
    Ex4DGS\cite{lee2024fully}                      & 32.11 & 0.030 & \cellcolor{third}0.015 & 0.048 \\
    4DGS\cite{yang2023real}              & 32.01 & - & \cellcolor{second}0.014 & 0.055 \\
    STGS\cite{li2024spacetime}                     & 32.05 & \cellcolor{third}0.026 & \cellcolor{second}0.014 & 0.044 \\
    GIFStream\cite{li2025gifstream}                & 31.75 & - & - & 0.051 \\
    LocalDyGS\cite{wu2025localdygs}                & \cellcolor{third}32.28 & 0.028 & \cellcolor{second}0.014 & \cellcolor{third}0.043 \\
    DASH\cite{chen2025dash}                     & 32.22 & - & - & - \\
    Swift4D\cite{wu2025swift4d}                              & 32.23 & - & 0.014 & 0.043 \\
    FTGS$^{\dagger}$\cite{wang2025freetimegs}      & \cellcolor{second}32.97 & 0.028 & \cellcolor{second}0.014 & \cellcolor{third}0.043 \\
    FTGS$^{\ast}$                                  & 32.80 & \cellcolor{second}0.021 & - & \cellcolor{second}0.040 \\
    \midrule
    \textbf{Ours}       
      & \cellcolor{best}\textbf{33.36}
      & \cellcolor{best}\textbf{0.019}
      & \cellcolor{best}\textbf{0.010}
      & \cellcolor{best}\textbf{0.031} \\
    \bottomrule
  \end{tabular}%
  }
\end{table*}
We first evaluate our method on the extended sequences of the SelfCap dataset to demonstrate its robustness over long-term frame budgets. As shown in Table~\ref{tab:selfcap_quality}, TRiGS achieves the highest rendering quality across 600, 900, and 1200 frames. Notably, while explicit 4DGS baselines like FTGS suffer from significant quality degradation (e.g., PSNR dropping from 26.23 to 25.41) as the sequence lengthens, TRiGS maintains high fidelity (PSNR of 26.05 at 1200 frames) without severe performance drops.

This stable performance is further highlighted in our efficiency analysis (Table~\ref{tab:selfcap_efficiency}). When processing extended videos, existing explicit 4D primitive models consistently suffer from severe memory bottlenecks. For instance, FTGS doubles its primitive count and memory footprint (up to 977 MB) when scaling from 600 to 1200 frames. Furthermore, while streaming-based models like GIFStream are designed to handle extended frames, they incur a massive memory overhead (up to 2825 MB) and still yield suboptimal rendering quality. In contrast, by effectively capturing continuous geometric transformations, TRiGS maintains a highly compact and constant representation of only 0.5M Gaussians and 160 MB of memory, maintaining over 110 FPS across all frame budgets.

Finally, to verify that our method is not only effective for long videos but also excels on standard-length sequences, we evaluate TRiGS on the N3V dataset (typically $\sim$300 frames). As reported in Table~\ref{tab:comparison_results}, TRiGS achieves state-of-the-art performance with the highest PSNR (33.36) and lowest LPIPS (0.031). This confirms that our continuous motion modeling generally surpasses existing approaches in capturing complex dynamic scenes, regardless of the video length.
\begin{table*}[t]
  \centering
  \caption{Efficiency comparison on SelfCap across 600, 900, and 1200 frame budgets.}
  \label{tab:selfcap_efficiency}
  \tiny
  \setlength{\tabcolsep}{3pt}
  \renewcommand{\arraystretch}{0.96}
  \resizebox{\textwidth}{!}{%
  \begin{tabular}{l|ccc|ccc|ccc}
    \toprule
    \multirow{2}{*}[-0.4em]{Method} & \multicolumn{3}{c|}{600 frames} & \multicolumn{3}{c|}{900 frames} & \multicolumn{3}{c}{1200 frames} \\
    \cmidrule(lr){2-4} \cmidrule(lr){5-7} \cmidrule(lr){8-10}
    & FPS$\uparrow$ & \#Gaussians$\downarrow$ & MB$\downarrow$
    & FPS$\uparrow$ & \#Gaussians$\downarrow$ & MB$\downarrow$
    & FPS$\uparrow$ & \#Gaussians$\downarrow$ & MB$\downarrow$ \\
    \midrule

    GIFStream\cite{li2025gifstream}
      & 94 & 6.47M & 1586
      & 98 & 8.85M & 2169
      & 102 & 11.53M & 2825 \\

    4DGS\cite{yang2023real}
      & 43 & 2.51M & 4665
      & 43 & 2.70M & 5021
      & 44 & 2.70M & 5015 \\

    STGS\cite{li2024spacetime}
      & 90 & 1.31M & 190
      & 83 & 2.01M & 291
      & 90 & 2.45M & 355 \\

    FTGS$^{\ast}$\cite{wang2025freetimegs}
      & 113 & 2M & 490
      & 110 & 3M & 733
      & 106 & 4M & 977 \\

    \textbf{Ours}
      & \textbf{116} & \textbf{0.5M} & \textbf{160}
      & \textbf{111} & \textbf{0.5M} & \textbf{160}
      & \textbf{120} & \textbf{0.5M} & \textbf{160} \\

    \bottomrule
  \end{tabular}%
  }
\end{table*}
We attribute this to the increased per-primitive motion capacity, which enables each Gaussian to accurately capture local dynamics, successfully representing the entire sequence with drastically fewer primitives.
\vspace{-10pt}
\subsubsection{Qualitative comparisons}
Consistent with our quantitative results, visual comparisons in Fig. ~\ref{fig:visual_SelfCap1200_final} and Fig. ~\ref{fig:visual_comparison} confirm the superiority of TRiGS in rendering complex dynamic scenes, particularly over extended sequences. While baseline methods suffer from error accumulation over time—often resulting in severe ghosting, blurriness, and structural degradation in fast-moving regions (e.g., articulating limbs or rotating wheels)—TRiGS consistently preserves sharp details and accurate geometries. By leveraging continuous geometric transformations with local anchors, our approach effectively captures intricate local dynamics without the need for massive primitive duplication. This translates to highly realistic, temporally stable novel-view synthesis, which is further demonstrated in our supplementary video.

\begin{table}[t]
  \centering
  \caption{Ablation study of core components in TRiGS on the 1200-frame extended sequence. We evaluate the impact of trajectory modeling, anchor modes, and base parameterization on rendering quality and memory efficiency.}
  \label{tab:ablation_components}
  \scriptsize
  \setlength{\tabcolsep}{2pt}
  \renewcommand{\arraystretch}{0.95}
  \resizebox{\columnwidth}{!}{%
  \begin{tabular}{l|ccc|ccc|cc}
    \toprule
    \multirow{2}{*}{Configuration} & \multicolumn{3}{c|}{Modeling Choices} & \multicolumn{3}{c|}{Quality Metrics} & \multicolumn{2}{c}{Efficiency} \\
    \cmidrule(lr){2-4} \cmidrule(lr){5-7} \cmidrule(lr){8-9}
    & Trajectory & Anchor & Base Param. & PSNR$\uparrow$ & SSIM$\uparrow$ & LPIPS$\downarrow$ & \#Gaussians$\downarrow$ & MB$\downarrow$ \\
    \midrule
    Baseline & Linear & None & $\mathbb{R}^3$ & 24.83 & 0.865 & 0.162 & 2.10M & 512 \\
    Naive SE(3) & Rigid Body & Origin & $SO(3)\times\mathbb{R}^3$ & 23.95 & 0.842 & 0.198 & 2.85M & 694 \\
    Local SE(3) & Rigid Body & Local & $\mathfrak{se}(3)$ & 25.52 & 0.891 & 0.125 & 0.60M & 185 \\
    Linear + B\'ezier & Non-linear & None & $\mathbb{R}^3$ & 25.10 & 0.873 & 0.145 & 1.85M & 450 \\
    \midrule
    \textbf{Full TRiGS} & \textbf{Non-linear} & \textbf{Local} & $\mathfrak{se}(3)$ & \textbf{26.05} & \textbf{0.904} & \textbf{0.099} & \textbf{0.50M} & \textbf{160} \\
    \bottomrule
  \end{tabular}%
  }
\end{table}
\begin{figure*}[t]
\centering

\includegraphics[width=0.158\textwidth]{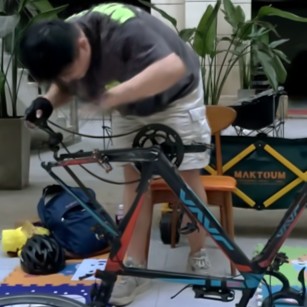}\hfill
\includegraphics[width=0.158\textwidth]{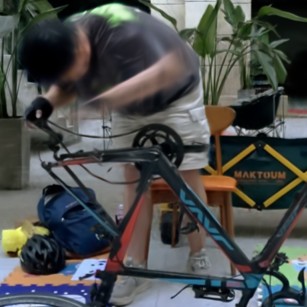}\hfill
\includegraphics[width=0.158\textwidth]{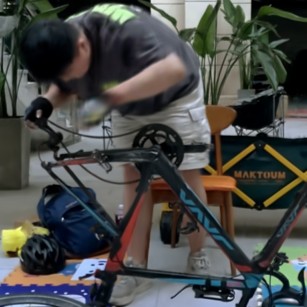}\hfill
\includegraphics[width=0.158\textwidth]{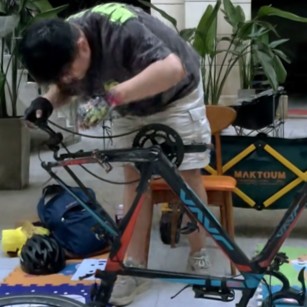}\hfill
\includegraphics[width=0.158\textwidth]{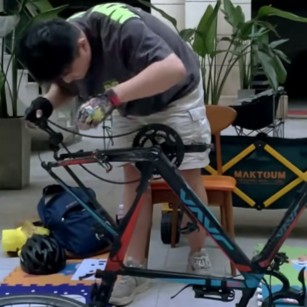}\hfill
\includegraphics[width=0.158\textwidth]{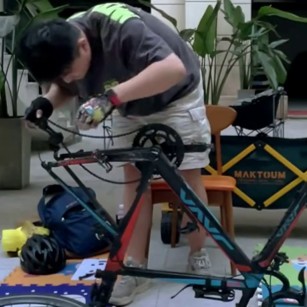}

\vspace{2pt}

{\footnotesize
\makebox[0.158\textwidth][c]{Baseline}\hfill
\makebox[0.158\textwidth][c]{NaiveSE3}\hfill
\makebox[0.158\textwidth][c]{Linear+B\`ezier}\hfill
\makebox[0.158\textwidth][c]{LocalSE3}\hfill
\makebox[0.158\textwidth][c]{Full}\hfill
\makebox[0.158\textwidth][c]{Ground Truth}
}

\vspace{-2mm}

\caption{\textbf{Visual ablation on the SelfCap \textit{bike2} scene.} Compared to incomplete configurations that suffer from motion artifacts and blurriness, \textbf{Full TRiGS} successfully preserves sharp geometric details and structural integrity.}
\label{fig:visual_ablation}

\vspace{-6mm}
\end{figure*}
\subsection{Analysis and Ablation Study}
\subsubsection{Ablation of Core Components}
To validate the necessity of each structural design in TRiGS, we conduct an ablation study on the challenging 1200-frame sequence, as summarized in Table~\ref{tab:ablation_components}.

The \textit{Baseline} relies on a simple linear velocity model without geometric constraints. Because it fails to track continuous curved trajectories, it naturally suffers from spatial mismatches. A straightforward attempt to apply geometric constraints is \textit{Naive SE(3)}, which models rigid transformations around a global origin with independent rotation and translation parameters ($SO(3)\times\mathbb{R}^3$). Interestingly, this configuration yields the lowest performance (23.95 PSNR) among all settings. This empirically validates our theoretical analysis in Sec.~\ref{sec:method}: Utilizing a single shared anchor misrepresents independent motions, and optimizing uncoupled parameters leads to entangled optimization and accumulated drift.

In contrast, introducing learnable local anchors and coupled Lie algebra parameterization (\textit{Local SE(3)}) yields a quantum jump in performance. The PSNR significantly improves to 25.52, confirming that decoupling global motions into geometrically consistent local rigid transformations is crucial for stable optimization. Furthermore, we evaluate \textit{Linear + Bézier}, which applies non-linear Bézier curves only to unconstrained translations. While it slightly improves upon the baseline, it demonstrates that non-linear trajectories alone cannot guarantee geometric consistency without rigid-body constraints. Ultimately, our \textbf{Full TRiGS} model, which synergistically combines local SE(3) transformations with hierarchical decomposition with Bézier Residual, achieves the highest visual fidelity (26.05 PSNR).

\textbf{Analysis of Representation Capacity} 
It is important to clarify the significant differences in Gaussian counts across the configurations in Table~\ref{tab:ablation_components}. TRiGS intentionally operates under a strict, fixed memory budget (0.50M primitives) by utilizing a relocation strategy without standard densification. However, for the ablated baselines (e.g., \textit{Baseline} and \textit{Naive SE(3)}), we permitted standard gradient-based densification. This was a deliberate experimental design to evaluate the upper-bound rendering capability of their motion models; restricting these baseline configurations to 0.50M primitives would result in catastrophic rendering collapse due to their inability to track complex trajectories. 

Interestingly, even when allowed to aggressively proliferate (up to 2.85M primitives) to compensate for spatial mismatches, these configurations still fail to reach the visual fidelity of TRiGS. Because they lack accurate geometric constraints, they attempt to patch trajectory errors by excessively splitting primitives, leading to severe memory bloat without resolving the underlying structural misalignment. This confirms that our highly expressive motion formulation effectively resolves complex dynamics under a minimal memory budget, demonstrating the advantage of structured motion parameterization over brute-force capacity scaling.
\vspace{-15pt}
\begin{table}[t]
  \centering
  \caption{Ablation study on optimization components in TRiGS. We evaluate the impact of gauge fixing, regularization losses, and relocation strategy on the 1200-frame sequence.}
  \label{tab:ablation_optimization}
  \tiny
  \setlength{\tabcolsep}{2pt}
  \renewcommand{\arraystretch}{0.9}
  \resizebox{0.70\columnwidth}{!}{%
  \begin{tabular}{l|ccc}
    \toprule
    Configuration & PSNR$\uparrow$ & SSIM$\uparrow$ & LPIPS$\downarrow$ \\
    \midrule
    w/o Gauge Fixing ($a_{i,\perp}$) & 25.15 & 0.880 & 0.134 \\
    w/o Rigid Reg. ($\mathcal{L}_{rigid}$) & 25.48 & 0.885 & 0.128 \\
    w/o Motion Smoothness ($\mathcal{L}_{motion}$) & 25.66 & 0.892 & 0.115 \\
    w/o Motion-guided Relocation & 25.80 & 0.897 & 0.108 \\
    \midrule
    \textbf{Full TRiGS Architecture} & \textbf{26.05} & \textbf{0.904} & \textbf{0.099} \\
    \bottomrule
  \end{tabular}%
  }
\end{table}
\subsubsection{Ablation of Optimization Components}
Beyond the core architecture, we ablate our specific optimization and regularization strategies in Table~\ref{tab:ablation_optimization}. Removing the gauge fixing step (\textit{w/o Gauge Fixing}) causes the largest quality drop (25.15 PSNR). Without it, the translational component suffers from unidentifiability, which severely destabilizes the optimization process. 

Motion regularizers are also crucial for rendering fidelity. Removing rigid regularization (\textit{w/o Rigid Reg.}) disrupts the spatial alignment of neighboring motions, causing subtle surface tearing and a drop in SSIM (0.885). Omitting the motion smoothness loss (\textit{w/o Motion Smoothness}) leads to jittery and temporally inconsistent trajectories. Finally, disabling opacity-based recycling (\textit{w/o Motion-guided Relocation}) prevents the active reallocation of redundant Gaussians to complex dynamic regions, lowering the overall representational capacity. Ultimately, integrating these strategies allows the full TRiGS architecture to achieve the highest visual quality (26.05 PSNR, 0.099 LPIPS).
\section{Conclusion}
\label{sec:conclusion}
We presented TRiGS, a new dynamic rendering framework that increases per-primitive temporal modeling capacity for novel-view synthesis. TRiGS mitigates the temporal fragmentation and memory growth from translation-only motion or short temporal opacity via three designs:
(i) Coupled rigid-body parameterization with a closed-form exponential map, (ii) Hierarchical B\'ezier residual within a visibility-driven temporal window, and (iii) Local anchor-centered deformation to capture faithful local motions. Moreover, our motion-aware training improves long-sequence stability by reducing drift while maintaining effective primitive coverage under a fixed budget. Experiments show that TRiGS achieves higher rendering quality with fewer Gaussians and lower memory overhead, and remains stable on longer videos where prior methods lose temporal identity and exhibit failure modes.


%
%
\bibliographystyle{unsrt}
\bibliography{references}

\clearpage
\appendix
\setcounter{figure}{0}
\setcounter{table}{0}
\setcounter{equation}{0}
\setcounter{section}{0}
\renewcommand{\thefigure}{S\arabic{figure}}
\renewcommand{\thetable}{S\arabic{table}}
\renewcommand{\theequation}{S\arabic{equation}}
\renewcommand{\thesection}{\Alph{section}}

\title{Supplementary Material}
\titlerunning{Supplementary Material}
\maketitle

\section{Limitations and Future Work}
While TRiGS significantly improves the temporal stability and memory efficiency of 4D Gaussian Splatting over extended sequences, it shares a common limitation with existing explicit dynamic representations regarding the optimization paradigm. Specifically, our method inherently relies on a lengthy per-scene optimization process, requiring the continuous motion parameters and Gaussians to be trained from scratch for each specific video. This prevents the model from being instantly applied to novel, unseen scenes. To overcome this limitation, our future work will focus on transforming the current per-scene optimization pipeline into a training-free, feed-forward architecture. By integrating our continuous rigid-body formulation with data-driven foundation models, we aim to achieve instant 4D reconstruction without the need for video-specific training, thereby drastically accelerating the pipeline for broader real-world applications.
\label{sec:suppl_Limitations}
\section{Extended Implementation Details}
We provide additional implementation details to improve reproducibility and to clarify the fairness of our baseline comparisons, especially for the reproduced FreeTimeGS (FTGS)\cite{wang2025freetimegs} baseline. Unless otherwise stated, all methods are trained and evaluated under the same data preprocessing and evaluation protocol as in the main paper, including the same camera calibration, train/test split, image resolution, and metric computation pipeline. In addition, following the main paper, all methods are initialized from the same RoMa-derived point cloud to avoid confounding the comparison with different initialization quality.

\textbf{Shared rendering backend.}
Both TRiGS and our reproduced FTGS baseline are implemented using the same \texttt{gsplat}-based Gaussian rasterization backend. This choice is important for fair comparison: it ensures that the observed performance gap is not caused by differences in CUDA kernels, alpha compositing rules, or renderer-specific engineering details. Instead, the comparison more directly reflects the difference in scene representation and motion modeling.

\textbf{FTGS reproduction details.}
Because the official training code of FTGS\cite{wang2025freetimegs} was not publicly available at the time of our experiments, we reproduced FTGS in our own codebase while preserving the original formulation as closely as possible. Concretely, each FTGS primitive is parameterized by its position, time, duration, velocity, scale, orientation, opacity, and spherical harmonics coefficients. Its motion is modeled by the original translation-only linear function,
\[
\mu_i(t)=\mu_i + v_i (t-\mu_{t,i}),
\]
and its contribution is modulated by a Gaussian-shaped temporal opacity centered at the primitive time. We also retain the core optimization components described in FTGS, including opacity regularization, periodic relocation, and 4D initialization from RoMa-based multi-view points with velocity initialization from neighboring 3D correspondences.

\textbf{Optimization protocol for FTGS.}
Our reproduced FTGS follows the training protocol described in the original paper: Adam optimization with the standard 3DGS settings, a 30k-iteration schedule for standard-length sequences, and periodic relocation performed every 100 iterations. For controlled comparison on Neural3DV, we use the same primitive budget as the original FTGS setting, i.e., \textbf{500k Gaussians}. This matches both the configuration used in our reproduced implementation and the budget reported in the FTGS paper for the 500k setting.

\textbf{Evidence of faithful reproduction.}
We emphasize that our FTGS reproduction is not only conceptually faithful but also \emph{empirically validated}. Table~\ref{tab:ftgs_reproduction} compares the published FTGS result on the Neural3DV benchmark under the 500k-primitive setting with our reproduced FTGS evaluated under the same budget. The published FTGS reports \textbf{32.97} PSNR, \textbf{0.028} DSSIM$_1$, and \textbf{0.043} LPIPS, while our reproduction achieves \textbf{32.80} PSNR, \textbf{0.021} DSSIM$_1$, and \textbf{0.040} LPIPS. The PSNR gap is only \textbf{0.17 dB}, and the reproduced model is slightly better in DSSIM$_1$ and LPIPS. Since the primitive budget is also matched at \textbf{500k} in both cases, this close agreement indicates that our implementation faithfully recovers the optimization behavior and effective model capacity of the original FTGS formulation, rather than serving as a weakened approximation.

\textbf{Relation to the main-paper comparisons.}
The matched-500k comparison above is used specifically to validate the fidelity of our FTGS reproduction. In contrast, the long-sequence SelfCap comparisons in the main paper report each method in its natural operating regime. Under those settings, FTGS grows from \textbf{2M} to \textbf{4M} Gaussians as the sequence length increases from 600 to 1200 frames, whereas TRiGS maintains a fixed budget of \textbf{0.5M} primitives. Therefore, the N3V validation in Table~\ref{tab:ftgs_reproduction} and the extended-sequence comparison in the main paper serve complementary purposes: the former validates that our FTGS implementation is faithful, while the latter highlights the different scaling behaviors of fragmented linear motion and our continuous rigid-body formulation.

\textbf{Why this matters for fair comparison.}
This shared implementation protocol substantially reduces ambiguity in baseline reproduction. Since both methods use the same Gaussian rasterizer and the reproduced FTGS matches the published FTGS result under the same 500k budget, the performance differences reported in the main paper can be attributed more confidently to the underlying motion representation, rather than to mismatched rendering infrastructure, primitive budget, or an under-tuned baseline.

\begin{table}[t]
\centering
\caption{
Validation of our FTGS reproduction on Neural3DV under the matched 500k-primitive setting.
The reproduced FTGS closely matches the published FTGS result while using the same primitive budget, supporting the faithfulness of our implementation.
}
\label{tab:ftgs_reproduction}
\resizebox{0.9\linewidth}{!}{
\begin{tabular}{lcccc}
\toprule
Method & PSNR$\uparrow$ & DSSIM$_1\downarrow$ & LPIPS$\downarrow$ & \#Gaussians \\
\midrule
FTGS (published, 500k) & 32.97 & 0.028 & 0.043 & 500k \\
FTGS (ours, reproduced) & 32.80 & 0.021 & 0.040 & 500k \\
\bottomrule
\end{tabular}
}
\end{table}
\label{sec:suppl_implementation}
\section{More Experiments Results}
\subsection{More Quantitative Results on the SelfCap Dataset}

In this section, we provide a comprehensive per-scene quantitative evaluation on the SelfCap dataset to supplement the main paper. We report the rendering quality using standard metrics (PSNR, SSIM, and LPIPS) across six diverse dynamic scenes: \textit{bike1}, \textit{bike2}, \textit{corgi1}, \textit{corgi2}, \textit{dance}, and \textit{yoga}. 

To thoroughly demonstrate the temporal scalability and robustness of our approach, we break down the performance across different extended frame budgets. The detailed per-scene results for the 600-frame, 900-frame, and 1200-frame settings are presented in Table~\ref{tab:selfcap_600}, Table~\ref{tab:selfcap_900}, and Table~\ref{tab:selfcap_1200}, respectively. Consistent with the main text, these extended tables confirm that TRiGS maintains superior high-fidelity rendering across all scenes without suffering from severe performance degradation as the video length increases.

\subsection{Matched-Budget Comparison at 500k Gaussians}

A natural question is whether the advantage of TRiGS still holds when all methods are constrained to the same primitive budget. To address this, we additionally evaluate all compared explicit 4DGS methods under a \textbf{matched budget of 500k Gaussians} on the challenging 1200-frame SelfCap setting. This experiment directly answers the equal-capacity question and helps isolate the effect of motion modeling from the effect of unconstrained primitive growth.

In the main long-sequence experiments, 4DGS\cite{yang2023real}  and STGS\cite{li2024spacetime}  occasionally exceeded the practical memory budget of a single RTX 4090 GPU (24GB VRAM) during unconstrained training on extended sequences, and their reported results were therefore measured from the latest executable checkpoints before OOM. In contrast, when the primitive count is explicitly capped at \textbf{500k}, both 4DGS and STGS can be trained and evaluated without OOM, enabling a cleaner and more controlled comparison. This budget cap partially changes the behavior of the baselines: 4DGS recovers some performance under the stabilized budget, and STGS does not suffer a severe degradation despite the reduced capacity. Nevertheless, \textbf{TRiGS still achieves the best overall performance}, obtaining the highest average PSNR/SSIM and the lowest average LPIPS.

As shown in Table~\ref{tab:matched_budget_500k}, TRiGS achieves \textbf{26.05} PSNR, \textbf{0.904} SSIM, and \textbf{0.099} LPIPS on average, outperforming all competing methods under the same 500k-Gaussian budget. More specifically, TRiGS attains the best LPIPS on \textbf{all six scenes}, the best PSNR on \textbf{five out of six scenes}, and the best SSIM on \textbf{five out of six scenes}. These results indicate that the advantage of TRiGS is not merely a byproduct of fixed-budget control; even when all methods are given the same representational capacity, our continuous rigid-body motion formulation remains the strongest.

\begin{table*}[t]
\centering
\caption{
Matched-budget comparison on the 1200-frame SelfCap setting, where all methods are constrained to the same budget of 500k Gaussians. Under this setting, 4DGS and STGS run without OOM on a single RTX 4090 (24GB VRAM), enabling a controlled equal-capacity comparison. Even in this favorable setting for the baselines, TRiGS achieves the best average PSNR/SSIM/LPIPS.
}
\label{tab:matched_budget_500k}

\setlength{\tabcolsep}{6pt}
\renewcommand{\arraystretch}{1.15}

\begin{tabular}{l|ccccccc}
\toprule
\textbf{1200 frames} & \textbf{bike1} & \textbf{bike2} & \textbf{corgi1} & \textbf{corgi2} & \textbf{dance} & \textbf{yoga} & \textbf{avg} \\
\midrule
\rowcolor[gray]{0.95}
\multicolumn{8}{c}{\textbf{PSNR}$\uparrow$} \\
\midrule
STGS\cite{li2024spacetime}    & 25.95 & 25.67 & 24.87 & \textbf{24.97} & 24.95 & 24.58 & 25.16 \\
4DGS\cite{yang2023real}       & 25.06 & 24.65 & 24.35 & 23.03 & 23.59 & 22.15 & 23.81 \\
FTGS\cite{wang2025freetimegs} & 25.61 & 25.92 & 25.33 & 23.25 & 26.12 & 25.04 & 25.21 \\
Ours                          & \textbf{26.45} & \textbf{26.70} & \textbf{26.25} & 24.62 & \textbf{26.85} & \textbf{25.43} & \textbf{26.05} \\

\midrule
\rowcolor[gray]{0.95}
\multicolumn{8}{c}{\textbf{SSIM}$\uparrow$} \\
\midrule
STGS\cite{li2024spacetime}    & 0.910 & \textbf{0.923} & 0.820 & 0.829 & 0.907 & 0.901 & 0.882 \\
4DGS\cite{yang2023real}       & 0.893 & 0.903 & 0.780 & 0.766 & 0.884 & 0.864 & 0.848 \\
FTGS\cite{wang2025freetimegs} & 0.887 & 0.891 & 0.814 & 0.803 & 0.906 & 0.868 & 0.861 \\
Ours                          & \textbf{0.912} & 0.918 & \textbf{0.885} & \textbf{0.875} & \textbf{0.931} & \textbf{0.903} & \textbf{0.904} \\

\midrule
\rowcolor[gray]{0.95}
\multicolumn{8}{c}{\textbf{LPIPS}$\downarrow$} \\
\midrule
STGS\cite{li2024spacetime}    & 0.085 & 0.069 & 0.188 & 0.171 & 0.089 & 0.132 & 0.122 \\
4DGS\cite{yang2023real}       & 0.179 & 0.158 & 0.353 & 0.357 & 0.162 & 0.295 & 0.251 \\
FTGS\cite{wang2025freetimegs} & 0.103 & 0.096 & 0.194 & 0.238 & 0.100 & 0.211 & 0.157 \\
Ours                          & \textbf{0.082} & \textbf{0.068} & \textbf{0.121} & \textbf{0.154} & \textbf{0.075} & \textbf{0.094} & \textbf{0.099} \\

\bottomrule
\end{tabular}

\end{table*}

\subsection{Qualitative Results on the SelfCap Dataset}
We further present qualitative comparisons on the \textbf{SelfCap} dataset to illustrate the visual differences between methods under long video sequences. Following the quantitative evaluation in the main paper, we show results for three representative scenes (\textit{bike2}, \textit{corgi2}, and \textit{yoga}) at different sequence lengths: 600, 900, and 1200 frames. These scenes contain diverse motion patterns and appearance variations, providing a comprehensive view of each method's behavior as the sequence length increases.

\begin{figure*}[!t]
\centering
\setlength{\tabcolsep}{2pt}

\begin{tabular}{>{\centering\arraybackslash}m{0.03\textwidth} c c c c}
& \textbf{GT} & \textbf{Ours} & \textbf{FTGS}\cite{wang2025freetimegs} & \textbf{STGS}\cite{li2024spacetime} \\
\adjustbox{valign=c}{\rotatebox[origin=c]{90}{\scriptsize\textbf{1200}}} &
\adjustbox{valign=c}{\includegraphics[width=0.22\textwidth]{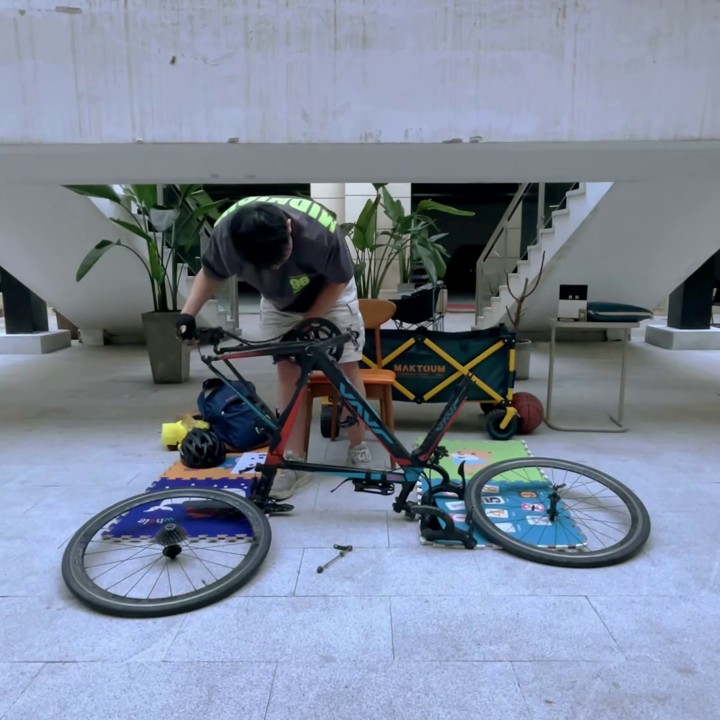}} &
\adjustbox{valign=c}{\includegraphics[width=0.22\textwidth]{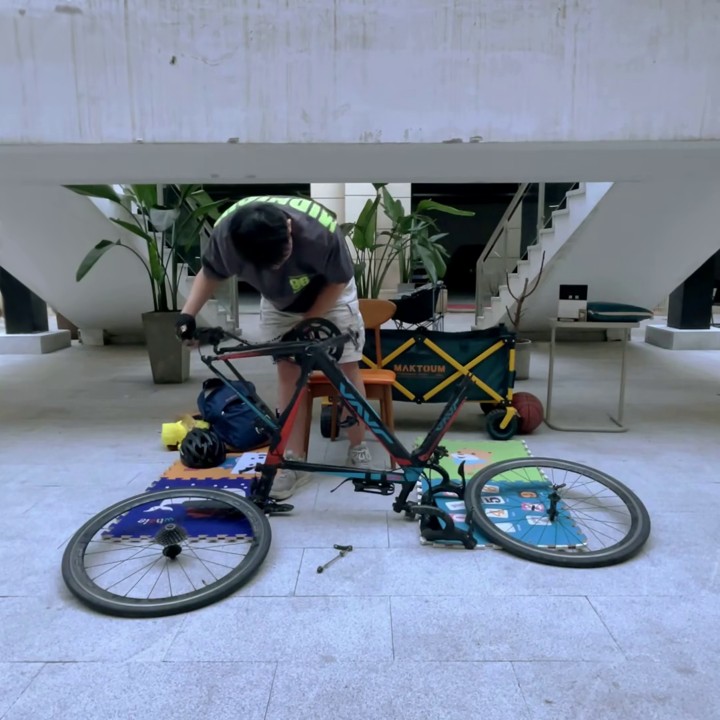}} &
\adjustbox{valign=c}{\includegraphics[width=0.22\textwidth]{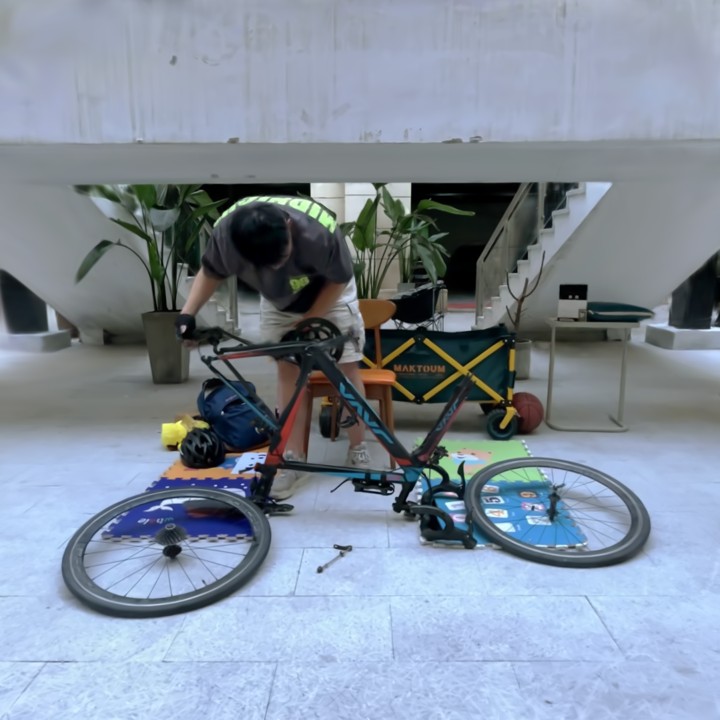}} &
\adjustbox{valign=c}{\includegraphics[width=0.22\textwidth]{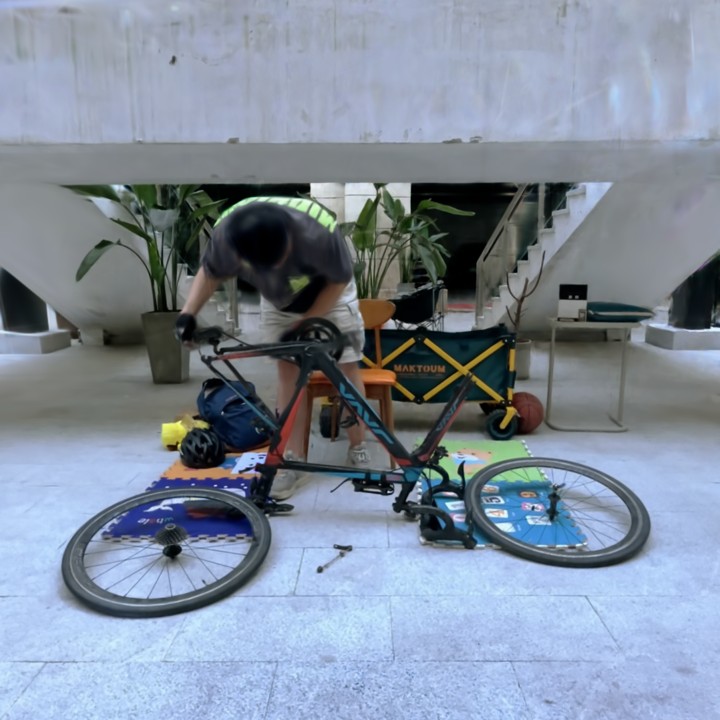}} \\
\adjustbox{valign=c}{\rotatebox[origin=c]{90}{\scriptsize\textbf{900}}} &
\adjustbox{valign=c}{\includegraphics[width=0.22\textwidth]{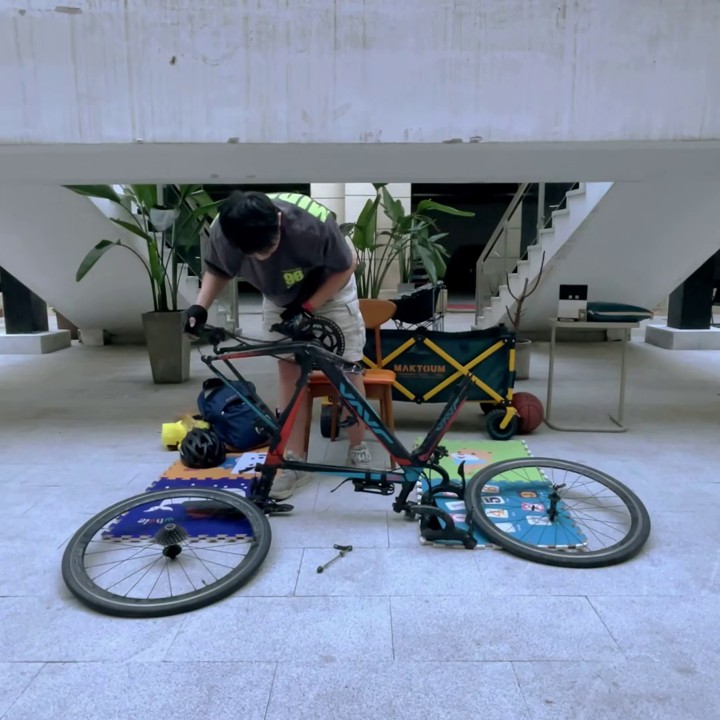}} &
\adjustbox{valign=c}{\includegraphics[width=0.22\textwidth]{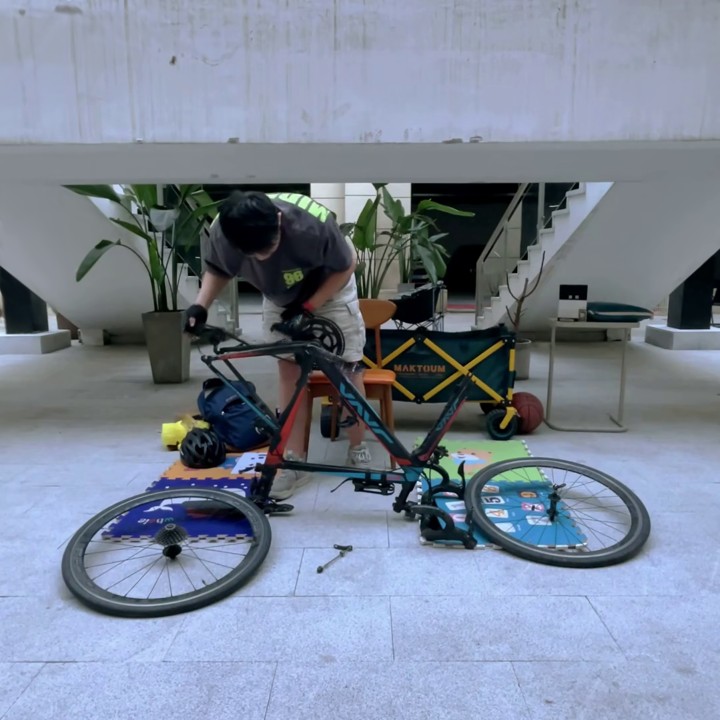}} &
\adjustbox{valign=c}{\includegraphics[width=0.22\textwidth]{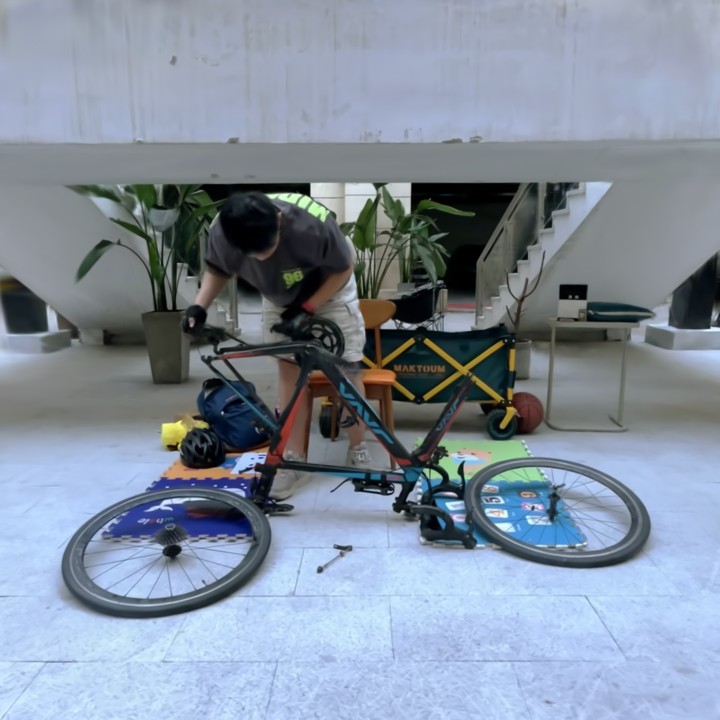}} &
\adjustbox{valign=c}{\includegraphics[width=0.22\textwidth]{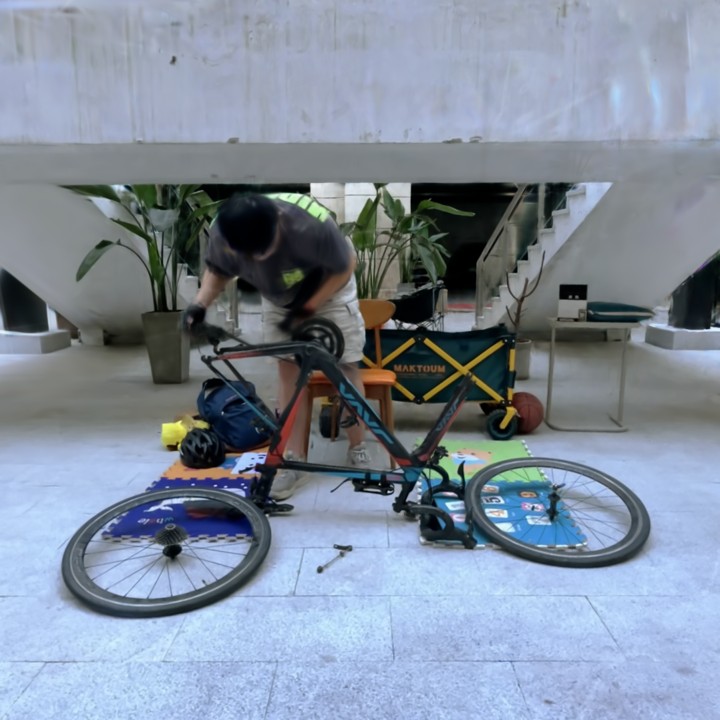}} \\
\adjustbox{valign=c}{\rotatebox[origin=c]{90}{\scriptsize\textbf{600}}} &
\adjustbox{valign=c}{\includegraphics[width=0.22\textwidth]{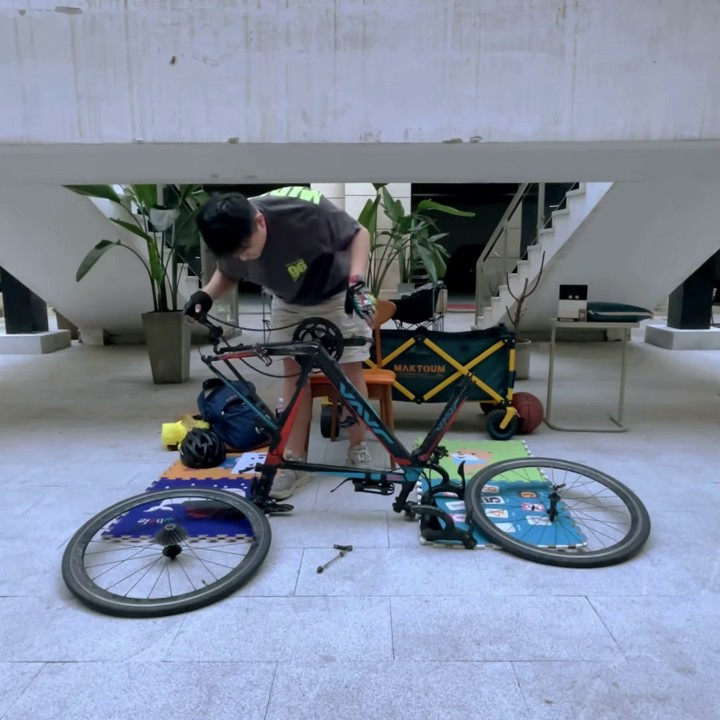}} &
\adjustbox{valign=c}{\includegraphics[width=0.22\textwidth]{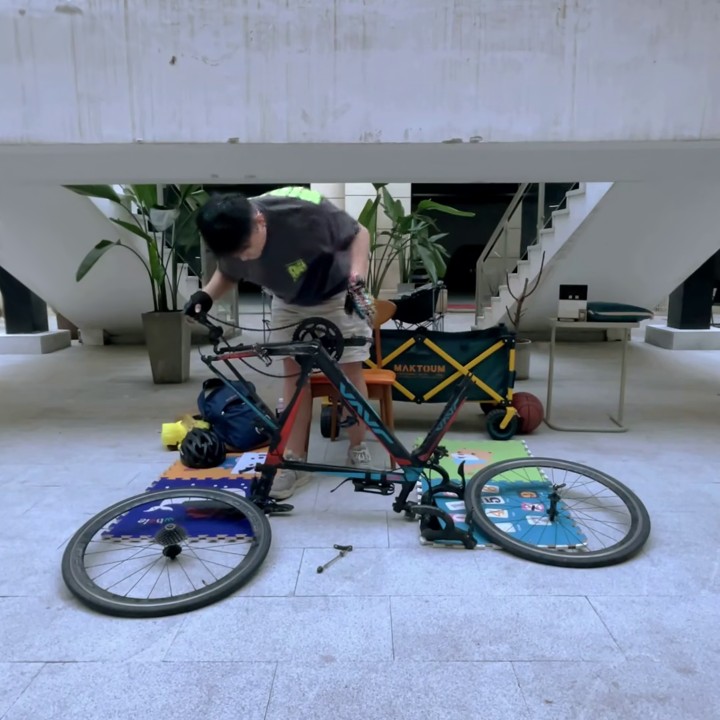}} &
\adjustbox{valign=c}{\includegraphics[width=0.22\textwidth]{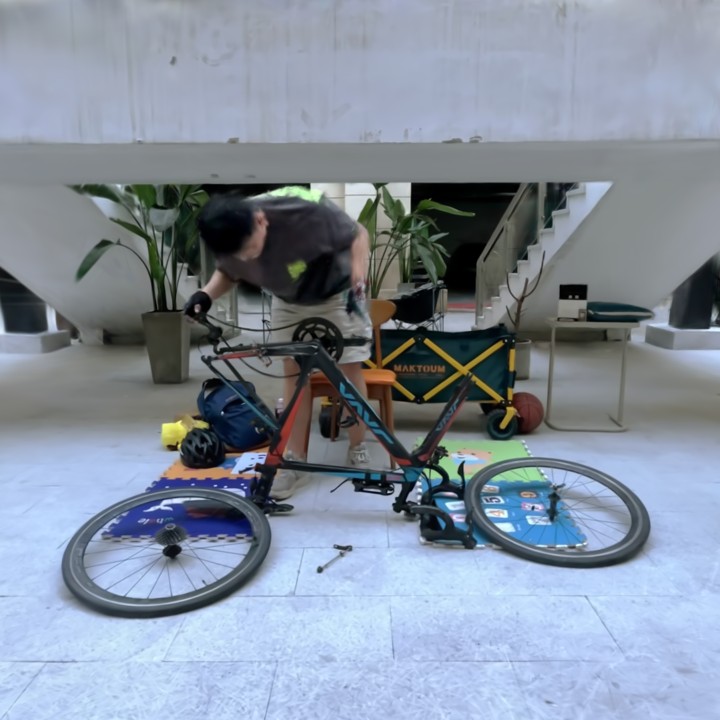}} &
\adjustbox{valign=c}{\includegraphics[width=0.22\textwidth]{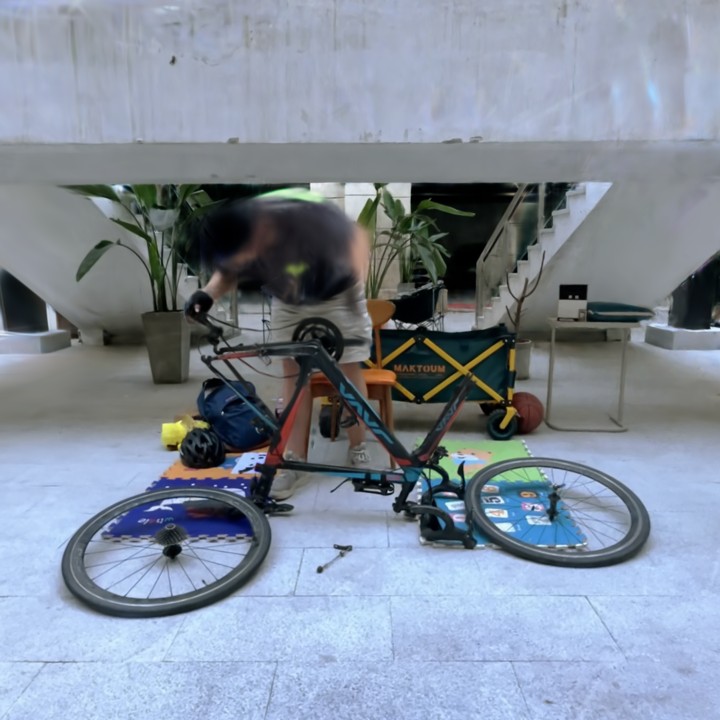}} \\
\end{tabular}

\caption{Qualitative comparison on the bike2 scene across different sequence lengths. Rows denote frame length (1200/900/600), and columns denote methods (GT, Ours, FTGS, STGS).}
\label{fig:visual_bike2_3x4}
\end{figure*}

\begin{figure*}[!t]
\centering
\setlength{\tabcolsep}{2pt}

\begin{tabular}{>{\centering\arraybackslash}m{0.03\textwidth} c c c c}
& \textbf{GT} & \textbf{Ours} & \textbf{FTGS}\cite{wang2025freetimegs} & \textbf{STGS}\cite{li2024spacetime} \\
\adjustbox{valign=c}{\rotatebox[origin=c]{90}{\scriptsize\textbf{1200}}} &
\adjustbox{valign=c}{\includegraphics[width=0.22\textwidth]{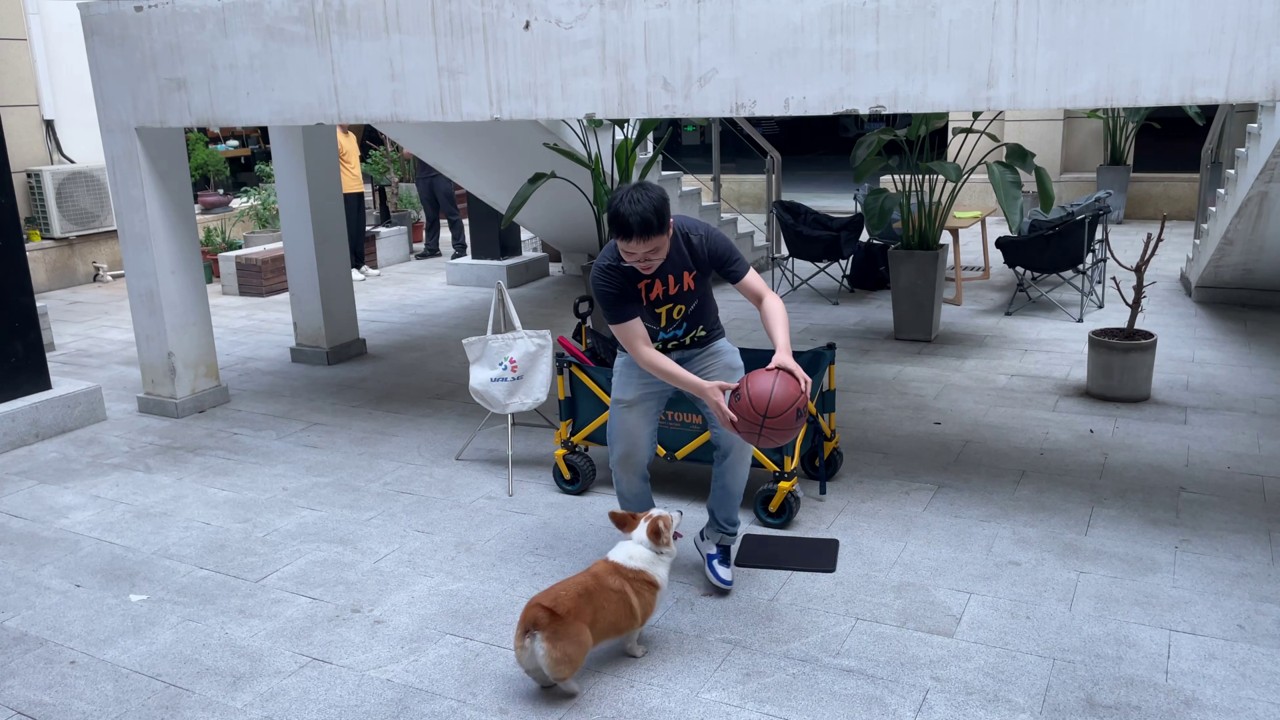}} &
\adjustbox{valign=c}{\includegraphics[width=0.22\textwidth]{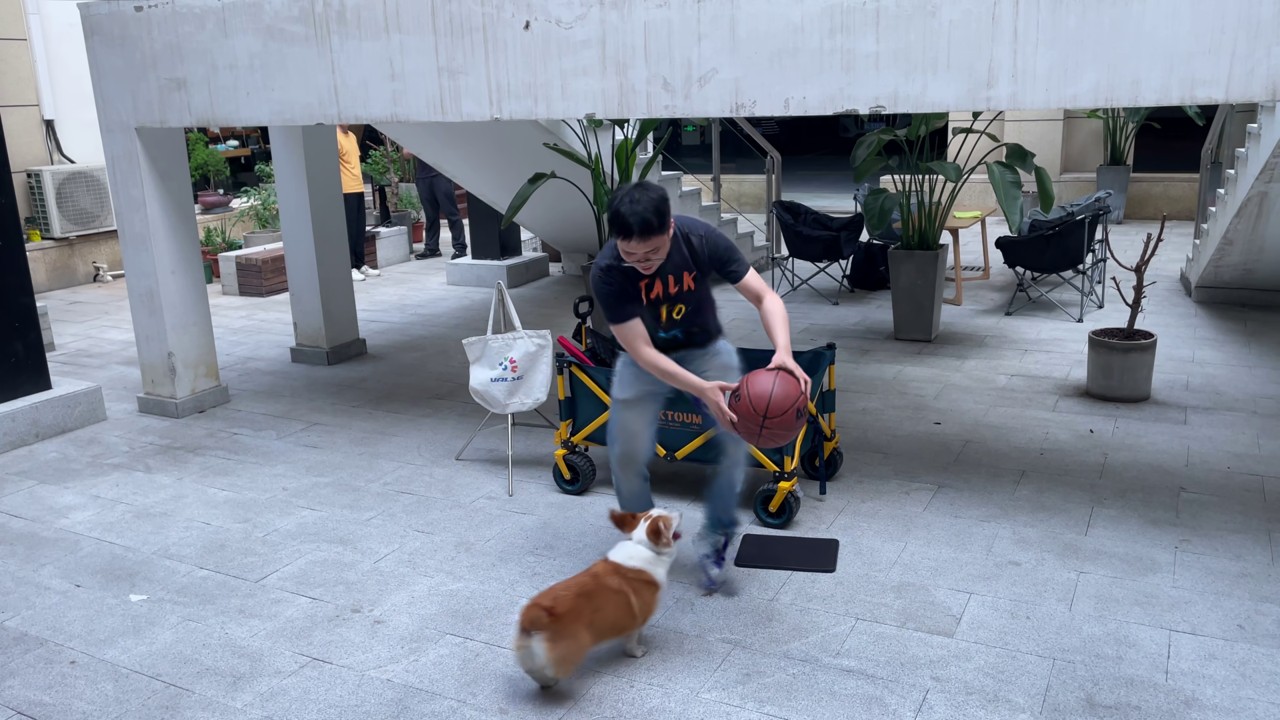}} &
\adjustbox{valign=c}{\includegraphics[width=0.22\textwidth]{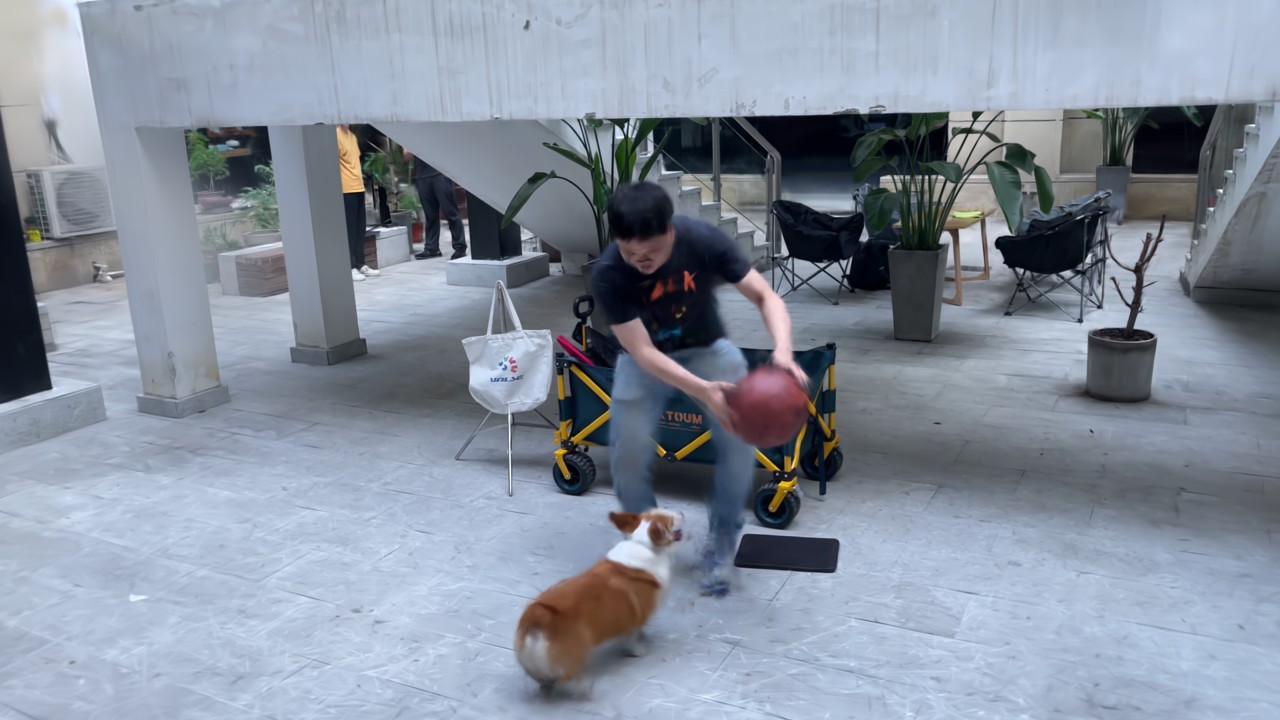}} &
\adjustbox{valign=c}{\includegraphics[width=0.22\textwidth]{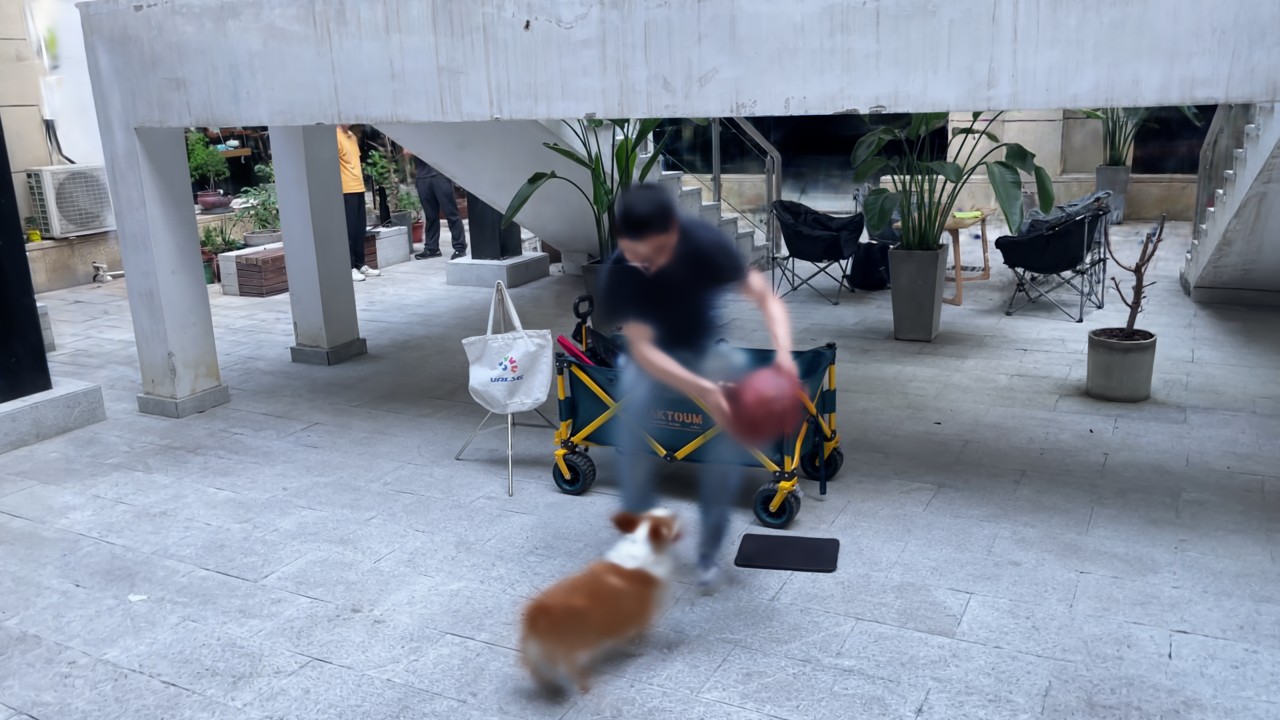}} \\
\adjustbox{valign=c}{\rotatebox[origin=c]{90}{\scriptsize\textbf{900}}} &
\adjustbox{valign=c}{\includegraphics[width=0.22\textwidth]{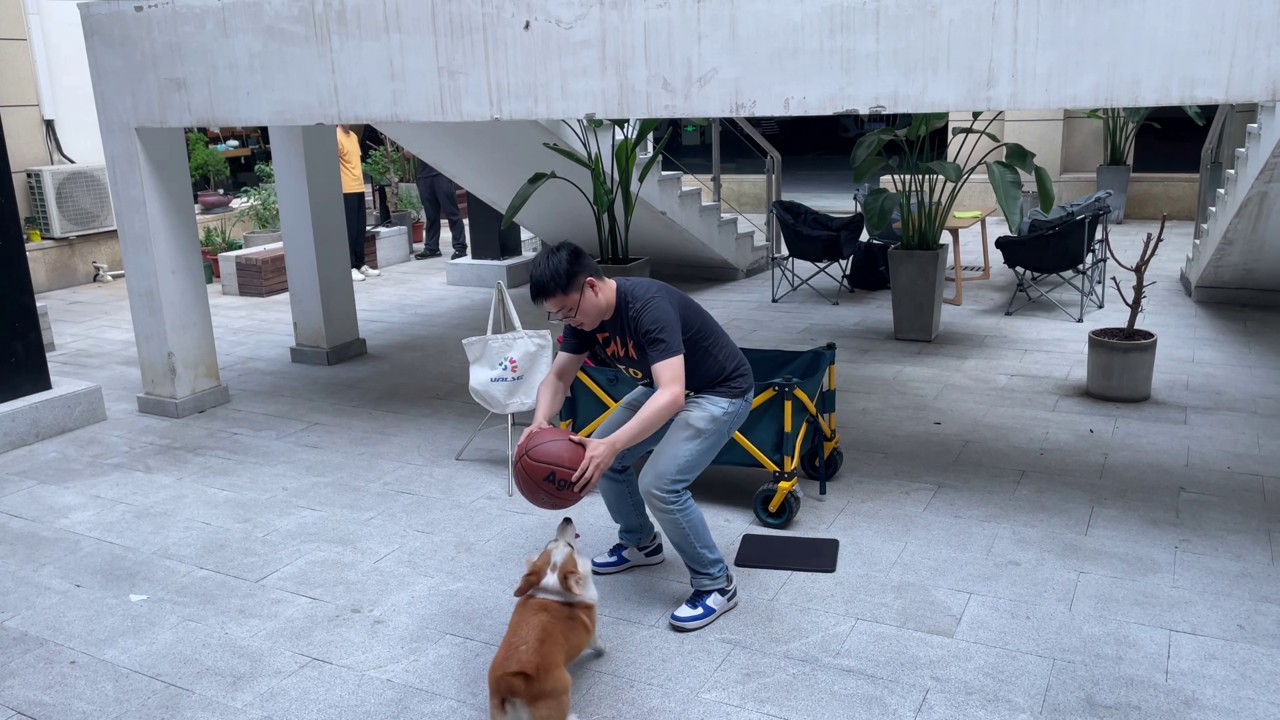}} &
\adjustbox{valign=c}{\includegraphics[width=0.22\textwidth]{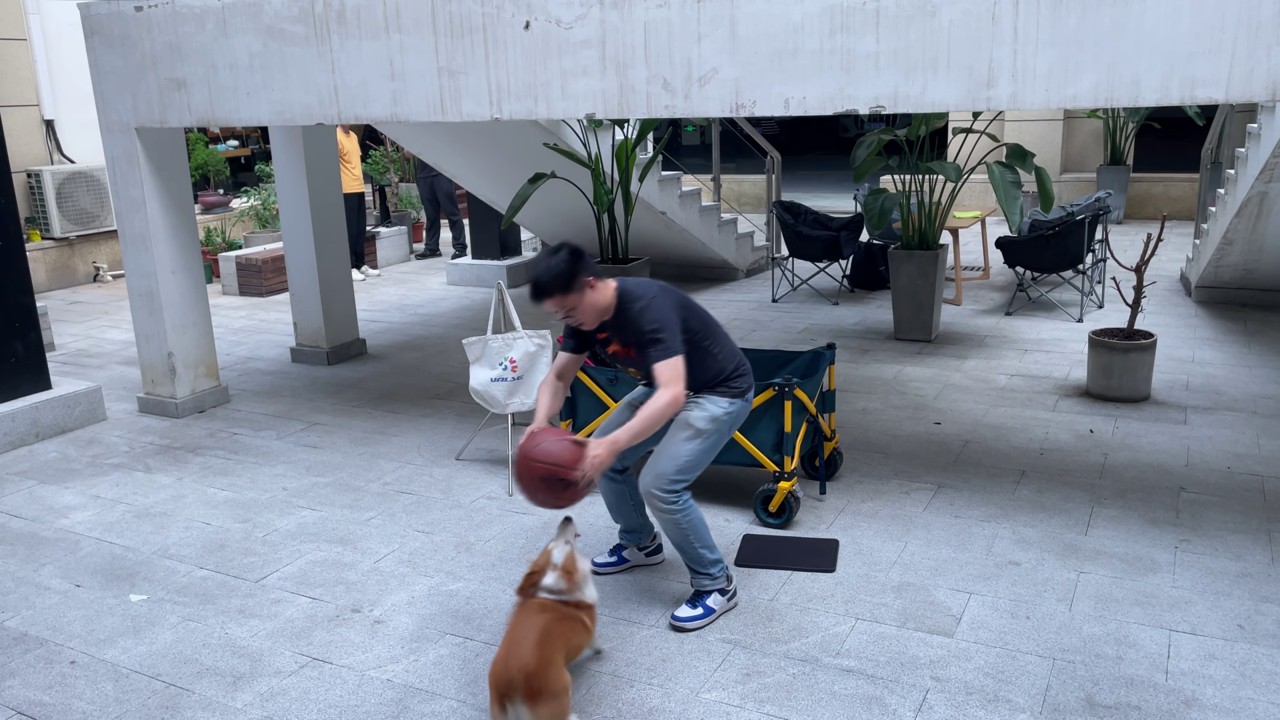}} &
\adjustbox{valign=c}{\includegraphics[width=0.22\textwidth]{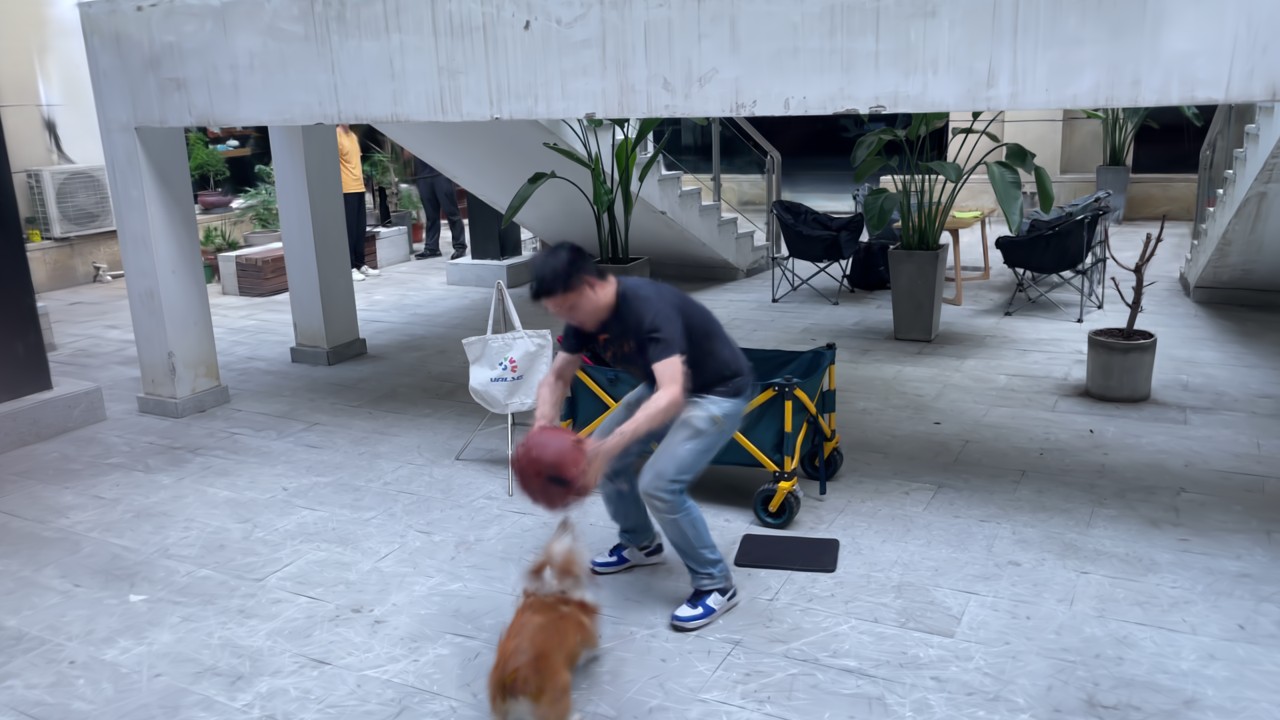}} &
\adjustbox{valign=c}{\includegraphics[width=0.22\textwidth]{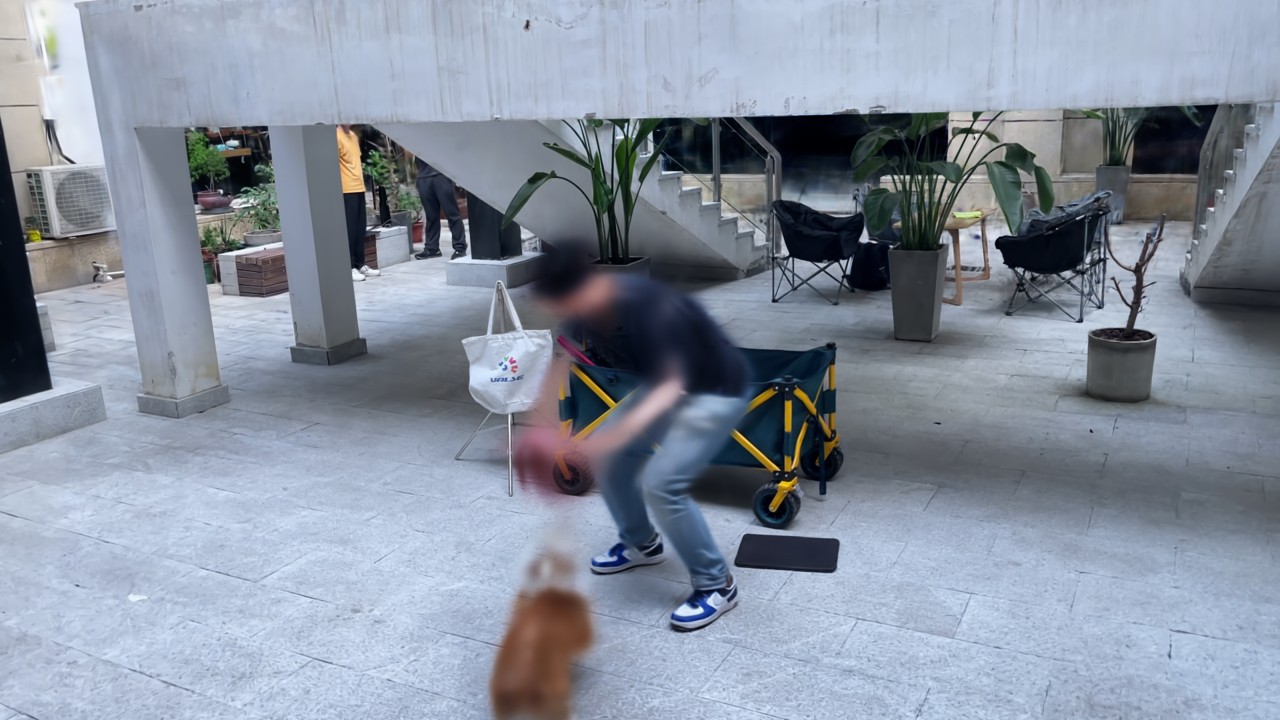}} \\
\adjustbox{valign=c}{\rotatebox[origin=c]{90}{\scriptsize\textbf{600}}} &
\adjustbox{valign=c}{\includegraphics[width=0.22\textwidth]{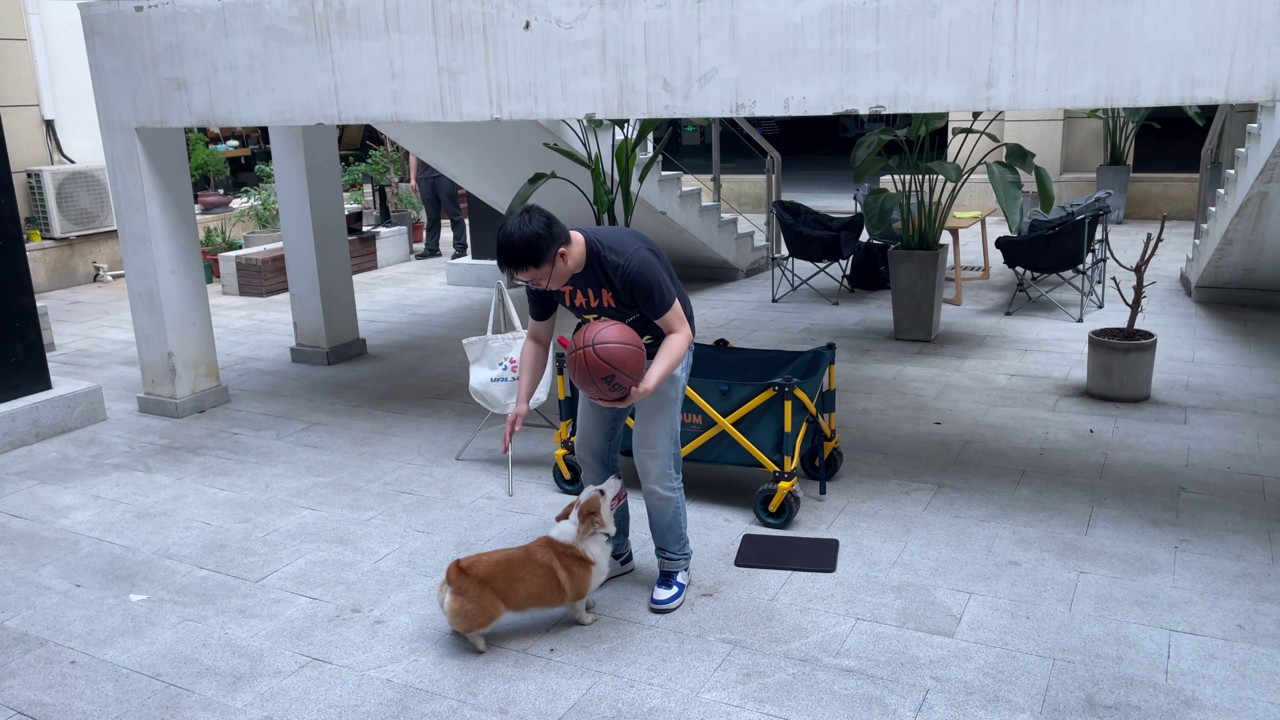}} &
\adjustbox{valign=c}{\includegraphics[width=0.22\textwidth]{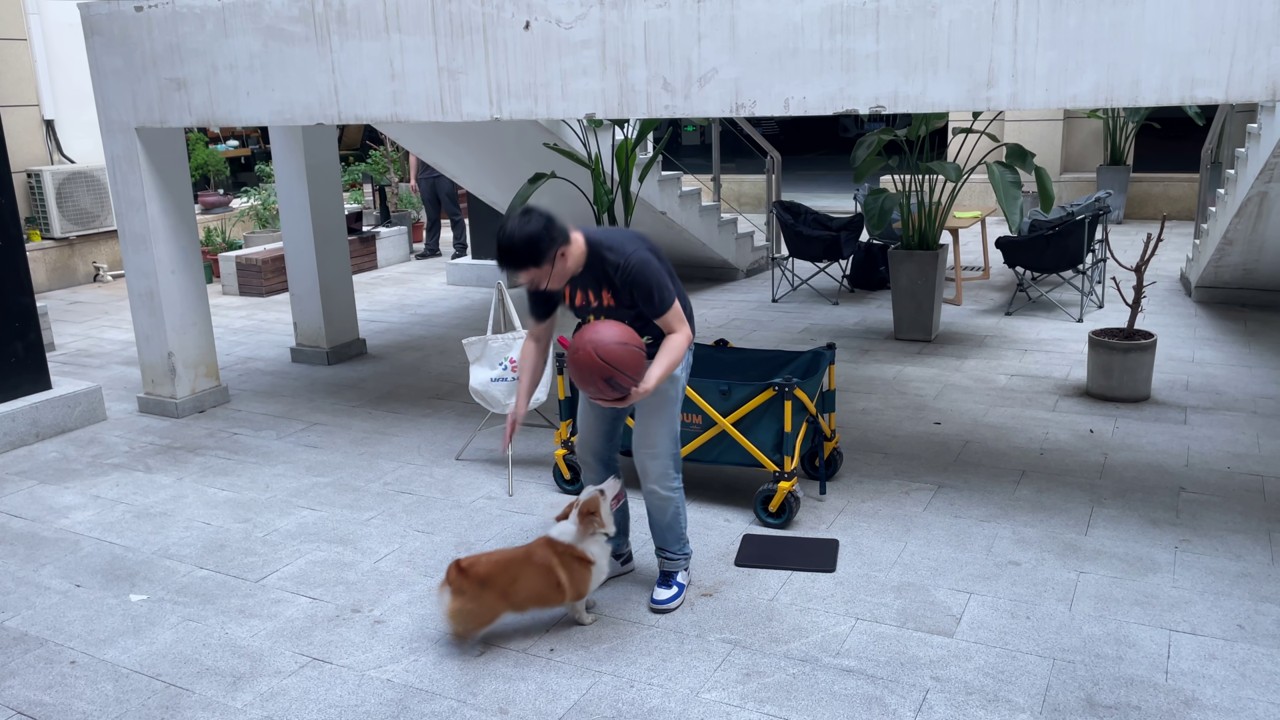}} &
\adjustbox{valign=c}{\includegraphics[width=0.22\textwidth]{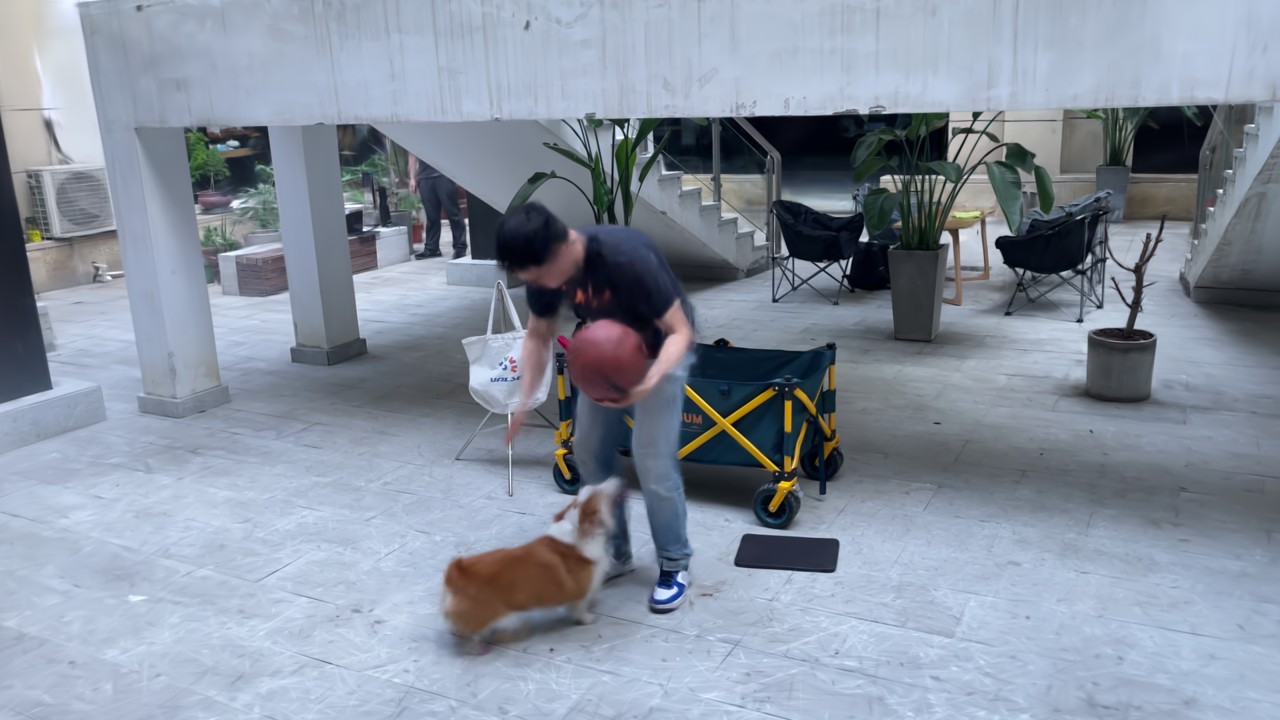}} &
\adjustbox{valign=c}{\includegraphics[width=0.22\textwidth]{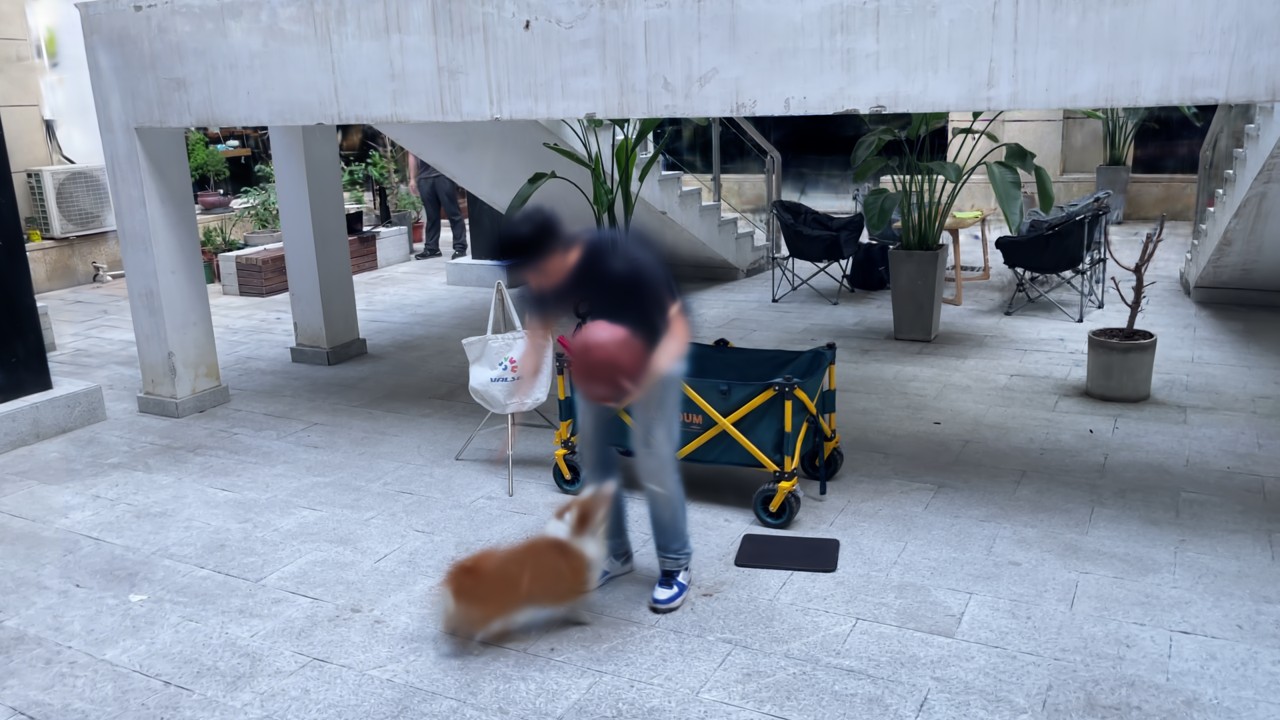}} \\
\end{tabular}

\caption{Qualitative comparison on the corgi2 scene across different sequence lengths. Rows denote frame length (1200/900/600), and columns denote methods (GT, Ours, FTGS, STGS).}
\label{fig:visual_corgi2_3x4}
\end{figure*}

\begin{figure*}[!t]
\centering
\setlength{\tabcolsep}{2pt}

\begin{tabular}{>{\centering\arraybackslash}m{0.03\textwidth} c c c c}
& \textbf{GT} & \textbf{Ours} & \textbf{FTGS}\cite{wang2025freetimegs} & \textbf{STGS}\cite{li2024spacetime} \\
\adjustbox{valign=c}{\rotatebox[origin=c]{90}{\scriptsize\textbf{1200}}} &
\adjustbox{valign=c}{\includegraphics[width=0.22\textwidth]{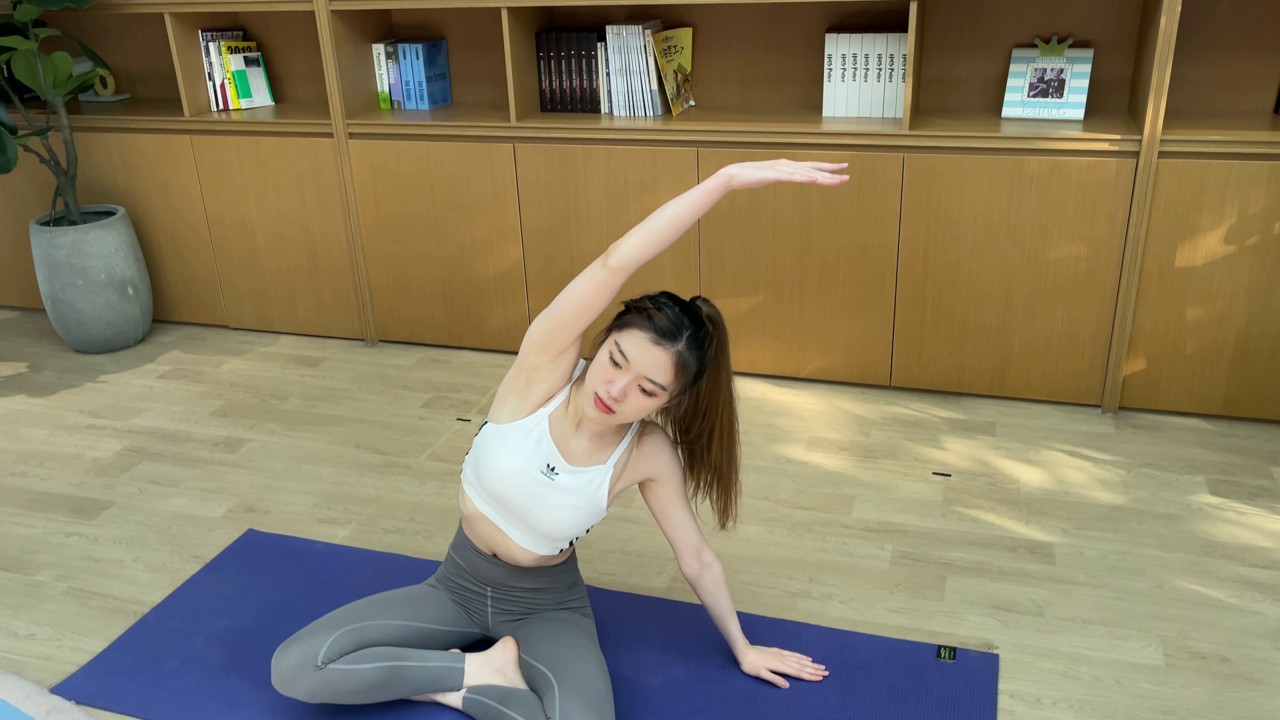}} &
\adjustbox{valign=c}{\includegraphics[width=0.22\textwidth]{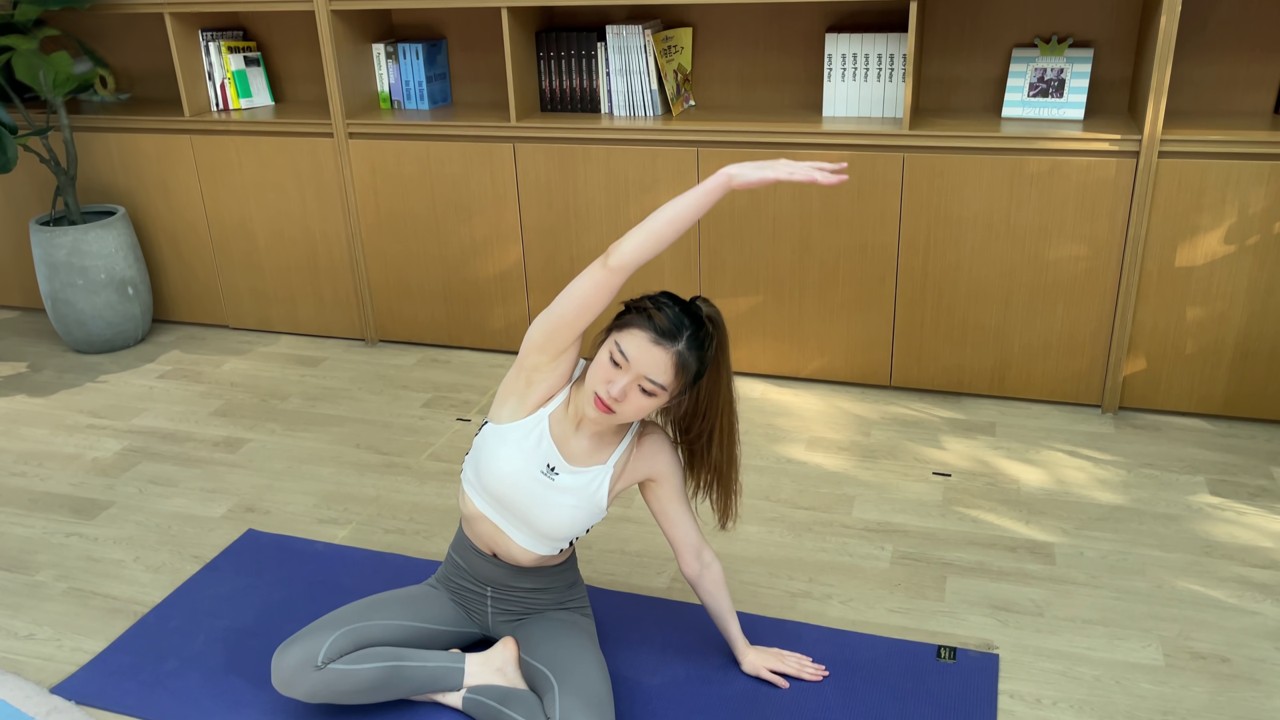}} &
\adjustbox{valign=c}{\includegraphics[width=0.22\textwidth]{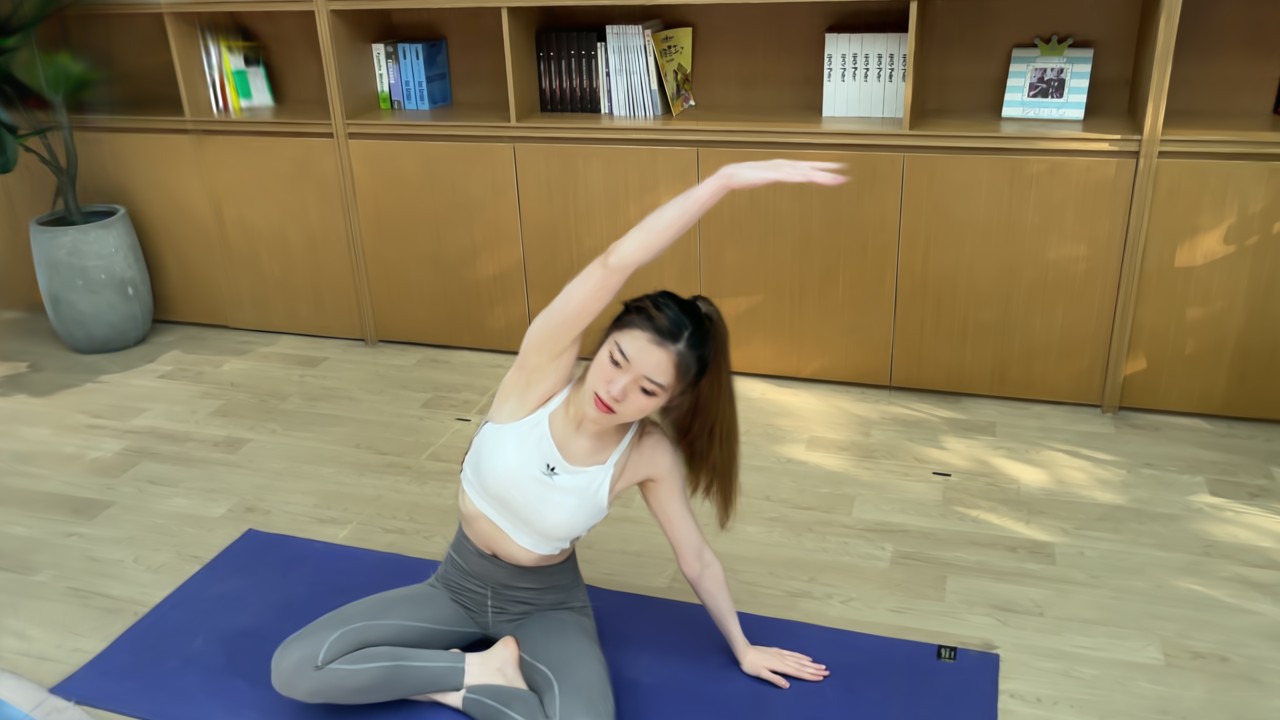}} &
\adjustbox{valign=c}{\includegraphics[width=0.22\textwidth]{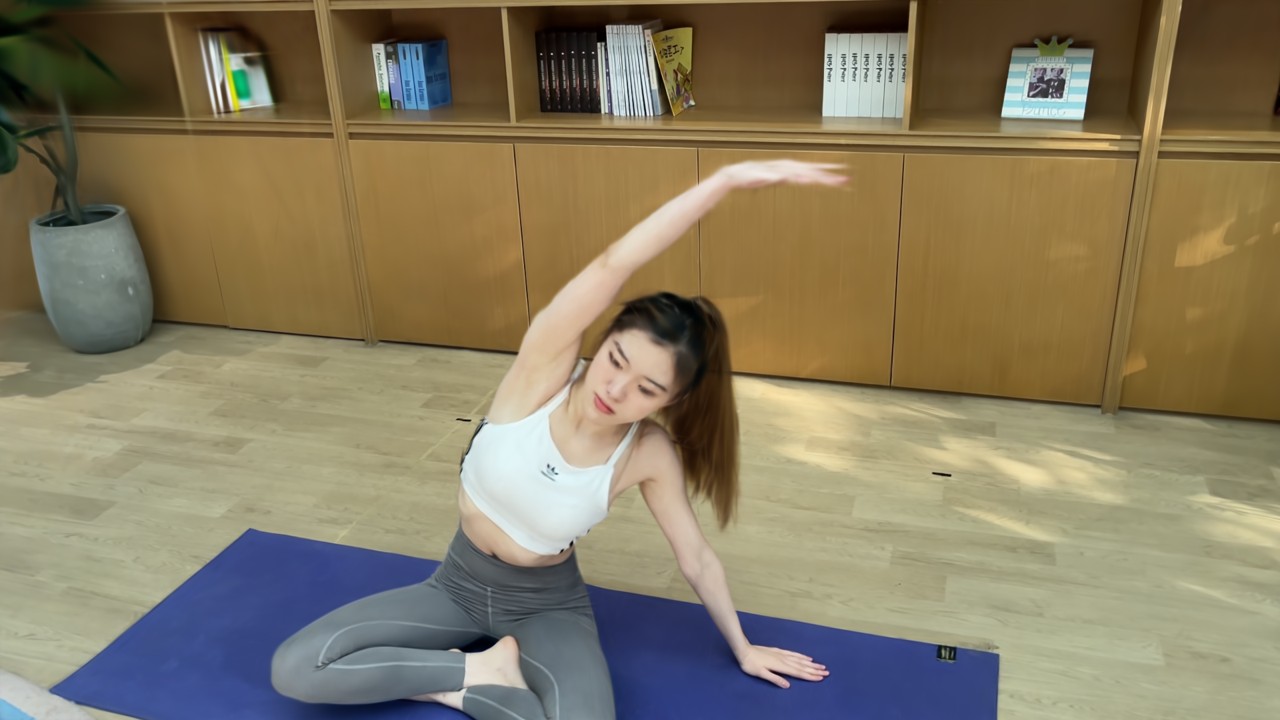}} \\
\adjustbox{valign=c}{\rotatebox[origin=c]{90}{\scriptsize\textbf{900}}} &
\adjustbox{valign=c}{\includegraphics[width=0.22\textwidth]{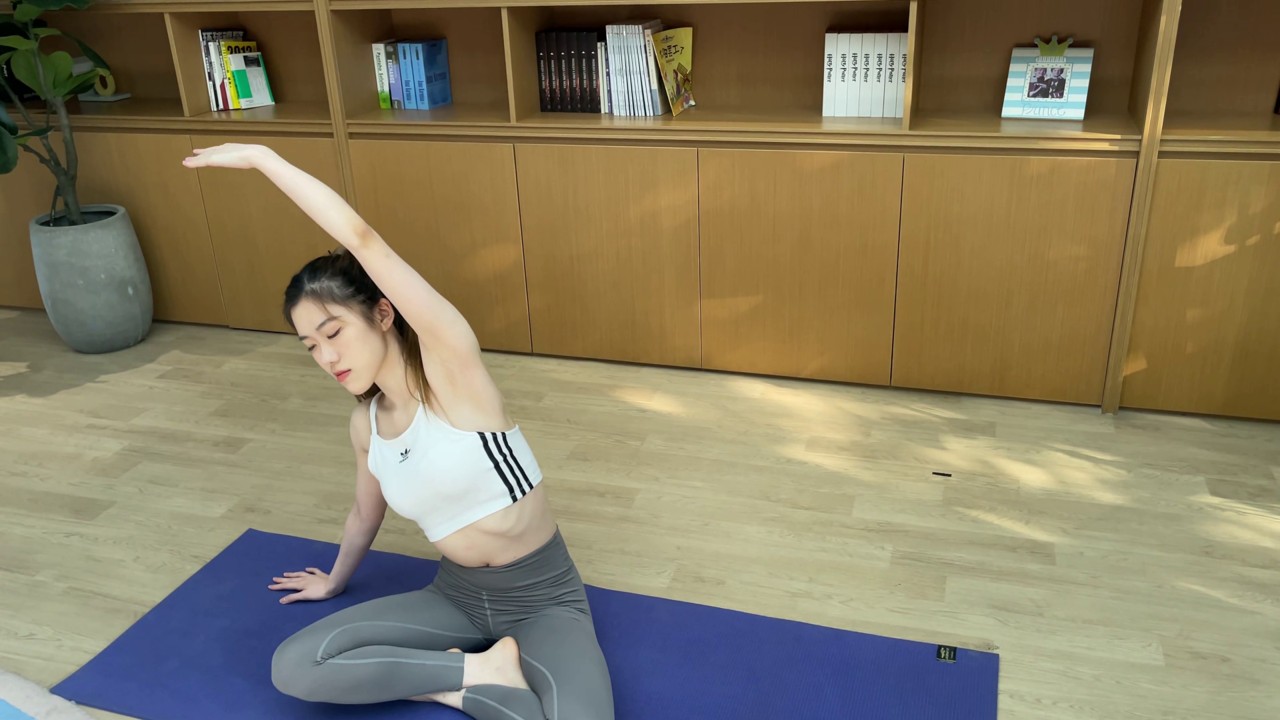}} &
\adjustbox{valign=c}{\includegraphics[width=0.22\textwidth]{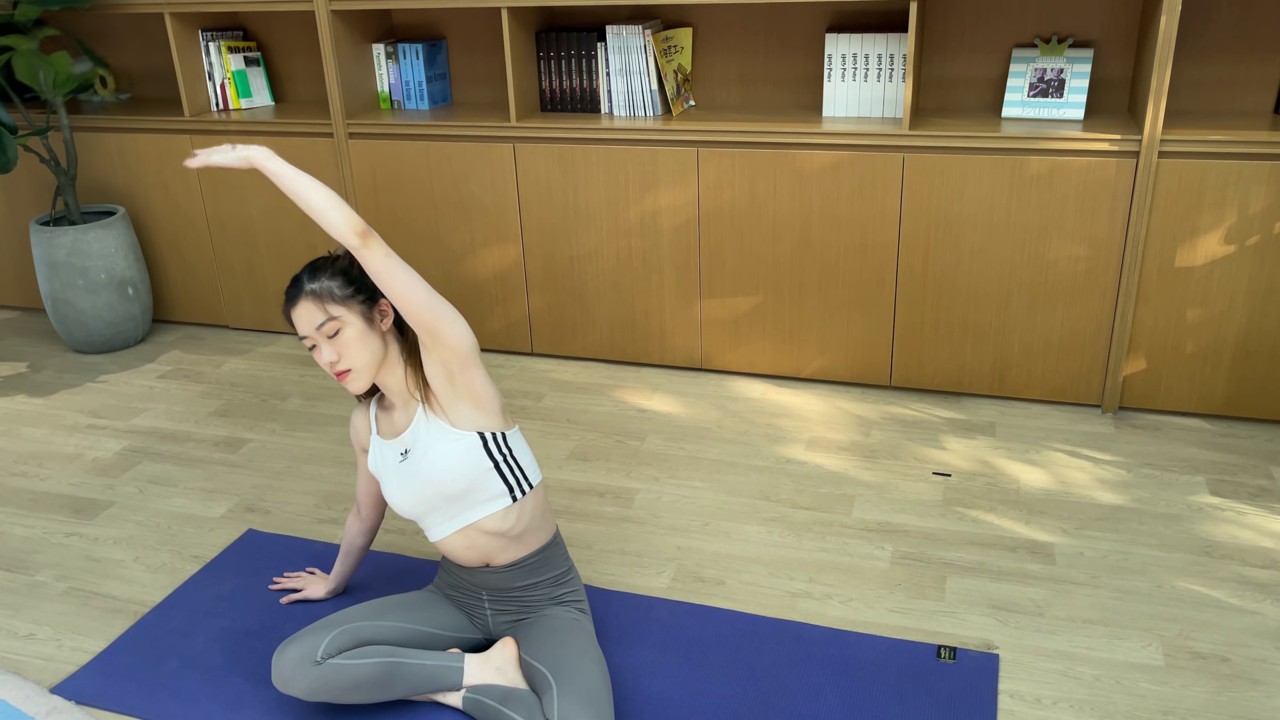}} &
\adjustbox{valign=c}{\includegraphics[width=0.22\textwidth]{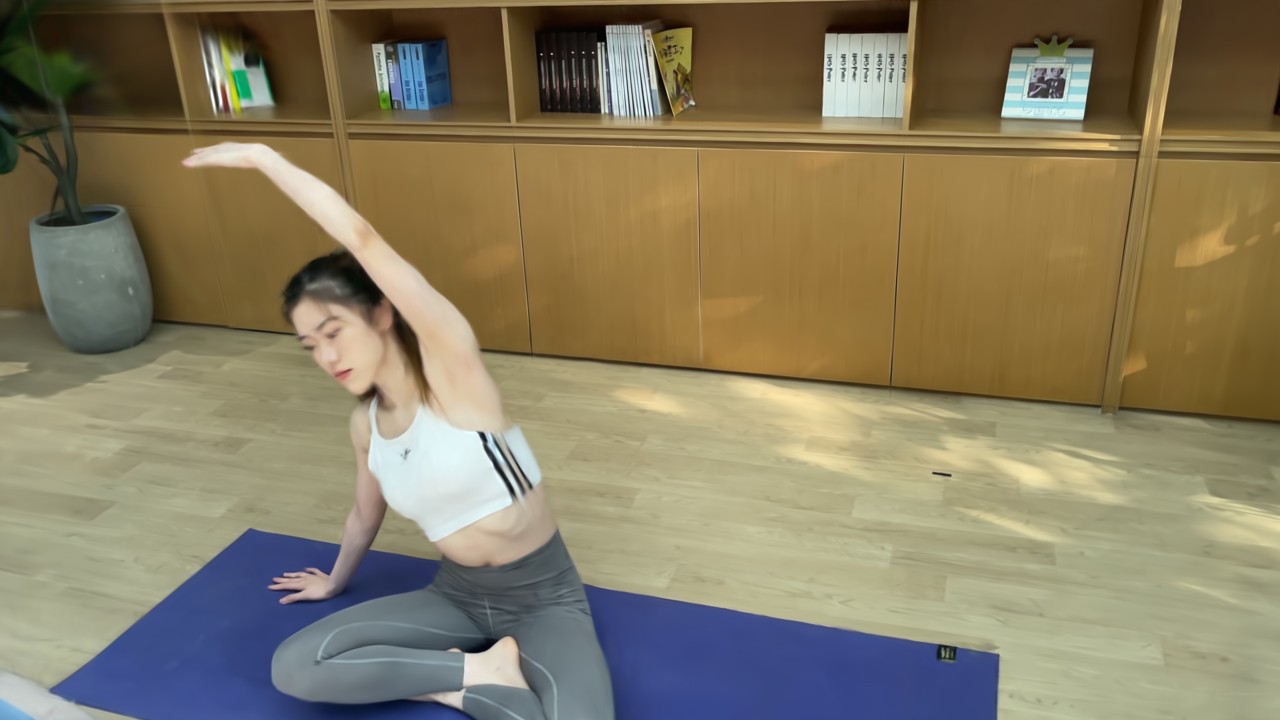}} &
\adjustbox{valign=c}{\includegraphics[width=0.22\textwidth]{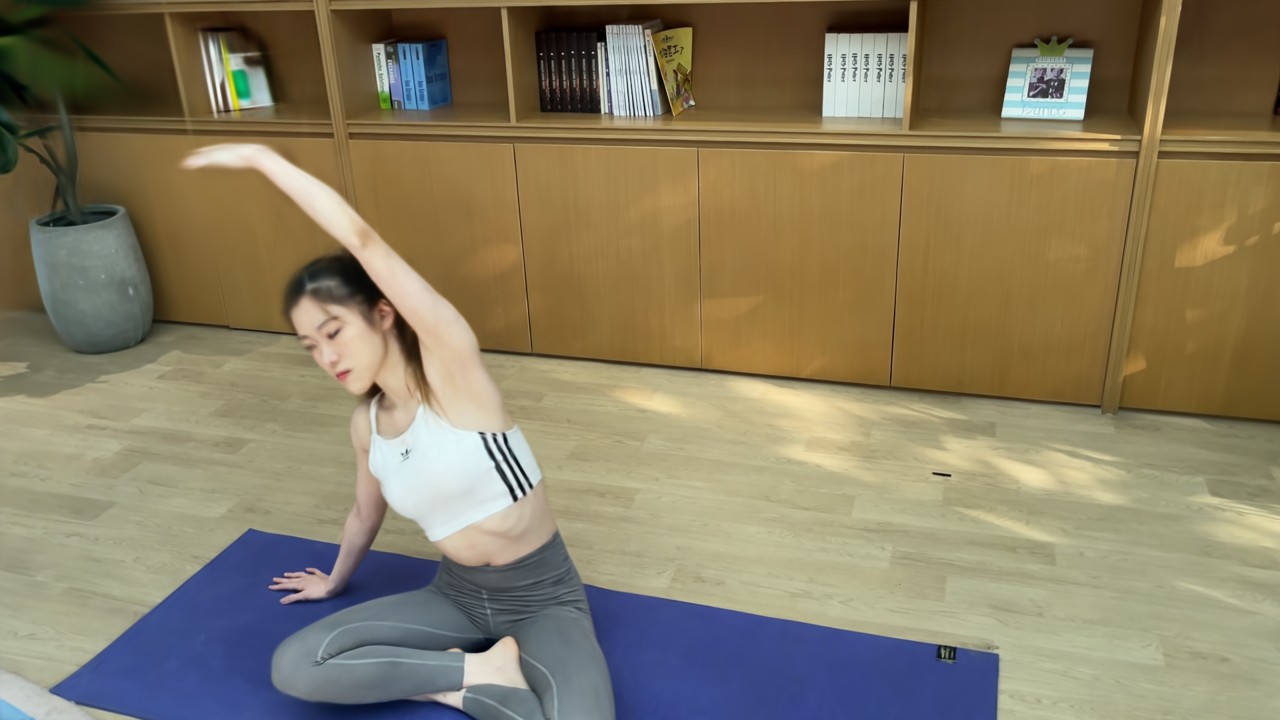}} \\
\adjustbox{valign=c}{\rotatebox[origin=c]{90}{\scriptsize\textbf{600}}} &
\adjustbox{valign=c}{\includegraphics[width=0.22\textwidth]{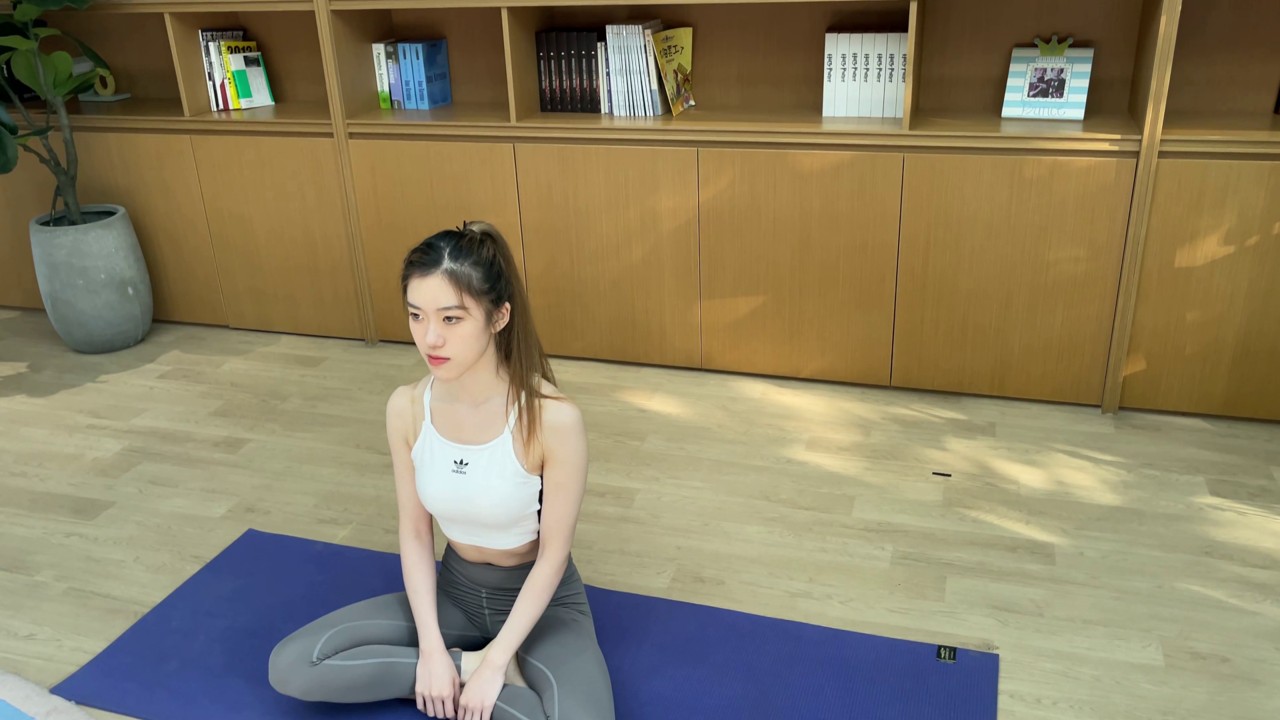}} &
\adjustbox{valign=c}{\includegraphics[width=0.22\textwidth]{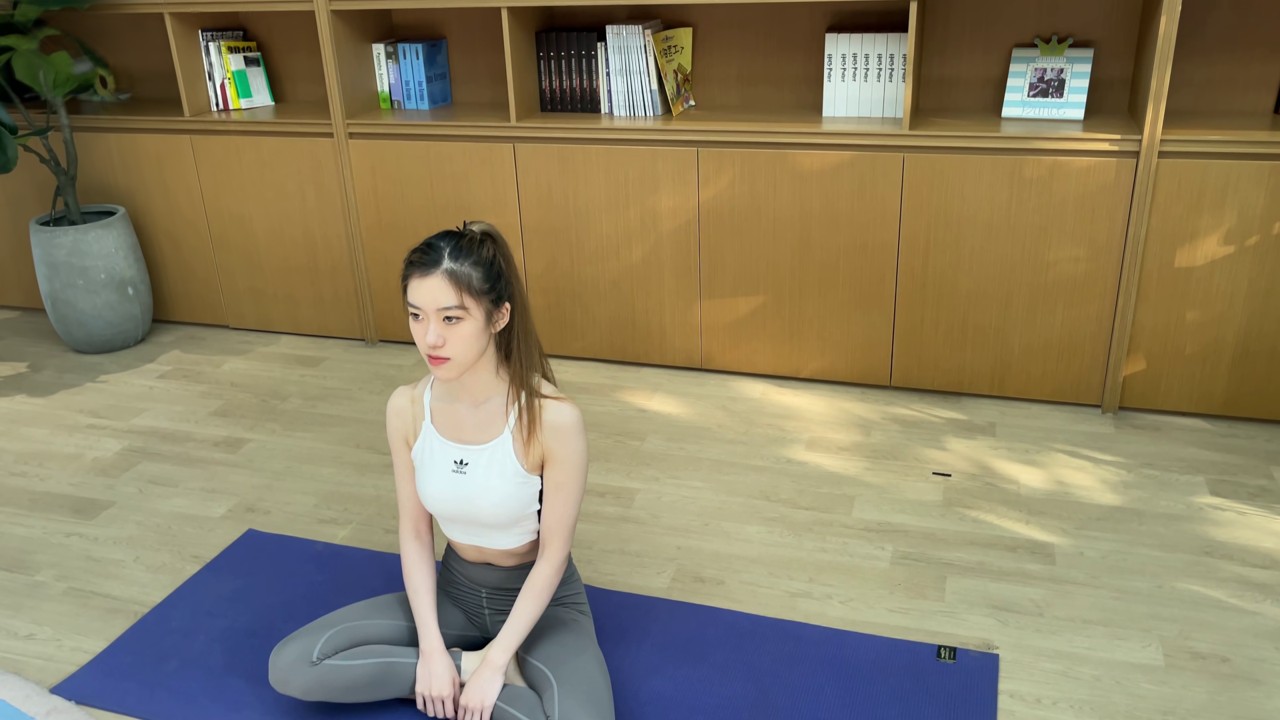}} &
\adjustbox{valign=c}{\includegraphics[width=0.22\textwidth]{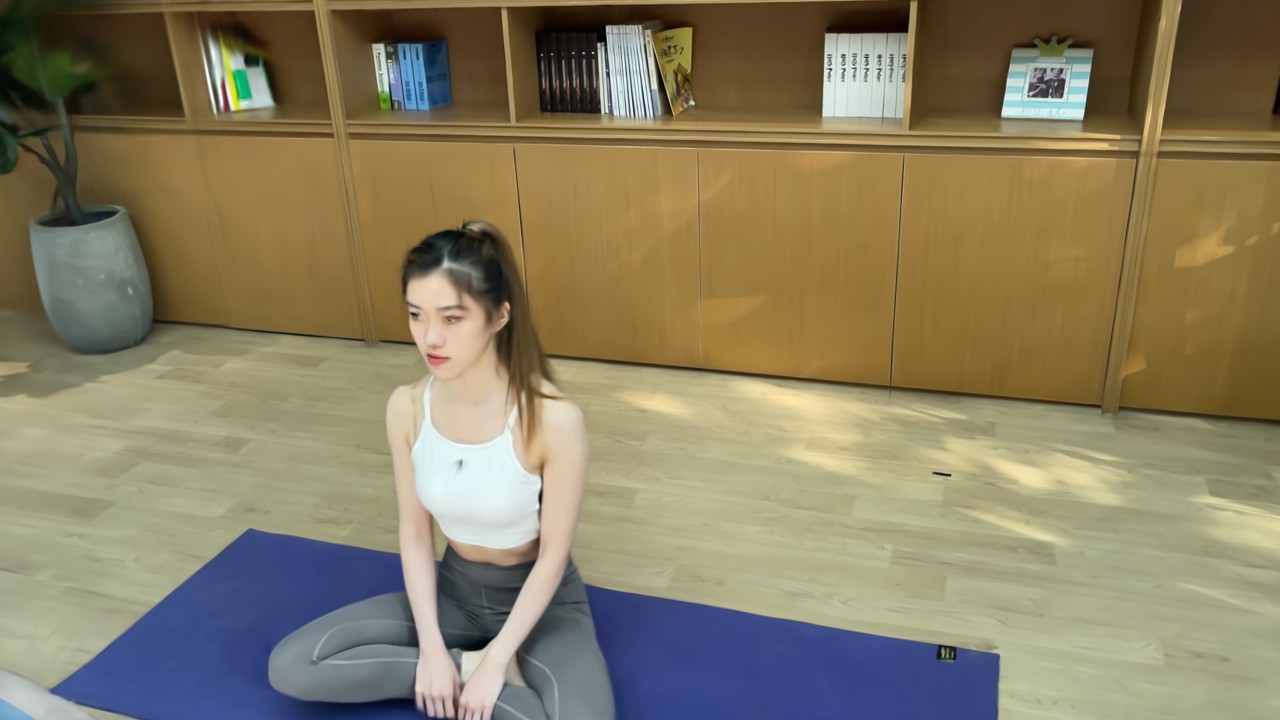}} &
\adjustbox{valign=c}{\includegraphics[width=0.22\textwidth]{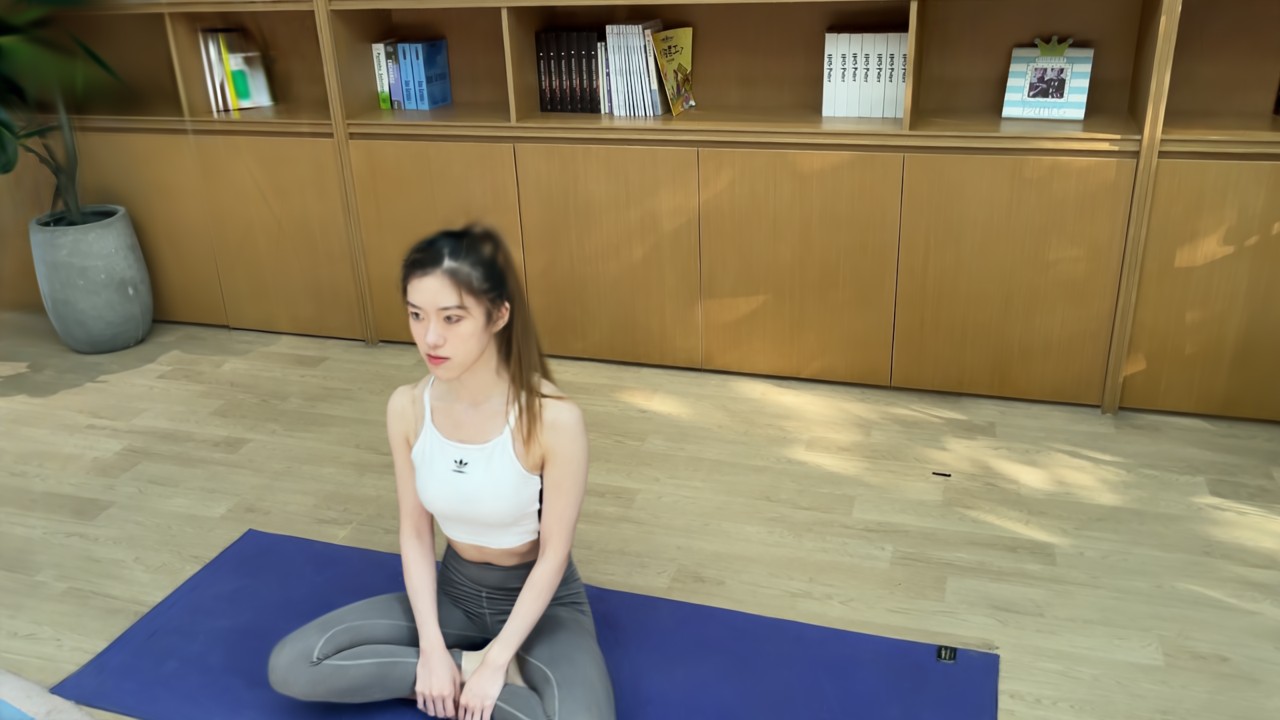}} \\
\end{tabular}

\caption{Qualitative comparison on the yoga scene across different sequence lengths. Rows denote frame length (1200/900/600), and columns denote methods (GT, Ours, FTGS, STGS).}
\label{fig:visual_yoga_3x4}
\end{figure*}

Figure~\ref{fig:visual_bike2_3x4}, Figure~\ref{fig:visual_corgi2_3x4}, and Figure~\ref{fig:visual_yoga_3x4} show qualitative comparisons for the three scenes. In general, prior methods tend to accumulate artifacts as the sequence becomes longer, including motion drift, duplicated structures, and blurred geometry. In contrast, our method maintains sharper details and more consistent geometry throughout the sequence.

\textbf{Bike2.}
The \textit{bike2} scene contains complex rigid motion with noticeable camera movement and object motion. As the sequence length increases, baseline methods gradually exhibit geometric distortions and temporal inconsistencies. In particular, fragmented motion modeling often leads to duplicated or drifting structures. Our method preserves the rigid structure of the bicycle and surrounding geometry more consistently across all frame lengths.

\textbf{Corgi2.}
The \textit{corgi2} scene involves articulated motion and fine structural details such as limbs and fur. As the sequence extends from 600 to 1200 frames, several baselines produce visible blurring and structural inconsistencies. In comparison, our approach maintains clearer object boundaries and better preserves fine-scale details, leading to more stable rendering results over time.

\textbf{Yoga.}
The \textit{yoga} scene contains significant human motion and viewpoint changes. With longer sequences, baseline methods often struggle to maintain consistent geometry, resulting in ghosting artifacts or degraded appearance. Our method produces more temporally stable reconstructions and retains sharper visual structures even at 1200 frames.

Overall, these qualitative results complement the quantitative evaluation by showing that the proposed motion formulation produces more stable and visually consistent reconstructions as the sequence length increases.


\definecolor{best}{RGB}{244,204,204}
\definecolor{second}{RGB}{252,229,205}
\definecolor{third}{RGB}{255,242,204}
\definecolor{blockgray}{RGB}{245,245,245}

\providecommand{\best}[1]{\cellcolor{best}{#1}}
\providecommand{\second}[1]{\cellcolor{second}{#1}}
\providecommand{\third}[1]{\cellcolor{third}{#1}}

\begin{table*}[t]
\centering
\small
\setlength{\tabcolsep}{5pt}
\renewcommand{\arraystretch}{1.15}
\begin{adjustbox}{max width=\textwidth}
\begin{tabular}{lccccccc}
\toprule
\textbf{600 frames} & \textbf{bike1} & \textbf{bike2} & \textbf{corgi1} & \textbf{corgi2} & \textbf{dance} & \textbf{yoga} & \textbf{avg} \\
\midrule
\rowcolor{blockgray}
\multicolumn{8}{c}{\textbf{PSNR$\uparrow$}} \\
\midrule
STGS\cite{li2024spacetime} & 26.27 & 25.55 & 23.19 & 18.93 & 26.29 & 24.78 & 24.17 \\
4DGS\cite{yang2023real} & 25.75 & 25.12 & \third{25.98} & 22.70 & 22.85 & 21.63 & 24.00 \\
Ex4DG\cite{lee2024fully} & 20.89 & 22.70 & 22.63 & 19.74 & 24.75 & 18.97 & 21.61 \\
GIFStream\cite{li2025gifstream} & \second{27.09} & \second{27.02} & 21.73 & 22.51 & 25.39 & 18.81 & 23.76 \\
LocalDyGS\cite{wu2025localdygs} & \best{27.15} & 26.19 & 24.22 & \best{24.58} & 25.60 & \third{25.24} & \third{25.49} \\
FTGS\cite{wang2025freetimegs} & 26.55 & \third{26.88} & \second{27.95} & \third{23.82} & \second{26.91} & \second{25.27} & \second{26.23} \\
\textbf{Ours} & \third{26.85} & \best{27.12} & \best{28.42} & \second{24.15} & \best{27.35} & \best{26.13} & \best{26.67} \\
\midrule
\rowcolor{blockgray}
\multicolumn{8}{c}{\textbf{SSIM$\uparrow$}} \\
\midrule
STGS\cite{li2024spacetime} & \third{0.916} & 0.916 & 0.750 & 0.656 & \third{0.921} & \third{0.907} & 0.844 \\
4DGS\cite{yang2023real} & 0.908 & 0.920 & 0.823 & 0.734 & 0.841 & 0.844 & 0.845 \\
Ex4DG\cite{lee2024fully} & 0.785 & 0.838 & 0.758 & 0.696 & 0.896 & 0.834 & 0.801 \\
GIFStream\cite{li2025gifstream} & 0.915 & \third{0.922} & \third{0.832} & \third{0.808} & 0.911 & 0.876 & \third{0.877} \\
LocalDyGS\cite{wu2025localdygs} & 0.902 & 0.910 & 0.790 & 0.770 & 0.894 & 0.885 & 0.858 \\
FTGS\cite{wang2025freetimegs} & \second{0.925} & \second{0.931} & \second{0.892} & \second{0.884} & \second{0.928} & \second{0.919} & \second{0.913} \\
\textbf{Ours} & \best{0.941} & \best{0.948} & \best{0.915} & \best{0.895} & \best{0.942} & \best{0.941} & \best{0.937} \\
\midrule
\rowcolor{blockgray}
\multicolumn{8}{c}{\textbf{LPIPS$\downarrow$}} \\
\midrule
STGS\cite{li2024spacetime} & \second{0.078} & \second{0.070} & 0.241 & 0.412 & \best{0.062} & \best{0.104} & 0.161 \\
4DGS\cite{yang2023real} & 0.137 & 0.133 & 0.270 & 0.461 & 0.262 & 0.396 & 0.276 \\
Ex4DG\cite{lee2024fully} & 0.231 & 0.185 & 0.267 & 0.350 & 0.095 & 0.271 & 0.233 \\
GIFStream\cite{li2025gifstream} & \best{0.075} & \third{0.122} & \third{0.220} & \best{0.179} & \second{0.063} & 0.148 & \third{0.134} \\
LocalDyGS\cite{wu2025localdygs} & 0.160 & 0.155 & 0.332 & 0.355 & 0.169 & 0.271 & 0.240 \\
FTGS\cite{wang2025freetimegs} & 0.095 & 0.079 & \second{0.155} & \third{0.208} & 0.084 & \second{0.105} & \second{0.121} \\
\textbf{Ours} & \third{0.082} & \best{0.065} & \best{0.138} & \second{0.195} & \third{0.078} & \third{0.120} & \best{0.113} \\
\bottomrule
\end{tabular}
\end{adjustbox}
\caption{Quantitative comparison on SelfCap (600 frames). Metrics are shown in separate blocks: PSNR, SSIM, and LPIPS. Best, second-best, and third-best are highlighted per column within each metric block.}
\label{tab:selfcap_600}
\end{table*}

\definecolor{best}{RGB}{244,204,204}
\definecolor{second}{RGB}{252,229,205}
\definecolor{third}{RGB}{255,242,204}
\definecolor{blockgray}{RGB}{245,245,245}

\providecommand{\best}[1]{\cellcolor{best}{#1}}
\providecommand{\second}[1]{\cellcolor{second}{#1}}
\providecommand{\third}[1]{\cellcolor{third}{#1}}
\providecommand{\besttext}[1]{\colorbox{best}{\strut #1}}
\providecommand{\secondtext}[1]{\colorbox{second}{\strut #1}}
\providecommand{\thirdtext}[1]{\colorbox{third}{\strut #1}}

\begin{table*}[t]
\centering
\small
\setlength{\tabcolsep}{8pt}
\renewcommand{\arraystretch}{1.15}
\begin{adjustbox}{max width=\textwidth}
\begin{tabular}{lccccccc}
\toprule
\textbf{900 frames} & \textbf{bike1} & \textbf{bike2} & \textbf{corgi1} & \textbf{corgi2} & \textbf{dance} & \textbf{yoga} & \textbf{avg} \\
\midrule
\rowcolor{blockgray}
\multicolumn{8}{c}{\textbf{PSNR$\uparrow$}} \\
\midrule
STGS\cite{li2024spacetime}      & 26.37 & 25.56 & 26.27 & 14.14 & \third{25.90} & 19.84 & 23.01 \\
4DGS\cite{yang2023real}      & 25.48 & 25.00 & 22.18 & 21.77 & 22.25 & 21.05 & 22.95 \\
Ex4DG\cite{lee2024fully}     & 20.38 & 22.38 & 21.47 & 19.74 & 23.00 & 18.62 & 20.93 \\
GIFStream\cite{li2025gifstream} & \best{26.76} & 26.29 & \third{26.83} & 22.54 & 24.05 & 16.38 & \third{23.81} \\
LocalDyGS\cite{wu2025localdygs} & 26.45 & \third{26.34} & 25.39 & \best{24.53} & 14.97 & \third{23.41} & 23.51 \\
FTGS\cite{wang2025freetimegs}      & \third{26.45} & \second{26.60} & \second{27.36} & \third{23.13} & \second{26.43} & \second{25.19} & \second{25.86} \\
\textbf{Ours} & \second{26.70} & \best{26.87} & \best{27.85} & \second{23.88} & \best{26.78} & \best{25.84} & \best{26.32} \\
\midrule
\rowcolor{blockgray}
\multicolumn{8}{c}{\textbf{SSIM$\uparrow$}} \\
\midrule
STGS\cite{li2024spacetime}      & \third{0.913} & \third{0.922} & \second{0.854} & 0.698 & \third{0.919} & \third{0.882} & 0.864 \\
4DGS\cite{yang2023real}      & 0.904 & 0.916 & 0.742 & 0.730 & 0.835 & 0.841 & 0.828 \\
Ex4DG\cite{lee2024fully}     & 0.775 & 0.834 & 0.750 & 0.690 & 0.882 & 0.826 & 0.793 \\
GIFStream\cite{li2025gifstream} & 0.905 & 0.913 & \third{0.843} & \third{0.810} & 0.903 & 0.852 & \third{0.871} \\
LocalDyGS\cite{wu2025localdygs} & 0.898 & 0.915 & 0.790 & 0.768 & 0.784 & 0.874 & 0.838 \\
FTGS\cite{wang2025freetimegs}      & \second{0.919} & \second{0.920} & 0.840 & \second{0.825} & \second{0.918} & \second{0.866} & \second{0.898} \\
\textbf{Ours} & \best{0.924} & \best{0.925} & \best{0.912} & \best{0.889} & \best{0.926} & \best{0.920} & \best{0.916} \\
\midrule
\rowcolor{blockgray}
\multicolumn{8}{c}{\textbf{LPIPS$\downarrow$}} \\
\midrule
STGS\cite{li2024spacetime}      & \best{0.080} & \best{0.066} & \second{0.126} & 0.326 & \best{0.068} & \second{0.106} & \second{0.124} \\
4DGS\cite{yang2023real}      & 0.148 & 0.125 & 0.455 & 0.465 & 0.276 & 0.428 & 0.316 \\
Ex4DG\cite{lee2024fully}     & 0.243 & 0.190 & 0.285 & 0.372 & 0.114 & 0.288 & 0.248 \\
GIFStream\cite{li2025gifstream} & \second{0.084} & \third{0.076} & \third{0.157} & \best{0.180} & \second{0.072} & 0.175 & \third{0.132} \\
LocalDyGS\cite{wu2025localdygs} & 0.178 & 0.147 & 0.333 & 0.355 & 0.386 & 0.284 & 0.280 \\
FTGS\cite{wang2025freetimegs}      & 0.093 & 0.081 & 0.202 & \third{0.213} & \third{0.097} & \best{0.106} & 0.132 \\
\textbf{Ours} & \third{0.089} & \second{0.073} & \best{0.125} & \second{0.185} & 0.104 & \third{0.144} & \best{0.120} \\
\bottomrule
\end{tabular}
\end{adjustbox}
\caption{Quantitative comparison on SelfCap (900 frames). \besttext{Best}, \secondtext{second-best}, and \thirdtext{third-best} results are highlighted.}
\label{tab:selfcap_900}
\end{table*}

\definecolor{best}{RGB}{244,204,204}
\definecolor{second}{RGB}{252,229,205}
\definecolor{third}{RGB}{255,242,204}
\definecolor{blockgray}{RGB}{245,245,245}

\providecommand{\best}[1]{\cellcolor{best}{#1}}
\providecommand{\second}[1]{\cellcolor{second}{#1}}
\providecommand{\third}[1]{\cellcolor{third}{#1}}

\begin{table*}[t]
\centering
\small
\setlength{\tabcolsep}{8pt}
\renewcommand{\arraystretch}{1.15}
\begin{adjustbox}{max width=\textwidth}
\begin{tabular}{lccccccc}
\toprule
\textbf{1200 frames} & \textbf{bike1} & \textbf{bike2} & \textbf{corgi1} & \textbf{corgi2} & \textbf{dance} & \textbf{yoga} & \textbf{avg} \\
\midrule
\rowcolor{blockgray}
\multicolumn{8}{c}{\textbf{PSNR$\uparrow$}} \\
\midrule
STGS\cite{li2024spacetime}      & \third{26.42} & \best{27.27} & \third{24.85} & \best{25.30} & 24.06 & \third{24.16} & \third{25.34} \\
4DGS\cite{yang2023real}      & 25.08 & 25.04 & 22.14 & 21.59 & 21.13 & 21.44 & 22.74 \\
Ex4DG\cite{lee2024fully}     & 20.31 & 22.37 & 21.14 & 19.42 & 21.94 & 17.54 & 20.45 \\
GIFStream\cite{li2025gifstream} & \best{26.54} & 23.42 & 24.74 & 22.98 & \third{25.97} & 22.00 & 24.27 \\
LocalDyGS\cite{wu2025localdygs} & 25.08 & 25.81 & 23.61 & \third{23.44} & 24.00 & 23.49 & 24.24 \\
FTGS\cite{wang2025freetimegs}       & 25.82 & \third{26.15} & \second{25.54} & 23.41 & \second{26.30} & \second{25.23} & \second{25.41} \\
\textbf{Ours} & \second{26.45} & \second{26.70} & \best{26.25} & \second{24.62} & \best{26.85} & \best{25.43} & \best{26.05} \\
\midrule
\rowcolor{blockgray}
\multicolumn{8}{c}{\textbf{SSIM$\uparrow$}} \\
\midrule
STGS\cite{li2024spacetime}      & \best{0.914} & \best{0.924} & \third{0.831} & \second{0.836} & 0.734 & \second{0.899} & 0.734 \\
4DGS\cite{yang2023real}      & 0.906 & 0.916 & 0.740 & 0.728 & 0.822 & 0.838 & 0.825 \\
Ex4DG\cite{lee2024fully}     & 0.772 & 0.834 & 0.742 & 0.691 & 0.871 & 0.814 & 0.787 \\
GIFStream\cite{li2025gifstream} & \third{0.907} & 0.893 & 0.827 & 0.815 & \second{0.911} & \third{0.889} & \second{0.874} \\
LocalDyGS\cite{wu2025localdygs} & 0.854 & \third{0.904} & 0.745 & 0.731 & 0.876 & 0.857 & 0.828 \\
FTGS\cite{wang2025freetimegs}       & 0.895 & 0.902 & \second{0.824} & \third{0.812} & \third{0.914} & 0.879 & \third{0.871} \\
\textbf{Ours} & \second{0.912} & \second{0.918} & \best{0.885} & \best{0.875} & \best{0.931} & \best{0.903} & \best{0.904} \\
\midrule
\rowcolor{blockgray}
\multicolumn{8}{c}{\textbf{LPIPS$\downarrow$}} \\
\midrule
STGS\cite{li2024spacetime}      & \best{0.079} & \best{0.067} & 0.188 & \best{0.148} & 0.100 & \second{0.120} & \second{0.100} \\
4DGS\cite{yang2023real}      & 0.145 & 0.133 & 0.452 & 0.466 & 0.292 & 0.407 & 0.316 \\
Ex4DG\cite{lee2024fully}     & 0.243 & 0.187 & 0.290 & 0.346 & 0.126 & 0.299 & 0.248 \\
GIFStream\cite{li2025gifstream} & \third{0.084} & \third{0.083} & \second{0.167} & \third{0.170} & \best{0.069} & \third{0.130} & \third{0.117} \\
LocalDyGS\cite{wu2025localdygs} & 0.219 & 0.153 & 0.396 & 0.424 & 0.191 & 0.344 & 0.288 \\
FTGS\cite{wang2025freetimegs}       & 0.095 & 0.088 & \third{0.182} & 0.224 & \third{0.092} & 0.195 & 0.146 \\
\textbf{Ours} & \second{0.082} & \second{0.068} & \best{0.121} & \second{0.154} & \second{0.075} & \best{0.094} & \best{0.099} \\
\bottomrule
\end{tabular}
\end{adjustbox}
\caption{Quantitative comparison on SelfCap (1200 frames). \besttext{Best}, \secondtext{second-best}, and \thirdtext{third-best} results are highlighted.}
\label{tab:selfcap_1200}
\end{table*}
\label{sec:suppl_Experiments}
\section{Mathematical Formulations of TRiGS}
\label{sec:suppl_Formulations}
This section provides additional details on the closed-form $SE(3)$ exponential map, the gauge-fixing argument in Sec.~3.3, and the small-angle-stable implementation.

\subsection{Closed-form \texorpdfstring{$SE(3)$}{SE(3)} exponential map}
\label{supp:closed_form_se3}

In the main paper, the motion of each primitive is parameterized by a time-conditioned Lie algebra coefficient
\begin{equation}
\zeta_i(t)=(\omega_i(t),\nu_i(t))\in\mathfrak{se}(3),
\end{equation}
and the corresponding relative log-parameter is defined as
\begin{equation}
\mathbf{u}_i(t):=\zeta_i(t)\Delta t,
\qquad
\Delta t:=t-\mu_{t,i}.
\end{equation}
Writing
\begin{equation}
\mathbf{u}_i(t)=(\phi_i(t),\upsilon_i(t)),
\qquad
\phi_i(t):=\omega_i(t)\Delta t,
\qquad
\upsilon_i(t):=\nu_i(t)\Delta t,
\end{equation}
the relative rigid transform from the central time $\mu_{t,i}$ to the query time $t$ is given by
\begin{equation}
\mathbf{T}_i(\mu_{t,i}\!\to\! t)
=
\exp\!\bigl(\hat{\mathbf{u}}_i(t)\bigr)
=
\begin{bmatrix}
R_i(t) & p_i(t)\\
0 & 1
\end{bmatrix}.
\end{equation}
Here, $\hat{\mathbf{u}}_i(t)$ denotes the standard wedge operator in $\mathfrak{se}(3)$:
\begin{equation}
\hat{\mathbf{u}}_i(t)=
\begin{bmatrix}
\hat\phi_i(t) & \upsilon_i(t)\\
0 & 0
\end{bmatrix},
\end{equation}
where, for any $a=[a_1,a_2,a_3]^\top\in\mathbb{R}^3$, its hat representation in $\mathfrak{so}(3)$ is
\begin{equation}
\hat a=
\begin{bmatrix}
0 & -a_3 & a_2\\
a_3 & 0 & -a_1\\
-a_2 & a_1 & 0
\end{bmatrix}.
\end{equation}
Let $\Omega:=\hat\phi_i(t)$ and $v:=\upsilon_i(t)$. For $n\ge1$, we have
\begin{equation}
\hat{\mathbf{u}}_i(t)^n=
\begin{bmatrix}
\Omega^n & \Omega^{n-1}v\\
0 & 0
\end{bmatrix}.
\end{equation}
Therefore, by the matrix exponential series,
\begin{equation}
\exp\!\bigl(\hat{\mathbf{u}}_i(t)\bigr)
=
I+\sum_{n=1}^{\infty}\frac{\hat{\mathbf{u}}_i(t)^n}{n!}
=
\begin{bmatrix}
I+\sum_{n=1}^{\infty}\frac{\Omega^n}{n!}
&
\sum_{n=1}^{\infty}\frac{\Omega^{n-1}}{n!}v\\
0 & 1
\end{bmatrix}.
\end{equation}
Hence, the rotation and translation blocks are
\begin{equation}
R_i(t)=\exp(\hat\phi_i(t)),
\qquad
p_i(t)
=
\left(\sum_{n=0}^{\infty}\frac{\hat\phi_i(t)^n}{(n+1)!}\right)\upsilon_i(t)
=
J(\phi_i(t))\,\upsilon_i(t),
\tag{S1}
\label{eq:supp_translation_jacobian}
\end{equation}
where $J(\phi_i(t))$ is the left Jacobian of $SO(3)$. Thus, the translation is induced by the same Lie algebra element as the rotation.
Let $\theta_i:=\|\phi_i(t)\|_2$. The rotation matrix is given by Rodrigues' formula and the left Jacobian has the closed form as
\begin{equation}
R_i(t)
=
I
+
\frac{\sin\theta_i}{\theta_i}\hat\phi_i
+
\frac{1-\cos\theta_i}{\theta_i^2}\hat\phi_i^2,
\tag{S2}
\end{equation}

\begin{equation}
J(\phi_i(t))
=
I
+
\frac{1-\cos\theta_i}{\theta_i^2}\hat\phi_i
+
\frac{\theta_i-\sin\theta_i}{\theta_i^3}\hat\phi_i^2.
\tag{S3}
\end{equation}
Since $\mathbf{u}_i(t)=\zeta_i(t)\Delta t$ with $\Delta t=t-\mu_{t,i}$, we have $\mathbf{u}_i(\mu_{t,i})=\mathbf{0}$ and thus
\begin{equation}
\mathbf{T}_i(\mu_{t,i}\!\to\!\mu_{t,i})=\exp(\hat{\mathbf{0}})=\mathbf{I}.
\end{equation}
Therefore, $\mathbf{T}_i(\mu_{t,i}\!\to\! t)$ is a relative transform defined with respect to the primitive central time $\mu_{t,i}$.

\subsection{Gauge-fixing derivation for the local anchor parameterization}
\label{supp:gauge_fixing_derivation}

Before gauge fixing, the local anchor-centered deformation can be written as
\begin{equation}
\mu_i(t)=R_i(t)(\mu_i-a_i)+a_i+p_i(t),
\end{equation}
or equivalently,
\begin{equation}
\mu_i(t)=R_i(t)\mu_i+(I-R_i(t))a_i+p_i(t).
\tag{S4}
\end{equation}
Thus, the anchor appears only through the term $(I-R_i(t))a_i$. Let $\bar{\omega}_i(t)$ denote the unit rotation axis direction. We decompose the anchor into axis-parallel and axis-orthogonal components:
\begin{equation}
a_i=a_{i,\parallel}(t)+a_{i,\perp}(t),
\end{equation}
\begin{equation}
a_{i,\parallel}(t)=\langle a_i,\bar{\omega}_i(t)\rangle \bar{\omega}_i(t),
\qquad
a_{i,\perp}(t)=a_i-\langle a_i,\bar{\omega}_i(t)\rangle \bar{\omega}_i(t).
\tag{S5}
\end{equation}
Because a rotation leaves its own axis invariant,
\begin{equation}
R_i(t)\bar{\omega}_i(t)=\bar{\omega}_i(t),
\qquad
(I-R_i(t))\bar{\omega}_i(t)=0,
\end{equation}
which implies
\begin{equation}
(I-R_i(t))a_{i,\parallel}(t)=0.
\end{equation}
Hence, the axis-parallel component of $a_i$ lies in the null space of $(I-R_i(t))$ and is unidentifiable.
Substituting $a_i=a_{i,\parallel}(t)+a_{i,\perp}(t)$ into Eq.~(S4) yields
\begin{equation}
\mu_i(t)
=
R_i(t)\mu_i
+
(I-R_i(t))a_{i,\perp}(t)
+
p_i(t).
\tag{S6}
\end{equation}
However, even after removing $a_{i,\parallel}(t)$, there remains a residual ambiguity between the in-plane anchor term and the in-plane translation term. To see this, decompose the translation as
\begin{equation}
p_i(t)=p_{i,\parallel}(t)+p_{i,\perp}(t),
\end{equation}
where $p_{i,\parallel}(t)$ is axis-aligned and $p_{i,\perp}(t)$ lies in the plane orthogonal to $\bar{\omega}_i(t)$. For any $\delta_i(t)$ orthogonal to $\bar{\omega}_i(t)$, define
\begin{equation}
a'_{i,\perp}(t)=a_{i,\perp}(t)+\delta_i(t),
\qquad
p'_{i,\perp}(t)=p_{i,\perp}(t)-(I-R_i(t))\delta_i(t).
\end{equation}
Then
\begin{equation}
(I-R_i(t))a'_{i,\perp}(t)+p'_{i,\perp}(t)
=
(I-R_i(t))a_{i,\perp}(t)+p_{i,\perp}(t),
\end{equation}
so the resulting deformation remains unchanged. To remove this residual redundancy, we retain only the axis-aligned translational component
\begin{equation}
\upsilon_{i,\parallel}(t)
=
\langle \upsilon_i(t),\bar{\omega}_i(t)\rangle\,\bar{\omega}_i(t),
\qquad
p_{i,\parallel}(t)=J(\phi_i(t))\,\upsilon_{i,\parallel}(t).
\tag{S7}
\end{equation}
The final gauge-fixed deformation, which recovers the final local-anchor deformation formula in the main paper, is given by
\begin{equation}
\mu_i(t)
=
R_i(t)(\mu_i-a_{i,\perp}(t))
+
a_{i,\perp}(t)
+
p_{i,\parallel}(t),
\qquad
\Sigma_i(t)=R_i(t)\Sigma_iR_i(t)^\top.
\tag{S8}
\end{equation}

\subsection{Implementation-aligned formulation}
\label{supp:implementation_gauge}

In implementation, we compute the same gauge directly in projection form using the instantaneous angular and translational coefficients $\omega_i(t)$ and $\nu_i(t)$, which avoids explicitly forming the unit axis in the near-zero angular regime. For $\|\omega_i(t)\|^2>\varepsilon$, the anchor projection and axis-aligned translational component are computed as
\begin{equation}
a_{i,\perp}(t)=a_i-\frac{a_i^\top \omega_i(t)}{\|\omega_i(t)\|^2}\omega_i(t),\qquad \nu_{i,\parallel}(t)=\frac{\nu_i(t)^\top \omega_i(t)}{\|\omega_i(t)\|^2}\omega_i(t).
\tag{S9}
\end{equation}
When the rotation axis becomes numerically unreliable, we treat this case as 
\begin{equation}
a_{i,\perp}(t)=a_i,
\qquad
\nu_{i,\parallel}(t)=\nu_i(t), 
\tag{S10}
\end{equation}
where $\|\omega_i(t)\|^2\le\varepsilon$.
It matches the translation-dominant small-angle regime below. In our implementation, we use $\varepsilon=10^{-8}$ for both the angular-magnitude test in Eq.~(S10) and the small-angle branch below, together with a denominator clamp of $10^{-10}$ for numerical safety.

\subsection{Small-angle numerical stabilization}
\label{supp:small_angle}

The closed-form expressions in Eqs.~(S1)--(S3) involve coefficients that become numerically unstable when $\theta$ is close to zero. Let $\theta_i=\|\phi_i(t)\|_2$. We define each coefficient as
\begin{equation}
A(\theta_i):=\frac{\sin\theta_i}{\theta_i},
\qquad
B(\theta_i):=\frac{1-\cos\theta_i}{\theta_i^2},
\qquad
C(\theta_i):=\frac{\theta_i-\sin\theta_i}{\theta_i^3}.
\tag{S11}
\end{equation}
To ensure numerical stability, we evaluate $A(\theta_i)$ and $B(\theta_i)$ using the $\operatorname{sinc}(x)=\sin(\pi x)/(\pi x)$ function:
\begin{equation}
A(\theta_i)=\operatorname{sinc}\!\left(\frac{\theta_i}{\pi}\right),
\qquad
B(\theta_i)=\frac12\,\operatorname{sinc}^2\!\left(\frac{\theta_i}{2\pi}\right).
\tag{S12}
\end{equation}
Since the $\operatorname{sinc}$ function cannot be directly applied to the remaining coefficient $C(\theta_i)$, we approximate it using a Taylor expansion for small angles to prevent numerical instability:
\begin{equation}
C(\theta_i)
=
\begin{cases}
\dfrac16-\dfrac{\theta_i^2}{120}, & \theta_i^2<10^{-8},\\[8pt]
\dfrac{\theta_i-\sin\theta_i}{\theta_i^3+10^{-10}}, & \text{otherwise}.
\end{cases}
\tag{S13}
\end{equation}

\section{Supplement for Motion-aware Relocation} 
\label{supp:relocation}

As discussed in the main paper, we employ a motion-aware relocation strategy to effectively recycle free-riding primitives. To select crucial regions for relocation, we define a sampling score $s_i$ for each alive primitive $i \in \mathcal{A}$ as the sum of four normalized metrics:
\begin{equation}
s_i \propto p_i^{(\alpha)} + p_i^{(d\nu)} + p_i^{(2D)} + p_i^{(t)}, \quad \text{where } p_i^{(\cdot)} \in [0, 1]
\end{equation}
These metrics jointly evaluate the intrinsic state (opacity and motion magnitude) and optimization error signals (spatial and temporal gradients) of the primitive:
\begin{align}
p_i^{(\alpha)} &= \frac{\sigma(\alpha_i)}{\sum_{j \in \mathcal{A}} \sigma(\alpha_j)}, &
p_i^{(d\nu)} &= \frac{\|\nu_i\| \exp(s_{t,i})}{\sum_{j \in \mathcal{A}} \|\nu_j\| \exp(s_{t,j})}, \\
p_i^{(2D)} &= \frac{\nabla_{\text{2D}}\mu_i}{\sum_{j \in \mathcal{A}} \nabla_{\text{2D}}\mu_j}, &
p_i^{(t)} &= \frac{\nabla_t \mu_i}{\sum_{j \in \mathcal{A}} \nabla_t \mu_j},
\end{align}
where $\sigma(\alpha_i)$ is the sigmoid opacity, $\nu_i$ and $s_{t,i}$ are the local translation and temporal scale, and $\nabla_{\text{2D}}\mu_i$ and $\nabla_t \mu_i$ denote the accumulated 2D spatial and temporal gradients, respectively. 
To replace inactive primitives, sampled ones are cloned with identically duplicated properties, except for their scales, which are reduced to approximately 66\% of the original.
\end{document}